\documentclass[10pt,twocolumn,letterpaper]{article}

\usepackage[datasets]{wacv} %

\usepackage{iftex}
\ifXeTeX
  \usepackage{fontspec}
\fi

\usepackage[protrusion=true]{microtype}
\usepackage{enumitem}
\usepackage{needspace}
\usepackage{amsmath,amssymb,amsthm}
\usepackage{booktabs}
\usepackage{makecell}
\usepackage{multirow}
\usepackage{tikz}
\usetikzlibrary{arrows.meta,positioning,fit,backgrounds,decorations.pathreplacing}
\usepackage{pgfplots}
\pgfplotsset{compat=1.18}

\definecolor{wacvblue}{rgb}{0.21,0.49,0.74}
\definecolor{evictred}{rgb}{0.72,0.27,0.23}
\definecolor{newgreen}{rgb}{0.18,0.55,0.34}
\definecolor{diagorange}{rgb}{0.85,0.50,0.10}
\colorlet{pcpcool}{wacvblue}
\colorlet{pcpwarm}{evictred}
\usepackage[pagebackref,breaklinks,colorlinks,allcolors=wacvblue]{hyperref}

\newtheorem{theorem}{Theorem}

\newtheorem{definition}{Definition}
\theoremstyle{remark}

\newcommand{\Sph}{\mathbb{S}^{d-1}}

\newcommand{\ip}[2]{\langle #1,\,#2\rangle}
\newcommand{\score}{s}
\newcommand{\energy}{g}

\newcommand{\Set}{\Gamma}
\newcommand{\cov}{\tau}
\newcommand{\birth}{b}
\newcommand{\qid}{\iota}
\newcommand{\nver}{R}

\makeatletter
\newcommand{\pcpdef}[2]{\expandafter\gdef\csname pcp@#1\endcsname{#2}}
\newcommand{\pcpdash}{\ifmmode\mbox{--}\else--\fi}
\newcommand{\N}[1]{\csname pcp@#1\endcsname}
\makeatother

\pcpdef{cd.cars-clip-m20.countshare}{82}
\pcpdef{cd.cars-clip-m20.mean.add}{0.93}
\pcpdef{cd.cars-clip-m20.mean.dcov}{+0.003}
\pcpdef{cd.cars-clip-m20.mean.flip}{2.11}
\pcpdef{cd.cars-clip-m20.mean.reach}{67.3}
\pcpdef{cd.cars-clip-m20.mean.rev}{8.5}
\pcpdef{cd.cars-clip-m20.sum.dcov}{-0.077}
\pcpdef{cd.cars-clip-m20.sum.flip}{5.41}
\pcpdef{cd.cars-clip-m20.sum.reach}{85.1}
\pcpdef{cd.cars-clip-m20.sum.rev}{34.6}
\pcpdef{cd.cars-siglip-m20.countshare}{39}
\pcpdef{cd.cars-siglip-m20.mean.add}{0.43}
\pcpdef{cd.cars-siglip-m20.mean.dcov}{+0.000}
\pcpdef{cd.cars-siglip-m20.mean.flip}{2.23}
\pcpdef{cd.cars-siglip-m20.mean.reach}{59.9}
\pcpdef{cd.cars-siglip-m20.mean.rev}{19.7}
\pcpdef{cd.cars-siglip-m20.sum.dcov}{-0.060}
\pcpdef{cd.cars-siglip-m20.sum.flip}{3.39}
\pcpdef{cd.cars-siglip-m20.sum.reach}{62.7}
\pcpdef{cd.cars-siglip-m20.sum.rev}{36.5}
\pcpdef{cd.cells}{25}
\pcpdef{cd.control.v1match}{25/25}
\pcpdef{cd.control.zerogrids}{5/5}
\pcpdef{cd.countshare.range}{39\pcpdash{}82}
\pcpdef{cd.food-siglip-m20.countshare}{49}
\pcpdef{cd.food-siglip-m20.mean.add}{0.84}
\pcpdef{cd.food-siglip-m20.mean.dcov}{-0.006}
\pcpdef{cd.food-siglip-m20.mean.flip}{5.08}
\pcpdef{cd.food-siglip-m20.mean.reach}{64.8}
\pcpdef{cd.food-siglip-m20.mean.rev}{32.5}
\pcpdef{cd.food-siglip-m20.sum.dcov}{-0.052}
\pcpdef{cd.food-siglip-m20.sum.flip}{6.72}
\pcpdef{cd.food-siglip-m20.sum.reach}{66.9}
\pcpdef{cd.food-siglip-m20.sum.rev}{50.1}
\pcpdef{cd.food-siglip-m40.countshare}{49}
\pcpdef{cd.food-siglip-m40.mean.add}{0.73}
\pcpdef{cd.food-siglip-m40.mean.dcov}{-0.009}
\pcpdef{cd.food-siglip-m40.mean.flip}{4.57}
\pcpdef{cd.food-siglip-m40.mean.reach}{62.6}
\pcpdef{cd.food-siglip-m40.mean.rev}{34.1}
\pcpdef{cd.food-siglip-m40.sum.dcov}{-0.055}
\pcpdef{cd.food-siglip-m40.sum.flip}{6.08}
\pcpdef{cd.food-siglip-m40.sum.reach}{65.8}
\pcpdef{cd.food-siglip-m40.sum.rev}{52.8}
\pcpdef{cd.food-siglip-m80.countshare}{49}
\pcpdef{cd.food-siglip-m80.mean.add}{0.71}
\pcpdef{cd.food-siglip-m80.mean.dcov}{-0.010}
\pcpdef{cd.food-siglip-m80.mean.flip}{4.51}
\pcpdef{cd.food-siglip-m80.mean.reach}{62.1}
\pcpdef{cd.food-siglip-m80.mean.rev}{34.7}
\pcpdef{cd.food-siglip-m80.sum.dcov}{-0.053}
\pcpdef{cd.food-siglip-m80.sum.flip}{5.98}
\pcpdef{cd.food-siglip-m80.sum.reach}{65.3}
\pcpdef{cd.food-siglip-m80.sum.rev}{53.5}
\pcpdef{cd.grids}{5}
\pcpdef{cd.grids.word}{five}
\pcpdef{cd.mean.add.range}{0.4\pcpdash{}0.9}
\pcpdef{cd.mean.dcov.absmax}{0.010}
\pcpdef{cd.mean.dcov.absmax.r2}{0.01}
\pcpdef{cd.mean.dcov.hi}{+0.003}
\pcpdef{cd.mean.dcov.lo}{-0.010}
\pcpdef{cd.mean.flip.range}{2.11\pcpdash{}5.08}
\pcpdef{cd.mean.reach.range}{60\pcpdash{}67}
\pcpdef{cd.mean.rev.range}{8.5\pcpdash{}34.7}
\pcpdef{cd.sum.dcov.hi}{-0.052}
\pcpdef{cd.sum.dcov.lo}{-0.077}
\pcpdef{cd.sum.flip.range}{3.39\pcpdash{}6.72}
\pcpdef{cd.sum.reach.range}{63\pcpdash{}85}
\pcpdef{cd.sum.rev.range}{34.6\pcpdash{}53.5}
\pcpdef{cd.table.churn.body}{%
\multicolumn{6}{@{}l@{}}{\emph{sum form (carries the $T\log|V|$ cardinality offset)}}\\
\;Stanford Cars / CLIP & 5.41 & 34.6 & 0.00 & 85.1 & -0.077\\
\;Stanford Cars / SigLIP & 3.39 & 36.5 & 0.00 & 62.7 & -0.060\\
\;Food-101 / SigLIP $m{=}20$ & 6.72 & 50.1 & 0.00 & 66.9 & -0.052\\
\;Food-101 / SigLIP $m{=}40$ & 6.08 & 52.8 & 0.00 & 65.8 & -0.055\\
\;Food-101 / SigLIP $m{=}80$ & 5.98 & 53.5 & 0.00 & 65.3 & -0.053\\
\midrule
\multicolumn{6}{@{}l@{}}{\emph{count-only counterfactual (cardinality term alone; composition frozen)}}\\
\;Stanford Cars / CLIP & 5.02 & 32.0 & 0.00 & 90.2 & -0.089\\
\;Stanford Cars / SigLIP & 2.27 & 24.5 & 0.00 & 67.4 & -0.140\\
\;Food-101 / SigLIP $m{=}20$ & 4.13 & 30.8 & 0.00 & 72.8 & -0.069\\
\;Food-101 / SigLIP $m{=}40$ & 3.84 & 33.3 & 0.00 & 69.9 & -0.062\\
\;Food-101 / SigLIP $m{=}80$ & 3.79 & 33.9 & 0.00 & 69.2 & -0.062\\
\midrule
\multicolumn{6}{@{}l@{}}{\emph{mean form (cardinality-corrected; \textbf{primary})}}\\
\;Stanford Cars / CLIP & 2.11 & 8.5 & 0.93 & 67.3 & +0.003\\
\;Stanford Cars / SigLIP & 2.23 & 19.7 & 0.43 & 59.9 & +0.000\\
\;Food-101 / SigLIP $m{=}20$ & 5.08 & 32.5 & 0.84 & 64.8 & -0.006\\
\;Food-101 / SigLIP $m{=}40$ & 4.57 & 34.1 & 0.73 & 62.6 & -0.009\\
\;Food-101 / SigLIP $m{=}80$ & 4.51 & 34.7 & 0.71 & 62.1 & -0.010\\}
\pcpdef{cd.table.design.body}{%
Stanford Cars / CLIP & 98 & 196 & 98 & 8{,}007 & 4{,}003 & 122{,}983 & 661{,}703\\
Stanford Cars / SigLIP & 98 & 196 & 98 & 8{,}007 & 4{,}003 & 72{,}922 & 711{,}764\\
Food-101 / SigLIP $m{=}20$ & 50 & 101 & 50 & 25{,}224 & 12{,}486 & 170{,}111 & 1{,}091{,}089\\
Food-101 / SigLIP $m{=}40$ & 50 & 101 & 50 & 25{,}224 & 12{,}486 & 145{,}569 & 1{,}115{,}631\\
Food-101 / SigLIP $m{=}80$ & 50 & 101 & 50 & 25{,}224 & 12{,}486 & 141{,}404 & 1{,}119{,}796\\}
\pcpdef{cr.food-siglip-m20.K}{101}
\pcpdef{cr.food-siglip-m20.closed.ci}{46.3\,[38.0, 52.2]}
\pcpdef{cr.food-siglip-m20.closed.p025}{38.0}
\pcpdef{cr.food-siglip-m20.closed.p975}{52.2}
\pcpdef{cr.food-siglip-m20.closed.ratio-of-means}{46.6}
\pcpdef{cr.food-siglip-m20.closed.undefined-count}{0}
\pcpdef{cr.food-siglip-m20.closed.valid-mean}{46.3}
\pcpdef{cr.food-siglip-m20.closed.valid-n}{20}
\pcpdef{cr.food-siglip-m20.dir.above-refresh}{20/20}
\pcpdef{cr.food-siglip-m20.dir.below-pre}{20/20}
\pcpdef{cr.food-siglip-m20.dir.between}{20/20}
\pcpdef{cr.food-siglip-m20.frozen.add.ci}{0.86\,[0.74, 0.97]}
\pcpdef{cr.food-siglip-m20.frozen.add.max}{0.98}
\pcpdef{cr.food-siglip-m20.frozen.add.mean}{0.86}
\pcpdef{cr.food-siglip-m20.frozen.add.min}{0.70}
\pcpdef{cr.food-siglip-m20.frozen.add.p025}{0.74}
\pcpdef{cr.food-siglip-m20.frozen.add.p975}{0.97}
\pcpdef{cr.food-siglip-m20.frozen.add.sd}{0.067}
\pcpdef{cr.food-siglip-m20.frozen.bit.ci}{5.08\,[4.54, 5.66]}
\pcpdef{cr.food-siglip-m20.frozen.bit.max}{5.72}
\pcpdef{cr.food-siglip-m20.frozen.bit.mean}{5.08}
\pcpdef{cr.food-siglip-m20.frozen.bit.min}{4.30}
\pcpdef{cr.food-siglip-m20.frozen.bit.p025}{4.54}
\pcpdef{cr.food-siglip-m20.frozen.bit.p975}{5.66}
\pcpdef{cr.food-siglip-m20.frozen.bit.sd}{0.306}
\pcpdef{cr.food-siglip-m20.frozen.reach.ci}{65.2\,[62.5, 66.8]}
\pcpdef{cr.food-siglip-m20.frozen.reach.max}{66.8}
\pcpdef{cr.food-siglip-m20.frozen.reach.mean}{65.2}
\pcpdef{cr.food-siglip-m20.frozen.reach.min}{61.5}
\pcpdef{cr.food-siglip-m20.frozen.reach.p025}{62.5}
\pcpdef{cr.food-siglip-m20.frozen.reach.p975}{66.8}
\pcpdef{cr.food-siglip-m20.frozen.reach.sd}{1.33}
\pcpdef{cr.food-siglip-m20.frozen.rev.ci}{31.1\,[29.9, 32.4]}
\pcpdef{cr.food-siglip-m20.frozen.rev.max}{32.4}
\pcpdef{cr.food-siglip-m20.frozen.rev.mean}{31.1}
\pcpdef{cr.food-siglip-m20.frozen.rev.min}{29.7}
\pcpdef{cr.food-siglip-m20.frozen.rev.p025}{29.9}
\pcpdef{cr.food-siglip-m20.frozen.rev.p975}{32.4}
\pcpdef{cr.food-siglip-m20.frozen.rev.sd}{0.82}
\pcpdef{cr.food-siglip-m20.gate.repro-values}{24}
\pcpdef{cr.food-siglip-m20.label}{Food-101 / SigLIP}
\pcpdef{cr.food-siglip-m20.pfx.coverage.macro.ci}{0.9265\,[0.9163, 0.9334]}
\pcpdef{cr.food-siglip-m20.pfx.coverage.macro.mean}{0.9265}
\pcpdef{cr.food-siglip-m20.pfx.coverage.macro.min}{0.9160}
\pcpdef{cr.food-siglip-m20.pfx.coverage.macro.p025}{0.9163}
\pcpdef{cr.food-siglip-m20.pfx.coverage.macro.p975}{0.9334}
\pcpdef{cr.food-siglip-m20.pfx.coverage.macro.sd}{0.00489}
\pcpdef{cr.food-siglip-m20.pfx.coverage.marginal.ci}{0.9265\,[0.9163, 0.9334]}
\pcpdef{cr.food-siglip-m20.pfx.coverage.marginal.mean}{0.9265}
\pcpdef{cr.food-siglip-m20.pfx.coverage.marginal.min}{0.9160}
\pcpdef{cr.food-siglip-m20.pfx.coverage.marginal.p025}{0.9163}
\pcpdef{cr.food-siglip-m20.pfx.coverage.marginal.p975}{0.9334}
\pcpdef{cr.food-siglip-m20.pfx.coverage.marginal.sd}{0.00489}
\pcpdef{cr.food-siglip-m20.pfx.coverage.p10.ci}{0.858\,[0.844, 0.878]}
\pcpdef{cr.food-siglip-m20.pfx.coverage.p10.mean}{0.858}
\pcpdef{cr.food-siglip-m20.pfx.coverage.p10.min}{0.843}
\pcpdef{cr.food-siglip-m20.pfx.coverage.p10.p025}{0.844}
\pcpdef{cr.food-siglip-m20.pfx.coverage.p10.p975}{0.878}
\pcpdef{cr.food-siglip-m20.pfx.coverage.p10.sd}{0.0105}
\pcpdef{cr.food-siglip-m20.pfx.coverage.worst.ci}{0.718\,[0.587, 0.788]}
\pcpdef{cr.food-siglip-m20.pfx.coverage.worst.mean}{0.718}
\pcpdef{cr.food-siglip-m20.pfx.coverage.worst.min}{0.536}
\pcpdef{cr.food-siglip-m20.pfx.coverage.worst.p025}{0.587}
\pcpdef{cr.food-siglip-m20.pfx.coverage.worst.p975}{0.788}
\pcpdef{cr.food-siglip-m20.pfx.coverage.worst.sd}{0.0601}
\pcpdef{cr.food-siglip-m20.pfx.size.ci}{12.29\,[10.68, 13.39]}
\pcpdef{cr.food-siglip-m20.pfx.size.mean}{12.29}
\pcpdef{cr.food-siglip-m20.pfx.size.sd}{0.765}
\pcpdef{cr.food-siglip-m20.pre-minus-pfx.ci}{+1.25\,[+0.83, +1.60]}
\pcpdef{cr.food-siglip-m20.pre-minus-pfx.max}{+1.62}
\pcpdef{cr.food-siglip-m20.pre-minus-pfx.mean}{+1.25}
\pcpdef{cr.food-siglip-m20.pre-minus-pfx.min}{+0.80}
\pcpdef{cr.food-siglip-m20.pre-minus-pfx.p025}{+0.83}
\pcpdef{cr.food-siglip-m20.pre-minus-pfx.p975}{+1.60}
\pcpdef{cr.food-siglip-m20.pre-minus-pfx.sd}{0.276}
\pcpdef{cr.food-siglip-m20.pre.coverage.macro.ci}{0.9273\,[0.9159, 0.9335]}
\pcpdef{cr.food-siglip-m20.pre.coverage.macro.mean}{0.9273}
\pcpdef{cr.food-siglip-m20.pre.coverage.macro.min}{0.9151}
\pcpdef{cr.food-siglip-m20.pre.coverage.macro.p025}{0.9159}
\pcpdef{cr.food-siglip-m20.pre.coverage.macro.p975}{0.9335}
\pcpdef{cr.food-siglip-m20.pre.coverage.macro.sd}{0.00504}
\pcpdef{cr.food-siglip-m20.pre.size.ci}{13.53\,[11.76, 14.81]}
\pcpdef{cr.food-siglip-m20.pre.size.mean}{13.53}
\pcpdef{cr.food-siglip-m20.pre.size.sd}{0.812}
\pcpdef{cr.food-siglip-m20.raw.size.ci}{18.13\,[15.78, 19.59]}
\pcpdef{cr.food-siglip-m20.raw.size.mean}{18.13}
\pcpdef{cr.food-siglip-m20.raw.size.sd}{1.187}
\pcpdef{cr.food-siglip-m20.refresh-minus-pfx.ci}{-1.43\,[-1.74, -1.07]}
\pcpdef{cr.food-siglip-m20.refresh-minus-pfx.max}{-1.06}
\pcpdef{cr.food-siglip-m20.refresh-minus-pfx.mean}{-1.43}
\pcpdef{cr.food-siglip-m20.refresh-minus-pfx.min}{-1.77}
\pcpdef{cr.food-siglip-m20.refresh-minus-pfx.p025}{-1.74}
\pcpdef{cr.food-siglip-m20.refresh-minus-pfx.p975}{-1.07}
\pcpdef{cr.food-siglip-m20.refresh-minus-pfx.sd}{0.206}
\pcpdef{cr.food-siglip-m20.refresh.coverage.macro.ci}{0.9276\,[0.9170, 0.9348]}
\pcpdef{cr.food-siglip-m20.refresh.coverage.macro.mean}{0.9276}
\pcpdef{cr.food-siglip-m20.refresh.coverage.macro.min}{0.9166}
\pcpdef{cr.food-siglip-m20.refresh.coverage.macro.p025}{0.9170}
\pcpdef{cr.food-siglip-m20.refresh.coverage.macro.p975}{0.9348}
\pcpdef{cr.food-siglip-m20.refresh.coverage.macro.sd}{0.00497}
\pcpdef{cr.food-siglip-m20.refresh.reach.ci}{77.9\,[76.1, 81.4]}
\pcpdef{cr.food-siglip-m20.refresh.reach.max}{81.6}
\pcpdef{cr.food-siglip-m20.refresh.reach.mean}{77.9}
\pcpdef{cr.food-siglip-m20.refresh.reach.min}{75.8}
\pcpdef{cr.food-siglip-m20.refresh.reach.p025}{76.1}
\pcpdef{cr.food-siglip-m20.refresh.reach.p975}{81.4}
\pcpdef{cr.food-siglip-m20.refresh.reach.sd}{1.60}
\pcpdef{cr.food-siglip-m20.refresh.size.ci}{10.86\,[9.49, 11.99]}
\pcpdef{cr.food-siglip-m20.refresh.size.mean}{10.86}
\pcpdef{cr.food-siglip-m20.refresh.size.sd}{0.733}
\pcpdef{cr.gates.passed}{9}
\pcpdef{cr.gates.passed.word}{nine}
\pcpdef{cr.meta.cells}{40}
\pcpdef{cr.meta.design}{minimum}
\pcpdef{cr.meta.grids}{2}
\pcpdef{cr.meta.grids.word}{two}
\pcpdef{cr.meta.n-calibration-draws}{20}
\pcpdef{cr.meta.n-calibration-draws.word}{twenty}
\pcpdef{cr.meta.n-partitions}{1}
\pcpdef{cr.meta.n-partitions.word}{one}
\pcpdef{cr.meta.pfx-below-pre-cells}{40/40}
\pcpdef{cr.meta.pfx-between-cells}{40/40}
\pcpdef{cr.meta.zerochurn-cells}{40/40}
\pcpdef{cr.scope.calibration-seed-list}{8100000\pcpdash{}8100019}
\pcpdef{cr.scope.churn-subset}{future}
\pcpdef{cr.scope.grid-list}{Food-101 / SigLIP; ImageNet-Sketch / SigLIP}
\pcpdef{cr.scope.interval-definition}{calibration-draw variation intervals conditional on this fixed corpus, model, evaluation split, and label partition}
\pcpdef{cr.scope.partition-list}{100}
\pcpdef{cr.scope.row-population}{the classwise calibration sample only, drawn without replacement within class from the fixed eligible pool}
\pcpdef{cr.scope.seed-range}{8{,}100{,}000\pcpdash{}8{,}100{,}999}
\pcpdef{cr.sketch-siglip-m20.K}{930}
\pcpdef{cr.sketch-siglip-m20.closed.ci}{37.0\,[31.2, 41.2]}
\pcpdef{cr.sketch-siglip-m20.closed.p025}{31.2}
\pcpdef{cr.sketch-siglip-m20.closed.p975}{41.2}
\pcpdef{cr.sketch-siglip-m20.closed.ratio-of-means}{36.9}
\pcpdef{cr.sketch-siglip-m20.closed.undefined-count}{0}
\pcpdef{cr.sketch-siglip-m20.closed.valid-mean}{37.0}
\pcpdef{cr.sketch-siglip-m20.closed.valid-n}{20}
\pcpdef{cr.sketch-siglip-m20.dir.above-refresh}{20/20}
\pcpdef{cr.sketch-siglip-m20.dir.below-pre}{20/20}
\pcpdef{cr.sketch-siglip-m20.dir.between}{20/20}
\pcpdef{cr.sketch-siglip-m20.frozen.add.ci}{0.09\,[0.09, 0.10]}
\pcpdef{cr.sketch-siglip-m20.frozen.add.max}{0.10}
\pcpdef{cr.sketch-siglip-m20.frozen.add.mean}{0.09}
\pcpdef{cr.sketch-siglip-m20.frozen.add.min}{0.09}
\pcpdef{cr.sketch-siglip-m20.frozen.add.p025}{0.09}
\pcpdef{cr.sketch-siglip-m20.frozen.add.p975}{0.10}
\pcpdef{cr.sketch-siglip-m20.frozen.add.sd}{0.004}
\pcpdef{cr.sketch-siglip-m20.frozen.bit.ci}{0.88\,[0.81, 0.97]}
\pcpdef{cr.sketch-siglip-m20.frozen.bit.max}{0.99}
\pcpdef{cr.sketch-siglip-m20.frozen.bit.mean}{0.88}
\pcpdef{cr.sketch-siglip-m20.frozen.bit.min}{0.79}
\pcpdef{cr.sketch-siglip-m20.frozen.bit.p025}{0.81}
\pcpdef{cr.sketch-siglip-m20.frozen.bit.p975}{0.97}
\pcpdef{cr.sketch-siglip-m20.frozen.bit.sd}{0.043}
\pcpdef{cr.sketch-siglip-m20.frozen.reach.ci}{70.5\,[69.2, 72.0]}
\pcpdef{cr.sketch-siglip-m20.frozen.reach.max}{72.0}
\pcpdef{cr.sketch-siglip-m20.frozen.reach.mean}{70.5}
\pcpdef{cr.sketch-siglip-m20.frozen.reach.min}{69.0}
\pcpdef{cr.sketch-siglip-m20.frozen.reach.p025}{69.2}
\pcpdef{cr.sketch-siglip-m20.frozen.reach.p975}{72.0}
\pcpdef{cr.sketch-siglip-m20.frozen.reach.sd}{0.75}
\pcpdef{cr.sketch-siglip-m20.frozen.rev.ci}{42.0\,[40.5, 43.3]}
\pcpdef{cr.sketch-siglip-m20.frozen.rev.max}{43.3}
\pcpdef{cr.sketch-siglip-m20.frozen.rev.mean}{42.0}
\pcpdef{cr.sketch-siglip-m20.frozen.rev.min}{40.4}
\pcpdef{cr.sketch-siglip-m20.frozen.rev.p025}{40.5}
\pcpdef{cr.sketch-siglip-m20.frozen.rev.p975}{43.3}
\pcpdef{cr.sketch-siglip-m20.frozen.rev.sd}{0.91}
\pcpdef{cr.sketch-siglip-m20.gate.repro-values}{24}
\pcpdef{cr.sketch-siglip-m20.label}{ImageNet-Sketch / SigLIP}
\pcpdef{cr.sketch-siglip-m20.pfx.coverage.macro.ci}{0.9030\,[0.9006, 0.9054]}
\pcpdef{cr.sketch-siglip-m20.pfx.coverage.macro.mean}{0.9030}
\pcpdef{cr.sketch-siglip-m20.pfx.coverage.macro.min}{0.9002}
\pcpdef{cr.sketch-siglip-m20.pfx.coverage.macro.p025}{0.9006}
\pcpdef{cr.sketch-siglip-m20.pfx.coverage.macro.p975}{0.9054}
\pcpdef{cr.sketch-siglip-m20.pfx.coverage.macro.sd}{0.00138}
\pcpdef{cr.sketch-siglip-m20.pfx.coverage.marginal.ci}{0.9030\,[0.9007, 0.9055]}
\pcpdef{cr.sketch-siglip-m20.pfx.coverage.marginal.mean}{0.9030}
\pcpdef{cr.sketch-siglip-m20.pfx.coverage.marginal.min}{0.9002}
\pcpdef{cr.sketch-siglip-m20.pfx.coverage.marginal.p025}{0.9007}
\pcpdef{cr.sketch-siglip-m20.pfx.coverage.marginal.p975}{0.9055}
\pcpdef{cr.sketch-siglip-m20.pfx.coverage.marginal.sd}{0.00140}
\pcpdef{cr.sketch-siglip-m20.pfx.coverage.p10.ci}{0.782\,[0.769, 0.800]}
\pcpdef{cr.sketch-siglip-m20.pfx.coverage.p10.mean}{0.782}
\pcpdef{cr.sketch-siglip-m20.pfx.coverage.p10.min}{0.769}
\pcpdef{cr.sketch-siglip-m20.pfx.coverage.p10.p025}{0.769}
\pcpdef{cr.sketch-siglip-m20.pfx.coverage.p10.p975}{0.800}
\pcpdef{cr.sketch-siglip-m20.pfx.coverage.p10.sd}{0.0110}
\pcpdef{cr.sketch-siglip-m20.pfx.coverage.worst.ci}{0.478\,[0.423, 0.520]}
\pcpdef{cr.sketch-siglip-m20.pfx.coverage.worst.mean}{0.478}
\pcpdef{cr.sketch-siglip-m20.pfx.coverage.worst.min}{0.423}
\pcpdef{cr.sketch-siglip-m20.pfx.coverage.worst.p025}{0.423}
\pcpdef{cr.sketch-siglip-m20.pfx.coverage.worst.p975}{0.520}
\pcpdef{cr.sketch-siglip-m20.pfx.coverage.worst.sd}{0.0351}
\pcpdef{cr.sketch-siglip-m20.pfx.size.ci}{12.04\,[11.38, 12.78]}
\pcpdef{cr.sketch-siglip-m20.pfx.size.mean}{12.04}
\pcpdef{cr.sketch-siglip-m20.pfx.size.sd}{0.371}
\pcpdef{cr.sketch-siglip-m20.pre-minus-pfx.ci}{+1.89\,[+1.61, +2.09]}
\pcpdef{cr.sketch-siglip-m20.pre-minus-pfx.max}{+2.14}
\pcpdef{cr.sketch-siglip-m20.pre-minus-pfx.mean}{+1.89}
\pcpdef{cr.sketch-siglip-m20.pre-minus-pfx.min}{+1.47}
\pcpdef{cr.sketch-siglip-m20.pre-minus-pfx.p025}{+1.61}
\pcpdef{cr.sketch-siglip-m20.pre-minus-pfx.p975}{+2.09}
\pcpdef{cr.sketch-siglip-m20.pre-minus-pfx.sd}{0.141}
\pcpdef{cr.sketch-siglip-m20.pre.coverage.macro.ci}{0.9021\,[0.8995, 0.9043]}
\pcpdef{cr.sketch-siglip-m20.pre.coverage.macro.mean}{0.9021}
\pcpdef{cr.sketch-siglip-m20.pre.coverage.macro.min}{0.8993}
\pcpdef{cr.sketch-siglip-m20.pre.coverage.macro.p025}{0.8995}
\pcpdef{cr.sketch-siglip-m20.pre.coverage.macro.p975}{0.9043}
\pcpdef{cr.sketch-siglip-m20.pre.coverage.macro.sd}{0.00139}
\pcpdef{cr.sketch-siglip-m20.pre.size.ci}{13.93\,[13.17, 14.71]}
\pcpdef{cr.sketch-siglip-m20.pre.size.mean}{13.93}
\pcpdef{cr.sketch-siglip-m20.pre.size.sd}{0.439}
\pcpdef{cr.sketch-siglip-m20.raw.size.ci}{22.35\,[20.88, 23.20]}
\pcpdef{cr.sketch-siglip-m20.raw.size.mean}{22.35}
\pcpdef{cr.sketch-siglip-m20.raw.size.sd}{0.665}
\pcpdef{cr.sketch-siglip-m20.refresh-minus-pfx.ci}{-3.23\,[-3.66, -2.93]}
\pcpdef{cr.sketch-siglip-m20.refresh-minus-pfx.max}{-2.93}
\pcpdef{cr.sketch-siglip-m20.refresh-minus-pfx.mean}{-3.23}
\pcpdef{cr.sketch-siglip-m20.refresh-minus-pfx.min}{-3.74}
\pcpdef{cr.sketch-siglip-m20.refresh-minus-pfx.p025}{-3.66}
\pcpdef{cr.sketch-siglip-m20.refresh-minus-pfx.p975}{-2.93}
\pcpdef{cr.sketch-siglip-m20.refresh-minus-pfx.sd}{0.224}
\pcpdef{cr.sketch-siglip-m20.refresh.coverage.macro.ci}{0.9026\,[0.9002, 0.9045]}
\pcpdef{cr.sketch-siglip-m20.refresh.coverage.macro.mean}{0.9026}
\pcpdef{cr.sketch-siglip-m20.refresh.coverage.macro.min}{0.9000}
\pcpdef{cr.sketch-siglip-m20.refresh.coverage.macro.p025}{0.9002}
\pcpdef{cr.sketch-siglip-m20.refresh.coverage.macro.p975}{0.9045}
\pcpdef{cr.sketch-siglip-m20.refresh.coverage.macro.sd}{0.00122}
\pcpdef{cr.sketch-siglip-m20.refresh.reach.ci}{77.3\,[75.5, 78.7]}
\pcpdef{cr.sketch-siglip-m20.refresh.reach.max}{78.7}
\pcpdef{cr.sketch-siglip-m20.refresh.reach.mean}{77.3}
\pcpdef{cr.sketch-siglip-m20.refresh.reach.min}{74.8}
\pcpdef{cr.sketch-siglip-m20.refresh.reach.p025}{75.5}
\pcpdef{cr.sketch-siglip-m20.refresh.reach.p975}{78.7}
\pcpdef{cr.sketch-siglip-m20.refresh.reach.sd}{0.91}
\pcpdef{cr.sketch-siglip-m20.refresh.size.ci}{8.81\,[8.33, 9.19]}
\pcpdef{cr.sketch-siglip-m20.refresh.size.mean}{8.81}
\pcpdef{cr.sketch-siglip-m20.refresh.size.sd}{0.256}
\pcpdef{cr.table.main.body}{%
Food-101 / SigLIP & 65.2\,[62.5, 66.8] & +1.25\,[+0.83, +1.60] & -1.43\,[-1.74, -1.07] & 0.9265 & 0.858 & 0.718 \\
ImageNet-Sketch / SigLIP & 70.5\,[69.2, 72.0] & +1.89\,[+1.61, +2.09] & -3.23\,[-3.66, -2.93] & 0.9030 & 0.782 & 0.478 \\}
\pcpdef{cr.table.supp.food-siglip-m20.body}{%
100 & 0 & 8100000 & \texttt{a30342b7} & 66.7 & 30.7 & 0.98 & +0.91 & -1.33 & 40.6 & 0.9306 & 0.864 & 0.759 & yes \\
100 & 1 & 8100001 & \texttt{c68ecd00} & 65.8 & 30.4 & 0.89 & +1.50 & -1.48 & 50.3 & 0.9287 & 0.872 & 0.643 & yes \\
100 & 2 & 8100002 & \texttt{06db6a91} & 61.5 & 31.6 & 0.70 & +0.85 & -1.06 & 44.5 & 0.9166 & 0.844 & 0.740 & yes \\
100 & 3 & 8100003 & \texttt{44c69d27} & 64.6 & 31.0 & 0.81 & +1.59 & -1.54 & 50.9 & 0.9200 & 0.844 & 0.716 & yes \\
100 & 4 & 8100004 & \texttt{3c3995ae} & 64.5 & 31.3 & 0.83 & +1.62 & -1.43 & 53.1 & 0.9269 & 0.852 & 0.759 & yes \\
100 & 5 & 8100005 & \texttt{42227310} & 66.6 & 30.8 & 0.93 & +1.46 & -1.39 & 51.3 & 0.9282 & 0.856 & 0.764 & yes \\
100 & 6 & 8100006 & \texttt{15e95522} & 66.5 & 29.7 & 0.91 & +1.58 & -1.60 & 49.6 & 0.9276 & 0.868 & 0.718 & yes \\
100 & 7 & 8100007 & \texttt{7383da53} & 65.0 & 32.4 & 0.81 & +1.32 & -1.71 & 43.5 & 0.9160 & 0.843 & 0.680 & yes \\
100 & 8 & 8100008 & \texttt{4301919b} & 64.9 & 30.4 & 0.80 & +0.92 & -1.49 & 38.1 & 0.9322 & 0.876 & 0.783 & yes \\
100 & 9 & 8100009 & \texttt{64bd1879} & 64.4 & 32.3 & 0.79 & +1.49 & -1.43 & 51.2 & 0.9263 & 0.867 & 0.668 & yes \\
100 & 10 & 8100010 & \texttt{3f87bb5b} & 65.7 & 30.4 & 0.89 & +1.09 & -1.62 & 40.3 & 0.9271 & 0.848 & 0.792 & yes \\
100 & 11 & 8100011 & \texttt{e65a4a58} & 63.6 & 31.4 & 0.78 & +1.08 & -1.15 & 48.4 & 0.9300 & 0.852 & 0.688 & yes \\
100 & 12 & 8100012 & \texttt{89a27d00} & 66.8 & 30.0 & 0.92 & +1.47 & -1.61 & 47.7 & 0.9292 & 0.855 & 0.704 & yes \\
100 & 13 & 8100013 & \texttt{e86f33b2} & 66.7 & 31.3 & 0.95 & +1.59 & -1.77 & 47.2 & 0.9284 & 0.860 & 0.536 & yes \\
100 & 14 & 8100014 & \texttt{bbe93774} & 64.5 & 30.5 & 0.83 & +1.08 & -1.41 & 43.3 & 0.9232 & 0.856 & 0.672 & yes \\
100 & 15 & 8100015 & \texttt{8caa1028} & 66.6 & 30.9 & 0.88 & +1.03 & -1.68 & 37.9 & 0.9245 & 0.852 & 0.712 & yes \\
100 & 16 & 8100016 & \texttt{d7451439} & 65.7 & 30.6 & 0.89 & +1.11 & -1.18 & 48.6 & 0.9322 & 0.860 & 0.772 & yes \\
100 & 17 & 8100017 & \texttt{bd967a36} & 66.0 & 31.9 & 0.87 & +1.14 & -1.35 & 45.8 & 0.9238 & 0.863 & 0.744 & yes \\
100 & 18 & 8100018 & \texttt{df39a6c7} & 64.4 & 32.4 & 0.83 & +1.33 & -1.32 & 50.2 & 0.9245 & 0.852 & 0.748 & yes \\
100 & 19 & 8100019 & \texttt{ce5a652e} & 64.8 & 32.3 & 0.83 & +0.80 & -1.08 & 42.8 & 0.9346 & 0.880 & 0.768 & yes \\}
\pcpdef{cr.table.supp.sketch-siglip-m20.body}{%
100 & 0 & 8100000 & \texttt{7010435e} & 70.6 & 41.1 & 0.09 & +1.90 & -3.06 & 38.3 & 0.9032 & 0.783 & 0.455 & yes \\
100 & 1 & 8100001 & \texttt{cc37e8e5} & 71.1 & 41.3 & 0.10 & +1.80 & -3.32 & 35.2 & 0.9017 & 0.769 & 0.500 & yes \\
100 & 2 & 8100002 & \texttt{0e82a4aa} & 70.5 & 43.2 & 0.10 & +1.88 & -3.32 & 36.1 & 0.9023 & 0.783 & 0.455 & yes \\
100 & 3 & 8100003 & \texttt{dfb5aff8} & 70.2 & 41.8 & 0.09 & +1.93 & -3.23 & 37.4 & 0.9023 & 0.772 & 0.520 & yes \\
100 & 4 & 8100004 & \texttt{139be79f} & 69.5 & 43.1 & 0.09 & +1.80 & -3.06 & 37.1 & 0.9035 & 0.791 & 0.423 & yes \\
100 & 5 & 8100005 & \texttt{ecd7c275} & 70.3 & 42.6 & 0.09 & +1.87 & -3.11 & 37.5 & 0.9011 & 0.769 & 0.520 & yes \\
100 & 6 & 8100006 & \texttt{d6dc8188} & 72.0 & 42.1 & 0.10 & +1.47 & -3.58 & 29.1 & 0.9031 & 0.782 & 0.520 & yes \\
100 & 7 & 8100007 & \texttt{7a8063b7} & 70.6 & 41.3 & 0.09 & +1.79 & -3.05 & 37.0 & 0.9017 & 0.782 & 0.455 & yes \\
100 & 8 & 8100008 & \texttt{6483576f} & 70.2 & 43.1 & 0.09 & +1.83 & -3.30 & 35.7 & 0.9034 & 0.791 & 0.455 & yes \\
100 & 9 & 8100009 & \texttt{bb05998b} & 70.5 & 41.2 & 0.09 & +2.02 & -3.00 & 40.1 & 0.9027 & 0.769 & 0.462 & yes \\
100 & 10 & 8100010 & \texttt{5ba637a6} & 70.8 & 41.7 & 0.10 & +1.99 & -3.22 & 38.2 & 0.9030 & 0.769 & 0.455 & yes \\
100 & 11 & 8100011 & \texttt{864d7380} & 69.0 & 40.7 & 0.09 & +1.75 & -2.96 & 37.2 & 0.9002 & 0.773 & 0.455 & yes \\
100 & 12 & 8100012 & \texttt{a8bc6293} & 70.6 & 43.3 & 0.10 & +2.00 & -3.37 & 37.2 & 0.9065 & 0.800 & 0.440 & yes \\
100 & 13 & 8100013 & \texttt{9e816df1} & 70.1 & 41.8 & 0.09 & +1.90 & -3.18 & 37.4 & 0.9040 & 0.792 & 0.520 & yes \\
100 & 14 & 8100014 & \texttt{7cd505e6} & 69.4 & 43.1 & 0.09 & +1.83 & -2.93 & 38.5 & 0.9043 & 0.783 & 0.520 & yes \\
100 & 15 & 8100015 & \texttt{76b57448} & 72.0 & 42.4 & 0.10 & +1.89 & -3.74 & 33.5 & 0.9016 & 0.769 & 0.462 & yes \\
100 & 16 & 8100016 & \texttt{d9b9937f} & 70.2 & 40.4 & 0.09 & +2.14 & -2.93 & 42.2 & 0.9030 & 0.783 & 0.423 & yes \\
100 & 17 & 8100017 & \texttt{8278f850} & 71.6 & 41.8 & 0.10 & +1.98 & -3.55 & 35.8 & 0.9040 & 0.800 & 0.520 & yes \\
100 & 18 & 8100018 & \texttt{0de91a0a} & 70.6 & 43.3 & 0.10 & +2.02 & -3.37 & 37.5 & 0.9041 & 0.792 & 0.500 & yes \\
100 & 19 & 8100019 & \texttt{432b33cd} & 70.5 & 41.7 & 0.10 & +2.04 & -3.29 & 38.3 & 0.9041 & 0.800 & 0.500 & yes \\}
\pcpdef{cub.clip.mean.reach.future}{69.2}
\pcpdef{cub.clip.mean.reach.hist}{70.2}
\pcpdef{cub.clip.sum.reach.future}{91.5}
\pcpdef{cub.clip.sum.reach.hist}{90.9}
\pcpdef{cub.siglip.mean.reach.future}{62.0}
\pcpdef{cub.siglip.mean.reach.hist}{63.5}
\pcpdef{cub.siglip.sum.reach.future}{76.9}
\pcpdef{cub.siglip.sum.reach.hist}{76.7}
\pcpdef{fig.pareto.rawrel.range}{1.26\pcpdash{}2.05}
\pcpdef{gd.cells}{120}
\pcpdef{gd.clip.all.ci}{[+0.151,\,+0.306]}
\pcpdef{gd.clip.all.closed}{37.9}
\pcpdef{gd.clip.all.delta}{+0.228}
\pcpdef{gd.clip.all.rawci}{[+0.279,\,+1.008]}
\pcpdef{gd.clip.all.rawdelta}{+0.643}
\pcpdef{gd.clip.all.rawdelta.abs}{0.643}
\pcpdef{gd.clip.all.rawwins}{6/6}
\pcpdef{gd.clip.all.wins}{6/6}
\pcpdef{gd.clip.frozen.bitchurn}{2.46}
\pcpdef{gd.clip.frozen.macro}{0.9137}
\pcpdef{gd.clip.frozen.median}{1.205}
\pcpdef{gd.clip.frozen.novel}{1.242}
\pcpdef{gd.clip.frozen.reach}{37.96}
\pcpdef{gd.clip.frozen.rev}{40.14}
\pcpdef{gd.clip.frozen.size}{1.286}
\pcpdef{gd.clip.gain.range}{0.111\pcpdash{}0.330}
\pcpdef{gd.clip.novel.ci}{[+0.345,\,+0.649]}
\pcpdef{gd.clip.novel.closed}{38.5}
\pcpdef{gd.clip.novel.delta}{+0.497}
\pcpdef{gd.clip.novel.rawci}{[-0.150,\,+0.541]}
\pcpdef{gd.clip.novel.rawdelta}{+0.195}
\pcpdef{gd.clip.novel.rawdelta.abs}{0.195}
\pcpdef{gd.clip.novel.rawwins}{4/6}
\pcpdef{gd.clip.novel.wins}{6/6}
\pcpdef{gd.clip.novelblock.closed}{100.0}
\pcpdef{gd.clip.pre.macro}{0.8991}
\pcpdef{gd.clip.pre.median}{1.693}
\pcpdef{gd.clip.pre.novel}{2.537}
\pcpdef{gd.clip.pre.novelcand}{0.835}
\pcpdef{gd.clip.pre.size}{1.871}
\pcpdef{gd.clip.prefix.empty}{2.4}
\pcpdef{gd.clip.prefix.macro}{0.9006}
\pcpdef{gd.clip.prefix.median}{1.575}
\pcpdef{gd.clip.prefix.novel}{2.040}
\pcpdef{gd.clip.prefix.novelcand}{0.607}
\pcpdef{gd.clip.prefix.p10}{0.805}
\pcpdef{gd.clip.prefix.size}{1.643}
\pcpdef{gd.clip.prefix.worstcov}{0.694}
\pcpdef{gd.clip.raw.macro}{0.8971}
\pcpdef{gd.clip.raw.median}{2.192}
\pcpdef{gd.clip.raw.novel}{2.235}
\pcpdef{gd.clip.raw.size}{2.286}
\pcpdef{gd.clip.refresh.bitchurn}{2.49}
\pcpdef{gd.clip.refresh.macro}{0.9008}
\pcpdef{gd.clip.refresh.median}{1.195}
\pcpdef{gd.clip.refresh.novel}{1.248}
\pcpdef{gd.clip.refresh.novelcand}{0.607}
\pcpdef{gd.clip.refresh.reach}{38.12}
\pcpdef{gd.clip.refresh.rev}{41.02}
\pcpdef{gd.clip.refresh.size}{1.268}
\pcpdef{gd.closed.novel.range}{38.5\pcpdash{}39.1}
\pcpdef{gd.closed.range}{37.9\pcpdash{}39.1}
\pcpdef{gd.gate.entries}{352}
\pcpdef{gd.gate.exempt}{8}
\pcpdef{gd.gate.maxdev}{1.22\times10^{-15}}
\pcpdef{gd.onboarding.maxdelta}{1.11\times10^{-16}}
\pcpdef{gd.regionlist}{Africa, EastAsia, SouthEastAsia, Americas, WestAsia, Europe}
\pcpdef{gd.regions}{6}
\pcpdef{gd.seeds}{10}
\pcpdef{gd.siglip.all.ci}{[+0.116,\,+0.167]}
\pcpdef{gd.siglip.all.closed}{39.1}
\pcpdef{gd.siglip.all.delta}{+0.141}
\pcpdef{gd.siglip.all.rawci}{[+0.038,\,+0.188]}
\pcpdef{gd.siglip.all.rawdelta}{+0.113}
\pcpdef{gd.siglip.all.rawdelta.abs}{0.113}
\pcpdef{gd.siglip.all.rawwins}{6/6}
\pcpdef{gd.siglip.all.wins}{6/6}
\pcpdef{gd.siglip.frozen.bitchurn}{1.51}
\pcpdef{gd.siglip.frozen.macro}{0.9269}
\pcpdef{gd.siglip.frozen.median}{1.018}
\pcpdef{gd.siglip.frozen.novel}{1.051}
\pcpdef{gd.siglip.frozen.reach}{26.69}
\pcpdef{gd.siglip.frozen.rev}{31.82}
\pcpdef{gd.siglip.frozen.size}{1.081}
\pcpdef{gd.siglip.gain.range}{0.095\pcpdash{}0.159}
\pcpdef{gd.siglip.novel.ci}{[+0.238,\,+0.358]}
\pcpdef{gd.siglip.novel.closed}{39.1}
\pcpdef{gd.siglip.novel.delta}{+0.298}
\pcpdef{gd.siglip.novel.rawci}{[-0.213,\,-0.048]}
\pcpdef{gd.siglip.novel.rawdelta}{-0.131}
\pcpdef{gd.siglip.novel.rawdelta.abs}{0.131}
\pcpdef{gd.siglip.novel.rawwins}{0/6}
\pcpdef{gd.siglip.novel.wins}{6/6}
\pcpdef{gd.siglip.novelblock.closed}{100.0}
\pcpdef{gd.siglip.pre.macro}{0.9042}
\pcpdef{gd.siglip.pre.median}{1.270}
\pcpdef{gd.siglip.pre.novel}{1.823}
\pcpdef{gd.siglip.pre.novelcand}{0.664}
\pcpdef{gd.siglip.pre.size}{1.425}
\pcpdef{gd.siglip.prefix.empty}{4.6}
\pcpdef{gd.siglip.prefix.macro}{0.9056}
\pcpdef{gd.siglip.prefix.median}{1.165}
\pcpdef{gd.siglip.prefix.novel}{1.525}
\pcpdef{gd.siglip.prefix.novelcand}{0.523}
\pcpdef{gd.siglip.prefix.p10}{0.821}
\pcpdef{gd.siglip.prefix.size}{1.284}
\pcpdef{gd.siglip.prefix.worstcov}{0.730}
\pcpdef{gd.siglip.raw.macro}{0.8966}
\pcpdef{gd.siglip.raw.median}{1.219}
\pcpdef{gd.siglip.raw.novel}{1.395}
\pcpdef{gd.siglip.raw.size}{1.397}
\pcpdef{gd.siglip.refresh.bitchurn}{1.47}
\pcpdef{gd.siglip.refresh.macro}{0.9071}
\pcpdef{gd.siglip.refresh.median}{0.990}
\pcpdef{gd.siglip.refresh.novel}{1.061}
\pcpdef{gd.siglip.refresh.novelcand}{0.523}
\pcpdef{gd.siglip.refresh.reach}{25.82}
\pcpdef{gd.siglip.refresh.rev}{32.30}
\pcpdef{gd.siglip.refresh.size}{1.064}
\pcpdef{gw.available}{yes}
\pcpdef{gw.blas}{accelerate}
\pcpdef{gw.cells}{8}
\pcpdef{gw.control.live}{yes}
\pcpdef{gw.control.maxulp}{98{,}304}
\pcpdef{gw.control.minulp}{26}
\pcpdef{gw.coordinates}{83{,}293{,}692}
\pcpdef{gw.flips}{0}
\pcpdef{gw.grids}{4}
\pcpdef{gw.grids.word}{four}
\pcpdef{gw.kmax}{930}
\pcpdef{gw.kmin}{101}
\pcpdef{gw.maxabs}{0.0}
\pcpdef{gw.maxulp}{0}
\pcpdef{gw.membership.cells}{47{,}727{,}372}
\pcpdef{gw.numpy}{2.4.6}
\pcpdef{gw.score.maxabs}{0.0}
\pcpdef{or.carsclip.pdmc.corr}{0.7796}
\pcpdef{or.carsclip.pdmc.rms}{0.009911}
\pcpdef{or.carsclip.pre.corr}{0.9079}
\pcpdef{or.carsclip.pre.rms}{0.005783}
\pcpdef{or.carssiglip.pdmc.corr}{0.6302}
\pcpdef{or.carssiglip.pdmc.rms}{0.015872}
\pcpdef{or.carssiglip.pre.corr}{0.6691}
\pcpdef{or.carssiglip.pre.rms}{0.014556}
\pcpdef{or.food.pdmc.corr}{0.4105}
\pcpdef{or.food.pdmc.rms}{0.011305}
\pcpdef{or.food.pre.corr}{0.5351}
\pcpdef{or.food.pre.rms}{0.010597}
\pcpdef{or.seeds}{5}
\pcpdef{or.synthetic}{20}
\pcpdef{ot.apssize.range}{9.01\pcpdash{}27.72}
\pcpdef{ot.carsclip.apssize}{27.72}
\pcpdef{ot.carsclip.cohortreach}{21.7}
\pcpdef{ot.carsclip.gap}{6.68}
\pcpdef{ot.carsclip.insrev}{29.4}
\pcpdef{ot.carsclip.lacmarg}{0.9049}
\pcpdef{ot.carsclip.lacsize}{25.55}
\pcpdef{ot.carsclip.macro}{0.9018}
\pcpdef{ot.carsclip.noadaptsize}{30.73}
\pcpdef{ot.carsclip.size}{27.775}
\pcpdef{ot.carssiglip.apssize}{16.56}
\pcpdef{ot.carssiglip.cohortreach}{22.3}
\pcpdef{ot.carssiglip.gap}{8.30}
\pcpdef{ot.carssiglip.insrev}{35.3}
\pcpdef{ot.carssiglip.lacmarg}{0.8974}
\pcpdef{ot.carssiglip.lacsize}{13.21}
\pcpdef{ot.carssiglip.macro}{0.8971}
\pcpdef{ot.carssiglip.noadaptsize}{15.73}
\pcpdef{ot.carssiglip.size}{14.753}
\pcpdef{ot.cohortreach.range}{13.3\pcpdash{}22.3}
\pcpdef{ot.food.apssize}{9.54}
\pcpdef{ot.food.cohortreach}{13.3}
\pcpdef{ot.food.gap}{4.72}
\pcpdef{ot.food.insrev}{49.0}
\pcpdef{ot.food.lacmarg}{0.9305}
\pcpdef{ot.food.lacsize}{7.82}
\pcpdef{ot.food.macro}{0.9277}
\pcpdef{ot.food.noadaptsize}{9.55}
\pcpdef{ot.food.size}{10.449}
\pcpdef{ot.food40.apssize}{9.01}
\pcpdef{ot.food40.noadaptsize}{9.03}
\pcpdef{ot.food80.apssize}{9.23}
\pcpdef{ot.food80.noadaptsize}{9.35}
\pcpdef{ot.gap.range}{4.7\pcpdash{}8.3}
\pcpdef{ot.gap.range5}{4.6\pcpdash{}8.3}
\pcpdef{ot.insrev.range}{29\pcpdash{}49}
\pcpdef{ot.noadaptsize.range}{9.03\pcpdash{}30.73}
\pcpdef{ps.closure.range}{45.6\pcpdash{}72.4}
\pcpdef{ps.gate.entries}{675}
\pcpdef{ps.noisefloor.cosine}{0.9999998}
\pcpdef{ps.noisefloor.delta}{5.96\times10^{-7}}
\pcpdef{ps.novelclosure.maxspread}{2.8}
\pcpdef{ps.novelclosure.range}{45.6\pcpdash{}53.9}
\pcpdef{ps.recoded.maxdelta}{1.2\times10^{-4}}
\pcpdef{ps.temp.cells}{3/3}
\pcpdef{ps.template.cells}{12/12}
\pcpdef{rb.available}{yes}
\pcpdef{rb.batch.real}{24{,}003}
\pcpdef{rb.batch.syn}{1{,}024}
\pcpdef{rb.batch1.blocked}{18.5\pcpdash{}22.0}
\pcpdef{rb.batch1.blockedoverfloor}{2.71\pcpdash{}3.62}
\pcpdef{rb.batch1.direct}{25.5\pcpdash{}32.8}
\pcpdef{rb.batch1.seq}{7.2\pcpdash{}9.8}
\pcpdef{rb.breakeven}{2}
\pcpdef{rb.breakeven.allwin}{yes}
\pcpdef{rb.breakeven.novalidate}{2}
\pcpdef{rb.breakeven.shipped}{2}
\pcpdef{rb.cells}{198}
\pcpdef{rb.correct.blockeddirect}{18/198}
\pcpdef{rb.correct.blockedseq}{0/198}
\pcpdef{rb.correct.directseq}{0/198}
\pcpdef{rb.dev.blockeddirect}{1.33\times10^{-15}}
\pcpdef{rb.dev.blockedseq}{1.16\times10^{-13}}
\pcpdef{rb.dev.directseq}{1.16\times10^{-13}}
\pcpdef{rb.dev.raw}{1.16\times10^{-13}}
\pcpdef{rb.devq.blockeddirect}{1.56\times10^{-13}}
\pcpdef{rb.devq.blockedseq}{1.36\times10^{-11}}
\pcpdef{rb.devq.directseq}{1.36\times10^{-11}}
\pcpdef{rb.e2e.T2.overhead}{4.7}
\pcpdef{rb.e2e.T6.overhead}{18.3}
\pcpdef{rb.kernel.sha}{b9f44a3915b7}
\pcpdef{rb.lambda}{8}
\pcpdef{rb.maxrel}{2.39\times10^{-12}}
\pcpdef{rb.mem.blocked}{122}
\pcpdef{rb.mem.blocked.T10}{254}
\pcpdef{rb.mem.direct}{14{,}645}
\pcpdef{rb.mem.directover}{120}
\pcpdef{rb.mem.floor}{2{,}442}
\pcpdef{rb.mem.floorover}{20}
\pcpdef{rb.planner.agrees}{yes}
\pcpdef{rb.pre.T2.K1}{465}
\pcpdef{rb.pre.T2.blockedover}{2.16}
\pcpdef{rb.pre.T2.control}{51.15}
\pcpdef{rb.pre.T2.controldelta}{2.4}
\pcpdef{rb.pre.T2.floor}{26.98}
\pcpdef{rb.pre.T6.K1}{233}
\pcpdef{rb.pre.T6.blockedover}{5.57}
\pcpdef{rb.pre.T6.control}{51.15}
\pcpdef{rb.pre.T6.controldelta}{2.4}
\pcpdef{rb.pre.T6.floor}{13.45}
\pcpdef{rb.pre.mem}{24}
\pcpdef{rb.pre.mem.K1}{1{,}000}
\pcpdef{rb.pre.mem.control}{2{,}442}
\pcpdef{rb.pre.syn.K1}{200}
\pcpdef{rb.pre.syn.floor}{0.50}
\pcpdef{rb.real.blockedoverfloor}{1.07\pcpdash{}1.96}
\pcpdef{rb.real.blockedovermatmul}{0.73\pcpdash{}1.20}
\pcpdef{rb.real.directoverblocked}{1.36\pcpdash{}3.66}
\pcpdef{rb.real.seqoverblocked}{5.3\pcpdash{}9.1}
\pcpdef{rb.sketch.T2.blocked}{58.18}
\pcpdef{rb.sketch.T2.direct}{85.47}
\pcpdef{rb.sketch.T2.floor}{52.38}
\pcpdef{rb.sketch.T2.matmul}{70.68}
\pcpdef{rb.sketch.T2.seq}{494.20}
\pcpdef{rb.sketch.T6.blocked}{74.95}
\pcpdef{rb.sketch.T6.direct}{226.08}
\pcpdef{rb.sketch.T6.floor}{52.38}
\pcpdef{rb.sketch.T6.matmul}{70.68}
\pcpdef{rb.sketch.T6.seq}{494.40}
\pcpdef{rb.skew.K}{10{,}000}
\pcpdef{rb.skew.T}{21}
\pcpdef{rb.skew.blockedoverfloor}{1.08\pcpdash{}1.09}
\pcpdef{rb.skew.ratio}{4.96\pcpdash{}17.84}
\pcpdef{rb.skew.sumratio}{18.9}
\pcpdef{rb.stack.bytes}{3.75}
\pcpdef{rb.stack.time}{3.0\pcpdash{}3.4}
\pcpdef{rb.syn.blockedoverfloor}{1.23\pcpdash{}1.48}
\pcpdef{rb.syn.directoverfloor}{118\pcpdash{}548}
\pcpdef{rb.trunc.blocked}{514/514}
\pcpdef{rb.trunc.direct}{498/498}
\pcpdef{rb.trunc.directskipped}{16}
\pcpdef{rc.boundaries}{6{,}710}
\pcpdef{rc.cadences}{4}
\pcpdef{rc.carsclip.T2}{29.555}
\pcpdef{rc.carsclip.T2.random}{29.555}
\pcpdef{rc.carsclip.T20.random}{29.727}
\pcpdef{rc.carsclip.T6.random}{29.670}
\pcpdef{rc.carsclip.Tper-class.random}{29.755}
\pcpdef{rc.carsclip.advantage}{0.535}
\pcpdef{rc.carsclip.base.range}{15.362}
\pcpdef{rc.carsclip.basecandidate}{15.362}
\pcpdef{rc.carsclip.best}{29.755}
\pcpdef{rc.carsclip.consumed.best}{37.3}
\pcpdef{rc.carsclip.consumed.worst}{54.9}
\pcpdef{rc.carsclip.cov.range}{0.9017\pcpdash{}0.9025}
\pcpdef{rc.carsclip.far-base-first.absdelta}{1.012}
\pcpdef{rc.carsclip.far-base-first.bitdiff}{0.66}
\pcpdef{rc.carsclip.far-base-first.imgdiff}{65.5}
\pcpdef{rc.carsclip.far-base-first.maxdelta}{10.3}
\pcpdef{rc.carsclip.far-base-first.meandelta}{+0.220}
\pcpdef{rc.carsclip.far-base-first.p95delta}{3.14}
\pcpdef{rc.carsclip.far-base-first.perseed}{[+0.118,\;+0.241,\;+0.184,\;+0.341,\;+0.213]}
\pcpdef{rc.carsclip.far-base-first.symdiff}{1.29}
\pcpdef{rc.carsclip.frozen.churnT2}{1.06}
\pcpdef{rc.carsclip.frozen.churnfine}{1.57}
\pcpdef{rc.carsclip.frozen.reachT2}{67.3}
\pcpdef{rc.carsclip.frozen.reachfine}{81.1}
\pcpdef{rc.carsclip.high-confusion-first.absdelta}{1.207}
\pcpdef{rc.carsclip.high-confusion-first.bitdiff}{0.78}
\pcpdef{rc.carsclip.high-confusion-first.imgdiff}{65.8}
\pcpdef{rc.carsclip.high-confusion-first.maxdelta}{20.4}
\pcpdef{rc.carsclip.high-confusion-first.meandelta}{+0.275}
\pcpdef{rc.carsclip.high-confusion-first.p95delta}{4.15}
\pcpdef{rc.carsclip.high-confusion-first.perseed}{[-0.003,\;+0.470,\;+0.225,\;+0.353,\;+0.331]}
\pcpdef{rc.carsclip.high-confusion-first.symdiff}{1.52}
\pcpdef{rc.carsclip.near-base-first.absdelta}{1.241}
\pcpdef{rc.carsclip.near-base-first.bitdiff}{0.78}
\pcpdef{rc.carsclip.near-base-first.imgdiff}{65.4}
\pcpdef{rc.carsclip.near-base-first.maxdelta}{19.6}
\pcpdef{rc.carsclip.near-base-first.meandelta}{+0.294}
\pcpdef{rc.carsclip.near-base-first.p95delta}{4.42}
\pcpdef{rc.carsclip.near-base-first.perseed}{[+0.122,\;+0.475,\;+0.258,\;+0.264,\;+0.349]}
\pcpdef{rc.carsclip.near-base-first.symdiff}{1.53}
\pcpdef{rc.carsclip.novel.T2}{14.193}
\pcpdef{rc.carsclip.novel.worst}{14.486}
\pcpdef{rc.carsclip.nversions}{99}
\pcpdef{rc.carsclip.pre}{30.090}
\pcpdef{rc.carsclip.random.absdelta}{1.017}
\pcpdef{rc.carsclip.random.bitdiff}{0.66}
\pcpdef{rc.carsclip.random.imgdiff}{63.5}
\pcpdef{rc.carsclip.random.maxdelta}{16.1}
\pcpdef{rc.carsclip.random.meandelta}{+0.200}
\pcpdef{rc.carsclip.random.p95delta}{3.26}
\pcpdef{rc.carsclip.random.perseed}{[+0.381,\;+0.226,\;-0.003,\;+0.307,\;+0.088]}
\pcpdef{rc.carsclip.random.symdiff}{1.29}
\pcpdef{rc.carsclip.work.T2.blocked}{204}
\pcpdef{rc.carsclip.work.T2.direct}{294}
\pcpdef{rc.carsclip.work.fine.blocked}{980}
\pcpdef{rc.carsclip.work.fine.direct}{14{,}553}
\pcpdef{rc.carsclip.work.fine.ratio}{14.9}
\pcpdef{rc.carsclip.worst}{29.848}
\pcpdef{rc.carsclip.worstorder}{near-base-first}
\pcpdef{rc.carssiglip.T2}{16.506}
\pcpdef{rc.carssiglip.T2.random}{16.506}
\pcpdef{rc.carssiglip.T20.random}{16.902}
\pcpdef{rc.carssiglip.T6.random}{16.847}
\pcpdef{rc.carssiglip.Tper-class.random}{16.918}
\pcpdef{rc.carssiglip.advantage}{1.092}
\pcpdef{rc.carssiglip.base.range}{9.107}
\pcpdef{rc.carssiglip.basecandidate}{9.107}
\pcpdef{rc.carssiglip.best}{16.852}
\pcpdef{rc.carssiglip.consumed.best}{31.7}
\pcpdef{rc.carssiglip.consumed.worst}{64.0}
\pcpdef{rc.carssiglip.cov.range}{0.8975\pcpdash{}0.8989}
\pcpdef{rc.carssiglip.far-base-first.absdelta}{0.843}
\pcpdef{rc.carssiglip.far-base-first.bitdiff}{0.52}
\pcpdef{rc.carssiglip.far-base-first.imgdiff}{49.6}
\pcpdef{rc.carssiglip.far-base-first.maxdelta}{12.6}
\pcpdef{rc.carssiglip.far-base-first.meandelta}{+0.346}
\pcpdef{rc.carssiglip.far-base-first.p95delta}{3.18}
\pcpdef{rc.carssiglip.far-base-first.perseed}{[+0.285,\;+0.528,\;+0.332,\;+0.250,\;+0.335]}
\pcpdef{rc.carssiglip.far-base-first.symdiff}{1.02}
\pcpdef{rc.carssiglip.frozen.churnT2}{1.11}
\pcpdef{rc.carssiglip.frozen.churnfine}{1.60}
\pcpdef{rc.carssiglip.frozen.reachT2}{59.9}
\pcpdef{rc.carssiglip.frozen.reachfine}{68.6}
\pcpdef{rc.carssiglip.high-confusion-first.absdelta}{1.209}
\pcpdef{rc.carssiglip.high-confusion-first.bitdiff}{0.71}
\pcpdef{rc.carssiglip.high-confusion-first.imgdiff}{53.1}
\pcpdef{rc.carssiglip.high-confusion-first.maxdelta}{19.2}
\pcpdef{rc.carssiglip.high-confusion-first.meandelta}{+0.662}
\pcpdef{rc.carssiglip.high-confusion-first.p95delta}{5.20}
\pcpdef{rc.carssiglip.high-confusion-first.perseed}{[+0.528,\;+0.718,\;+0.786,\;+0.553,\;+0.727]}
\pcpdef{rc.carssiglip.high-confusion-first.symdiff}{1.39}
\pcpdef{rc.carssiglip.near-base-first.absdelta}{1.315}
\pcpdef{rc.carssiglip.near-base-first.bitdiff}{0.79}
\pcpdef{rc.carssiglip.near-base-first.imgdiff}{55.3}
\pcpdef{rc.carssiglip.near-base-first.maxdelta}{19.8}
\pcpdef{rc.carssiglip.near-base-first.meandelta}{+0.699}
\pcpdef{rc.carssiglip.near-base-first.p95delta}{5.55}
\pcpdef{rc.carssiglip.near-base-first.perseed}{[+0.739,\;+0.718,\;+0.731,\;+0.652,\;+0.652]}
\pcpdef{rc.carssiglip.near-base-first.symdiff}{1.54}
\pcpdef{rc.carssiglip.novel.T2}{7.399}
\pcpdef{rc.carssiglip.novel.worst}{8.098}
\pcpdef{rc.carssiglip.nversions}{99}
\pcpdef{rc.carssiglip.pre}{17.598}
\pcpdef{rc.carssiglip.random.absdelta}{0.975}
\pcpdef{rc.carssiglip.random.bitdiff}{0.59}
\pcpdef{rc.carssiglip.random.imgdiff}{50.5}
\pcpdef{rc.carssiglip.random.maxdelta}{16.4}
\pcpdef{rc.carssiglip.random.meandelta}{+0.412}
\pcpdef{rc.carssiglip.random.p95delta}{3.92}
\pcpdef{rc.carssiglip.random.perseed}{[+0.504,\;+0.517,\;+0.338,\;+0.395,\;+0.305]}
\pcpdef{rc.carssiglip.random.symdiff}{1.16}
\pcpdef{rc.carssiglip.work.T2.blocked}{204}
\pcpdef{rc.carssiglip.work.T2.direct}{294}
\pcpdef{rc.carssiglip.work.fine.blocked}{980}
\pcpdef{rc.carssiglip.work.fine.direct}{14{,}553}
\pcpdef{rc.carssiglip.work.fine.ratio}{14.9}
\pcpdef{rc.carssiglip.worst}{17.205}
\pcpdef{rc.carssiglip.worstorder}{near-base-first}
\pcpdef{rc.cells}{255}
\pcpdef{rc.consumed.best.range}{31.7\pcpdash{}40.5}
\pcpdef{rc.consumed.range}{31.7\pcpdash{}64.0}
\pcpdef{rc.consumed.worst.range}{53.9\pcpdash{}64.0}
\pcpdef{rc.cov.spread.max}{0.0015}
\pcpdef{rc.food.T2}{12.148}
\pcpdef{rc.food.T2.random}{12.148}
\pcpdef{rc.food.T20.random}{12.757}
\pcpdef{rc.food.T6.random}{12.651}
\pcpdef{rc.food.Tper-class.random}{12.788}
\pcpdef{rc.food.advantage}{1.578}
\pcpdef{rc.food.base.range}{6.745}
\pcpdef{rc.food.basecandidate}{6.745}
\pcpdef{rc.food.best}{12.788}
\pcpdef{rc.food.consumed.best}{40.5}
\pcpdef{rc.food.consumed.worst}{53.9}
\pcpdef{rc.food.cov.range}{0.9266\pcpdash{}0.9274}
\pcpdef{rc.food.far-base-first.absdelta}{1.373}
\pcpdef{rc.food.far-base-first.bitdiff}{1.59}
\pcpdef{rc.food.far-base-first.imgdiff}{64.8}
\pcpdef{rc.food.far-base-first.maxdelta}{15.4}
\pcpdef{rc.food.far-base-first.meandelta}{+0.658}
\pcpdef{rc.food.far-base-first.p95delta}{5.02}
\pcpdef{rc.food.far-base-first.perseed}{[+0.759,\;+0.361,\;+0.873,\;+0.755,\;+0.540]}
\pcpdef{rc.food.far-base-first.symdiff}{1.61}
\pcpdef{rc.food.frozen.churnT2}{2.52}
\pcpdef{rc.food.frozen.churnfine}{3.79}
\pcpdef{rc.food.frozen.reachT2}{64.8}
\pcpdef{rc.food.frozen.reachfine}{74.8}
\pcpdef{rc.food.high-confusion-first.absdelta}{1.400}
\pcpdef{rc.food.high-confusion-first.bitdiff}{1.54}
\pcpdef{rc.food.high-confusion-first.imgdiff}{59.9}
\pcpdef{rc.food.high-confusion-first.maxdelta}{16.7}
\pcpdef{rc.food.high-confusion-first.meandelta}{+0.851}
\pcpdef{rc.food.high-confusion-first.p95delta}{5.80}
\pcpdef{rc.food.high-confusion-first.perseed}{[+0.760,\;+0.427,\;+0.967,\;+1.192,\;+0.907]}
\pcpdef{rc.food.high-confusion-first.symdiff}{1.55}
\pcpdef{rc.food.near-base-first.absdelta}{1.298}
\pcpdef{rc.food.near-base-first.bitdiff}{1.43}
\pcpdef{rc.food.near-base-first.imgdiff}{57.9}
\pcpdef{rc.food.near-base-first.maxdelta}{16.7}
\pcpdef{rc.food.near-base-first.meandelta}{+0.783}
\pcpdef{rc.food.near-base-first.p95delta}{5.38}
\pcpdef{rc.food.near-base-first.perseed}{[+0.633,\;+0.466,\;+0.757,\;+1.188,\;+0.872]}
\pcpdef{rc.food.near-base-first.symdiff}{1.44}
\pcpdef{rc.food.novel.T2}{5.403}
\pcpdef{rc.food.novel.worst}{6.254}
\pcpdef{rc.food.nversions}{52}
\pcpdef{rc.food.pre}{13.727}
\pcpdef{rc.food.random.absdelta}{1.346}
\pcpdef{rc.food.random.bitdiff}{1.55}
\pcpdef{rc.food.random.imgdiff}{61.8}
\pcpdef{rc.food.random.maxdelta}{18.1}
\pcpdef{rc.food.random.meandelta}{+0.640}
\pcpdef{rc.food.random.p95delta}{5.26}
\pcpdef{rc.food.random.perseed}{[+0.657,\;+0.222,\;+0.808,\;+0.899,\;+0.613]}
\pcpdef{rc.food.random.symdiff}{1.56}
\pcpdef{rc.food.work.T2.blocked}{109}
\pcpdef{rc.food.work.T2.direct}{151}
\pcpdef{rc.food.work.fine.blocked}{509}
\pcpdef{rc.food.work.fine.direct}{3{,}926}
\pcpdef{rc.food.work.fine.ratio}{7.7}
\pcpdef{rc.food.worst}{12.999}
\pcpdef{rc.food.worstorder}{high-confusion-first}
\pcpdef{rc.gate.entries}{435}
\pcpdef{rc.imgdiff.range}{49.6\pcpdash{}65.8}
\pcpdef{rc.maxdelta.max}{20}
\pcpdef{rc.meandelta.range}{0.20\pcpdash{}0.85}
\pcpdef{rc.onboarding.maxdelta}{8.33\times10^{-16}}
\pcpdef{rc.orders}{5}
\pcpdef{rc.p95delta.range}{3.1\pcpdash{}5.8}
\pcpdef{rc.schedules}{17}
\pcpdef{rc.signconsistent.all}{58/60}
\pcpdef{rc.signconsistent.fine}{10/12}
\pcpdef{rc.symdiff.range}{1.0\pcpdash{}1.6}
\pcpdef{rc3p.food.tcalpha}{-0.1012}
\pcpdef{rc3p.food40.tcalpha}{-0.0423}
\pcpdef{rc3p.food80.tcalpha}{-0.0006}
\pcpdef{rc3p.tgap}{0.9}
\pcpdef{sc.apscf.identity}{15/15}
\pcpdef{sc.apsdet.add}{1.31\pcpdash{}4.12}
\pcpdef{sc.apsdet.flip}{3.14\pcpdash{}12.58}
\pcpdef{sc.apsdet.reach}{81.2\pcpdash{}88.0}
\pcpdef{sc.apsdet.rev}{7.7\pcpdash{}23.3}
\pcpdef{sc.apsfrozen.add}{0.57\pcpdash{}1.29}
\pcpdef{sc.apsfrozen.flip}{2.70\pcpdash{}7.63}
\pcpdef{sc.apsfrozen.reach}{63.8\pcpdash{}77.3}
\pcpdef{sc.apsfrozen.rev}{9.1\pcpdash{}35.3}
\pcpdef{sc.apsrefresh.add}{1.02\pcpdash{}2.62}
\pcpdef{sc.apsrefresh.flip}{3.43\pcpdash{}8.88}
\pcpdef{sc.apsrefresh.reach}{69.7\pcpdash{}85.4}
\pcpdef{sc.apsrefresh.rev}{13.4\pcpdash{}36.2}
\pcpdef{sc.apsrenorm.add}{79.02\pcpdash{}92.94}
\pcpdef{sc.apsrenorm.flip}{65.70\pcpdash{}75.71}
\pcpdef{sc.apsrenorm.reach}{97.6\pcpdash{}99.9}
\pcpdef{sc.apsrenorm.rev}{0.0}
\pcpdef{sc.apstail.add}{0.00}
\pcpdef{sc.apstail.flip}{10.87\pcpdash{}16.77}
\pcpdef{sc.apstail.reach}{78.1\pcpdash{}95.4}
\pcpdef{sc.apstail.rev}{83.8\pcpdash{}89.7}
\pcpdef{sc.gate.comparisons}{45}
\pcpdef{sc.gate.maxdiff}{0.0}
\pcpdef{sc.gate.statistics}{930}
\pcpdef{sc.grids}{3}
\pcpdef{sc.hpsmeanfrozen.add}{0.43\pcpdash{}0.93}
\pcpdef{sc.hpsmeanfrozen.flip}{2.11\pcpdash{}5.08}
\pcpdef{sc.hpsmeanfrozen.reach}{59.9\pcpdash{}67.3}
\pcpdef{sc.hpsmeanfrozen.rev}{8.5\pcpdash{}32.5}
\pcpdef{sc.hpsmeanrefresh.add}{0.72\pcpdash{}2.37}
\pcpdef{sc.hpsmeanrefresh.flip}{2.50\pcpdash{}6.34}
\pcpdef{sc.hpsmeanrefresh.reach}{67.7\pcpdash{}79.3}
\pcpdef{sc.hpsmeanrefresh.rev}{10.5\pcpdash{}31.8}
\pcpdef{sc.lacbits.max}{1{,}261{,}200}
\pcpdef{sc.lacbits.min}{784{,}686}
\pcpdef{sc.lacfrozen.add}{0.00}
\pcpdef{sc.lacfrozen.flip}{3.39\pcpdash{}6.72}
\pcpdef{sc.lacfrozen.reach}{62.7\pcpdash{}85.1}
\pcpdef{sc.lacfrozen.rev}{34.6\pcpdash{}50.1}
\pcpdef{sc.lacidentity}{15/15}
\pcpdef{sc.rapsfrozen.add}{0.06\pcpdash{}0.23}
\pcpdef{sc.rapsfrozen.flip}{10.22\pcpdash{}12.10}
\pcpdef{sc.rapsfrozen.reach}{96.9\pcpdash{}100.0}
\pcpdef{sc.rapsfrozen.rev}{45.7\pcpdash{}51.9}
\pcpdef{sc.rapslam.add}{0.03\pcpdash{}0.11}
\pcpdef{sc.rapslam.flip}{11.09\pcpdash{}14.77}
\pcpdef{sc.rapslam.reach}{99.3\pcpdash{}100.0}
\pcpdef{sc.rapslam.rev}{51.7\pcpdash{}55.6}
\pcpdef{sc.rapsrefresh.add}{1.79\pcpdash{}2.49}
\pcpdef{sc.rapsrefresh.flip}{3.06\pcpdash{}5.92}
\pcpdef{sc.rapsrefresh.reach}{90.8\pcpdash{}96.0}
\pcpdef{sc.rapsrefresh.rev}{6.7\pcpdash{}20.6}
\pcpdef{sc.refit.reordered.range}{98\pcpdash{}100}
\pcpdef{sg.cars.K}{196}
\pcpdef{sg.cars.allclosure.frozen}{53.6}
\pcpdef{sg.cars.allclosure.refresh}{56.5}
\pcpdef{sg.cars.closure.K108}{44.7}
\pcpdef{sg.cars.closure.K137}{49.7}
\pcpdef{sg.cars.closure.K167}{52.1}
\pcpdef{sg.cars.closure.K196}{56.5}
\pcpdef{sg.cars.closure.K78}{36.7}
\pcpdef{sg.cars.closure.first}{36.7}
\pcpdef{sg.cars.closure.last}{56.5}
\pcpdef{sg.cars.frozen.additions}{4{,}998}
\pcpdef{sg.cars.frozen.churnfinal}{2.44}
\pcpdef{sg.cars.frozen.flipexcess}{11{,}064}
\pcpdef{sg.cars.frozen.macrofinal}{0.8967}
\pcpdef{sg.cars.frozen.netchanges}{32{,}568}
\pcpdef{sg.cars.frozen.pathflips}{43{,}633}
\pcpdef{sg.cars.frozen.reachfinal}{81.2}
\pcpdef{sg.cars.frozen.reachfirst}{45.6}
\pcpdef{sg.cars.frozen.revocations}{27{,}570}
\pcpdef{sg.cars.frozen.sizefinal}{14.933}
\pcpdef{sg.cars.frozen.worstfinal}{0.4642}
\pcpdef{sg.cars.kernel.bitschanged}{0}
\pcpdef{sg.cars.kernel.maxdelta}{4.30\times10^{-16}}
\pcpdef{sg.cars.newclosure.frozen}{35.4}
\pcpdef{sg.cars.newclosure.refresh}{36.9}
\pcpdef{sg.cars.onboarding.maxdelta}{1.11\times10^{-16}}
\pcpdef{sg.cars.pinned.churnfinal}{2.76}
\pcpdef{sg.cars.pre.macrofinal}{0.9000}
\pcpdef{sg.cars.pre.sizefinal}{21.014}
\pcpdef{sg.cars.pre.worstfinal}{0.6360}
\pcpdef{sg.cars.prefix.macrofinal}{0.9002}
\pcpdef{sg.cars.prefix.sizefinal}{17.753}
\pcpdef{sg.cars.prefix.worstfinal}{0.6301}
\pcpdef{sg.cars.prefixes}{49, 78, 108, 137, 167, 196}
\pcpdef{sg.cars.refitfinal}{167}
\pcpdef{sg.cars.refresh.additions}{10{,}016}
\pcpdef{sg.cars.refresh.churnfinal}{3.00}
\pcpdef{sg.cars.refresh.macrofinal}{0.8986}
\pcpdef{sg.cars.refresh.reachfinal}{85.3}
\pcpdef{sg.cars.refresh.reachfirst}{56.5}
\pcpdef{sg.cars.refresh.revocations}{30{,}064}
\pcpdef{sg.cars.refresh.sizefinal}{15.242}
\pcpdef{sg.cars.refresh.worstfinal}{0.6258}
\pcpdef{sg.cars.rescoredfinal}{3{,}340}
\pcpdef{sg.cells}{60}
\pcpdef{sg.food.K}{101}
\pcpdef{sg.food.allclosure.frozen}{48.6}
\pcpdef{sg.food.allclosure.refresh}{51.6}
\pcpdef{sg.food.closure.K101}{51.6}
\pcpdef{sg.food.closure.K40}{43.2}
\pcpdef{sg.food.closure.K56}{47.8}
\pcpdef{sg.food.closure.K71}{49.5}
\pcpdef{sg.food.closure.K86}{51.2}
\pcpdef{sg.food.closure.first}{43.2}
\pcpdef{sg.food.closure.last}{51.6}
\pcpdef{sg.food.frozen.additions}{16{,}215}
\pcpdef{sg.food.frozen.churnfinal}{5.55}
\pcpdef{sg.food.frozen.flipexcess}{37{,}761}
\pcpdef{sg.food.frozen.macrofinal}{0.9255}
\pcpdef{sg.food.frozen.netchanges}{120{,}459}
\pcpdef{sg.food.frozen.pathflips}{158{,}220}
\pcpdef{sg.food.frozen.reachfinal}{86.6}
\pcpdef{sg.food.frozen.reachfirst}{48.9}
\pcpdef{sg.food.frozen.revocations}{104{,}244}
\pcpdef{sg.food.frozen.sizefinal}{10.632}
\pcpdef{sg.food.frozen.worstfinal}{0.7588}
\pcpdef{sg.food.kernel.bitschanged}{0}
\pcpdef{sg.food.kernel.maxdelta}{3.33\times10^{-16}}
\pcpdef{sg.food.newclosure.frozen}{27.2}
\pcpdef{sg.food.newclosure.refresh}{28.4}
\pcpdef{sg.food.onboarding.maxdelta}{8.33\times10^{-17}}
\pcpdef{sg.food.pinned.churnfinal}{6.22}
\pcpdef{sg.food.pre.macrofinal}{0.9273}
\pcpdef{sg.food.pre.sizefinal}{17.425}
\pcpdef{sg.food.pre.worstfinal}{0.7459}
\pcpdef{sg.food.prefix.macrofinal}{0.9296}
\pcpdef{sg.food.prefix.sizefinal}{14.122}
\pcpdef{sg.food.prefix.worstfinal}{0.7642}
\pcpdef{sg.food.prefixes}{25, 40, 56, 71, 86, 101}
\pcpdef{sg.food.refitfinal}{86}
\pcpdef{sg.food.refresh.additions}{32{,}458}
\pcpdef{sg.food.refresh.churnfinal}{6.59}
\pcpdef{sg.food.refresh.macrofinal}{0.9284}
\pcpdef{sg.food.refresh.reachfinal}{92.2}
\pcpdef{sg.food.refresh.reachfirst}{66.1}
\pcpdef{sg.food.refresh.revocations}{110{,}439}
\pcpdef{sg.food.refresh.sizefinal}{11.029}
\pcpdef{sg.food.refresh.worstfinal}{0.7536}
\pcpdef{sg.food.rescoredfinal}{1{,}720}
\pcpdef{sg.gate.entries}{2{,}340}
\pcpdef{sg.versions}{6}
\pcpdef{st.K}{196}
\pcpdef{st.flip.frozen}{96\pcpdash{}98}
\pcpdef{st.flip.frozen.max}{98}
\pcpdef{st.flip.pre}{0}
\pcpdef{st.flip.prefixcp}{0}
\pcpdef{st.flip.refresh}{96\pcpdash{}98}
\pcpdef{st.flip.refresh.max}{98}
\pcpdef{st.oldclasses}{98}
\pcpdef{st.refit.frozen}{0}
\pcpdef{st.refit.refresh}{98}
\pcpdef{st.scoredefs.frozen}{98}
\pcpdef{st.scoredefs.refresh}{98}
\pcpdef{st.seeds}{5}
\pcpdef{sup.cars-clip-m20.below.frozen}{78}
\pcpdef{sup.cars-clip-m20.below.pre}{79}
\pcpdef{sup.cars-clip-m20.below.prefix}{79}
\pcpdef{sup.cars-clip-m20.below.refresh}{79}
\pcpdef{sup.cars-clip-m20.nclasses}{196}
\pcpdef{sup.cars-siglip-m20.below.frozen}{84}
\pcpdef{sup.cars-siglip-m20.below.pre}{85}
\pcpdef{sup.cars-siglip-m20.below.prefix}{85}
\pcpdef{sup.cars-siglip-m20.below.refresh}{84}
\pcpdef{sup.cars-siglip-m20.nclasses}{196}
\pcpdef{sup.closure.size.range}{44.8\pcpdash{}53.8}
\pcpdef{sup.closure.tail.range}{52.5\pcpdash{}72.8}
\pcpdef{sup.cub200-clip-m20.below.frozen}{78}
\pcpdef{sup.cub200-clip-m20.below.pre}{74}
\pcpdef{sup.cub200-clip-m20.below.prefix}{75}
\pcpdef{sup.cub200-clip-m20.below.refresh}{74}
\pcpdef{sup.cub200-clip-m20.nclasses}{200}
\pcpdef{sup.cub200-siglip-m20.below.frozen}{78}
\pcpdef{sup.cub200-siglip-m20.below.pre}{76}
\pcpdef{sup.cub200-siglip-m20.below.prefix}{77}
\pcpdef{sup.cub200-siglip-m20.below.refresh}{76}
\pcpdef{sup.cub200-siglip-m20.nclasses}{200}
\pcpdef{sup.food-siglip-m20.below.frozen}{27}
\pcpdef{sup.food-siglip-m20.below.pre}{26}
\pcpdef{sup.food-siglip-m20.below.prefix}{25}
\pcpdef{sup.food-siglip-m20.below.refresh}{23}
\pcpdef{sup.food-siglip-m20.nclasses}{101}
\pcpdef{sup.food-siglip-m40.below.frozen}{27}
\pcpdef{sup.food-siglip-m40.below.pre}{26}
\pcpdef{sup.food-siglip-m40.below.prefix}{23}
\pcpdef{sup.food-siglip-m40.below.refresh}{23}
\pcpdef{sup.food-siglip-m40.nclasses}{101}
\pcpdef{sup.food-siglip-m80.below.frozen}{20}
\pcpdef{sup.food-siglip-m80.below.pre}{17}
\pcpdef{sup.food-siglip-m80.below.prefix}{17}
\pcpdef{sup.food-siglip-m80.below.refresh}{15}
\pcpdef{sup.food-siglip-m80.nclasses}{101}
\pcpdef{sup.imagenet-r-clip-m20.below.frozen}{77}
\pcpdef{sup.imagenet-r-clip-m20.below.pre}{82}
\pcpdef{sup.imagenet-r-clip-m20.below.prefix}{80}
\pcpdef{sup.imagenet-r-clip-m20.below.refresh}{78}
\pcpdef{sup.imagenet-r-clip-m20.nclasses}{200}
\pcpdef{sup.imagenet-r-siglip-m20.below.frozen}{66}
\pcpdef{sup.imagenet-r-siglip-m20.below.pre}{79}
\pcpdef{sup.imagenet-r-siglip-m20.below.prefix}{79}
\pcpdef{sup.imagenet-r-siglip-m20.below.refresh}{80}
\pcpdef{sup.imagenet-r-siglip-m20.nclasses}{200}
\pcpdef{sup.imagenet-sketch-clip-m20.below.frozen}{381}
\pcpdef{sup.imagenet-sketch-clip-m20.below.pre}{374}
\pcpdef{sup.imagenet-sketch-clip-m20.below.prefix}{372}
\pcpdef{sup.imagenet-sketch-clip-m20.below.refresh}{371}
\pcpdef{sup.imagenet-sketch-clip-m20.nclasses}{930}
\pcpdef{sup.imagenet-sketch-siglip-m20.below.frozen}{358}
\pcpdef{sup.imagenet-sketch-siglip-m20.below.pre}{375}
\pcpdef{sup.imagenet-sketch-siglip-m20.below.prefix}{373}
\pcpdef{sup.imagenet-sketch-siglip-m20.below.refresh}{378}
\pcpdef{sup.imagenet-sketch-siglip-m20.nclasses}{930}
\pcpdef{sup.tau.closed.range}{40.9\pcpdash{}58.5}
\pcpdef{sup.topq.frozenbest}{30/33}
\pcpdef{sup.topq.rawworst}{33/33}
\pcpdef{tab.confot.body}{%
\;Conf-OT, Food-101 / SigLIP $m{=}20$ & --- & --- & --- & 10.45 & --- & --- & --- & ---\\
\;Conf-OT, Stanford Cars / SigLIP & --- & --- & --- & 14.75 & --- & --- & --- & ---\\
\;Conf-OT, Stanford Cars / CLIP & --- & --- & --- & 27.78 & --- & --- & --- & ---\\}
\pcpdef{tab.frontier.body}{%
Food-101 / SigLIP $m{=}20$ & 101 & 17.74 & 13.73 & \textbf{12.15} & 11.03 & 10.33$^{\ddagger}$ & 58.5 & $+1.58$ & $-1.12\,[-1.66,-0.58]$\\
Food-101 / SigLIP $m{=}40$ & 101 & 15.20 & 11.78 & \textbf{10.47} & 9.31 & 8.83$^{\ddagger}$ & 52.8 & $+1.31$ & $-1.17\,[-1.51,-0.82]$\\
Food-101 / SigLIP $m{=}80$ & 101 & 14.31 & 11.32 & \textbf{10.05} & 8.78 & 8.43$^{\ddagger}$ & 50.0 & $+1.27$ & $-1.27\,[-1.53,-1.00]$\\
Stanford Cars / SigLIP & 196 & 30.13 & 17.60 & \textbf{16.51} & 15.24 & 15.08$^{\ddagger}$ & 46.4 & $+1.09$ & $-1.26\,[-1.52,-1.00]$\\
Stanford Cars / CLIP & 196 & 44.85 & 30.09 & \textbf{29.55} & 28.78 & 29.02$^{\ddagger}$ & 40.9 & $+0.54$ & $-0.77\,[-1.00,-0.55]$\\
ImageNet-R / CLIP & 200 & 14.16 & 7.03 & \textbf{5.83} & 4.86 & 4.54$^{\ddagger}$ & 55.3 & $+1.20$ & $-0.97\,[-1.38,-0.56]$\\
ImageNet-R / SigLIP & 200 & 5.67 & 3.30 & \textbf{2.63} & 1.82 & 1.77$^{\ddagger}$ & 45.7 & $+0.67$ & $-0.80\,[-1.26,-0.35]$\\
CUB-200 / CLIP & 200 & 60.29 & 30.25 & \textbf{29.72} & 29.23 & 28.85$^{\ddagger}$ & 52.1 & $+0.53$ & $-0.49\,[-0.72,-0.25]$\\
CUB-200 / SigLIP & 200 & 44.45 & 21.63 & \textbf{21.01} & 20.32 & 20.25$^{\ddagger}$ & 47.1 & $+0.62$ & $-0.69\,[-0.92,-0.46]$\\
ImageNet-Sketch / CLIP & 930 & 45.48 & 30.16 & \textbf{27.01} & 24.31 & 23.64$^{\ddagger}$ & 53.8 & $+3.14$ & $-2.70\,[-3.39,-2.01]$\\
ImageNet-Sketch / SigLIP & 930 & 22.20 & 14.05 & \textbf{11.68} & 8.93 & 8.66$^{\ddagger}$ & 46.3 & $+2.37$ & $-2.75\,[-3.33,-2.16]$\\}
\pcpdef{tab.frontier.body.m20}{%
Food-101 / SigLIP $m{=}20$ & 101 & 17.74 & 13.73 & \textbf{12.15} & 11.03 & 10.33$^{\ddagger}$ & 58.5 & $+1.58$ & $-1.12\,[-1.66,-0.58]$\\
Stanford Cars / SigLIP & 196 & 30.13 & 17.60 & \textbf{16.51} & 15.24 & 15.08$^{\ddagger}$ & 46.4 & $+1.09$ & $-1.26\,[-1.52,-1.00]$\\
Stanford Cars / CLIP & 196 & 44.85 & 30.09 & \textbf{29.55} & 28.78 & 29.02$^{\ddagger}$ & 40.9 & $+0.54$ & $-0.77\,[-1.00,-0.55]$\\
ImageNet-R / CLIP & 200 & 14.16 & 7.03 & \textbf{5.83} & 4.86 & 4.54$^{\ddagger}$ & 55.3 & $+1.20$ & $-0.97\,[-1.38,-0.56]$\\
ImageNet-R / SigLIP & 200 & 5.67 & 3.30 & \textbf{2.63} & 1.82 & 1.77$^{\ddagger}$ & 45.7 & $+0.67$ & $-0.80\,[-1.26,-0.35]$\\
CUB-200 / CLIP & 200 & 60.29 & 30.25 & \textbf{29.72} & 29.23 & 28.85$^{\ddagger}$ & 52.1 & $+0.53$ & $-0.49\,[-0.72,-0.25]$\\
CUB-200 / SigLIP & 200 & 44.45 & 21.63 & \textbf{21.01} & 20.32 & 20.25$^{\ddagger}$ & 47.1 & $+0.62$ & $-0.69\,[-0.92,-0.46]$\\
ImageNet-Sketch / CLIP & 930 & 45.48 & 30.16 & \textbf{27.01} & 24.31 & 23.64$^{\ddagger}$ & 53.8 & $+3.14$ & $-2.70\,[-3.39,-2.01]$\\
ImageNet-Sketch / SigLIP & 930 & 22.20 & 14.05 & \textbf{11.68} & 8.93 & 8.66$^{\ddagger}$ & 46.3 & $+2.37$ & $-2.75\,[-3.33,-2.16]$\\}
\pcpdef{tab.main.schedule}{%
Food-101 / SigLIP & 13.727 & 12.148 & high-confusion-first & 12.999 & 53.9\\
Stanford Cars / CLIP & 30.090 & 29.555 & near-base-first & 29.848 & 54.9\\
Stanford Cars / SigLIP & 17.598 & 16.506 & near-base-first & 17.205 & 64.0\\}
\pcpdef{tab.main.temperature}{%
0.80 & 17.553 & 16.536 & 15.357 & 0.8982 & 0\\
1.00 & 17.598 & 16.506 & 15.242 & 0.8983 & 0\\
1.20 & 17.853 & 16.692 & 15.405 & 0.8992 & 0\\}
\pcpdef{tab.mutation.body}{%
HPS-Frozen (mean)$^{\ddagger}$ & 8.4\pcpdash{}51.8 & 0.1\pcpdash{}0.9 & 48\pcpdash{}78 & 0.153\pcpdash{}0.789 & \textbf{no} & 0\\
HPS-Refresh (mean) & 11.6\pcpdash{}52.7 & 0.2\pcpdash{}2.5 & 53\pcpdash{}90 & 0.504\pcpdash{}0.833 & by refit & all\\
\midrule
raw cosine & \textbf{0} & \textbf{0} & \textbf{0} & 0.503\pcpdash{}0.823 & \textbf{yes} & \textbf{0}\\
PRE & \textbf{0} & \textbf{0} & \textbf{0} & 0.500\pcpdash{}0.833 & \textbf{yes} & \textbf{0}\\
\textbf{PrefixCP} & \textbf{0} & \textbf{0} & \textbf{0} & 0.483\pcpdash{}0.843 & \textbf{yes} & \textbf{0}\\}
\pcpdef{tab.runtime.body}{%
$K{=}930$, $\nver{=}2$, ms/batch & 26.98 & \textbf{58.18} & 85.47 & 52.38 & 52.38 & ---\\
$K{=}930$, $\nver{=}6$, ms/batch & 13.45 & \textbf{74.95} & 226.08 & 52.38 & 52.38 & ---\\
$K{=}10^5$, $\nver{=}500$, ms & 0.50 & \textbf{390.22} & 145\,s & 263.95 & 263.95 & ---\\
$K{=}10^5$, $\nver{=}100$, MiB & 24 & \textbf{122} & 14{,}645 & 2{,}442 & 2{,}442 & ---\\}
\pcpdef{tab.sup.arms}{%
\multicolumn{12}{@{}l@{}}{\emph{Food-101 / SigLIP $m{=}20$} ($K{=}101$)}\\
\;raw cosine & 17.736 & 17.735 & 18.06 & 24.68 & 25.79 & 27.80 & 1.08 & 0.16 & 0.9334 & 0.8692 & 0.7600\\
\;PRE & 13.727 & 13.725 & 13.89 & 21.07 & 22.60 & 26.74 & 8.44 & 0.00 & 0.9269 & 0.8575 & 0.7440\\
\;\textbf{PrefixCP} & 12.148 & 12.147 & 12.33 & 17.44 & 18.53 & 20.77 & 6.78 & 0.00 & 0.9267 & 0.8597 & 0.7424\\
\;HPS-Refresh & 11.029 & 11.027 & 11.04 & 16.60 & 17.72 & 20.02 & 10.89 & 0.00 & 0.9283 & 0.8536 & 0.7448\\
\;HPS-Frozen$^{\ddagger}$ & 10.330 & 10.328 & 10.38 & 15.70 & 16.60 & 18.85 & 11.66 & 0.00 & 0.9239 & 0.8512 & 0.7088\\
\midrule
\multicolumn{12}{@{}l@{}}{\emph{Food-101 / SigLIP $m{=}40$} ($K{=}101$)}\\
\;raw cosine & 15.199 & 15.198 & 15.52 & 21.75 & 22.68 & 24.35 & 2.25 & 0.35 & 0.9314 & 0.8751 & 0.7918\\
\;PRE & 11.781 & 11.779 & 11.85 & 18.33 & 19.72 & 23.75 & 10.36 & 0.00 & 0.9253 & 0.8693 & 0.7984\\
\;\textbf{PrefixCP} & 10.475 & 10.473 & 10.66 & 15.15 & 16.14 & 18.29 & 8.39 & 0.00 & 0.9256 & 0.8678 & 0.8046\\
\;HPS-Refresh & 9.308 & 9.306 & 9.26 & 14.34 & 15.17 & 17.27 & 13.86 & 0.00 & 0.9263 & 0.8758 & 0.7976\\
\;HPS-Frozen$^{\ddagger}$ & 8.832 & 8.831 & 8.79 & 13.63 & 14.54 & 16.50 & 14.63 & 0.00 & 0.9214 & 0.8653 & 0.7488\\
\midrule
\multicolumn{12}{@{}l@{}}{\emph{Food-101 / SigLIP $m{=}80$} ($K{=}101$)}\\
\;raw cosine & 14.307 & 14.305 & 14.68 & 20.85 & 21.64 & 23.24 & 2.67 & 0.47 & 0.9342 & 0.8903 & 0.8228\\
\;PRE & 11.316 & 11.315 & 11.25 & 17.69 & 18.93 & 22.71 & 10.84 & 0.00 & 0.9292 & 0.8880 & 0.8332\\
\;\textbf{PrefixCP} & 10.051 & 10.050 & 10.13 & 14.66 & 15.59 & 17.70 & 8.75 & 0.00 & 0.9288 & 0.8880 & 0.8428\\
\;HPS-Refresh & 8.784 & 8.783 & 8.64 & 13.48 & 14.32 & 16.52 & 14.77 & 0.00 & 0.9289 & 0.8871 & 0.8333\\
\;HPS-Frozen$^{\ddagger}$ & 8.425 & 8.424 & 8.28 & 13.04 & 13.87 & 15.84 & 15.43 & 0.00 & 0.9241 & 0.8839 & 0.7887\\
\midrule
\multicolumn{12}{@{}l@{}}{\emph{Stanford Cars / SigLIP} ($K{=}196$)}\\
\;raw cosine & 30.133 & 30.225 & 30.84 & 45.70 & 49.85 & 57.86 & 0.51 & 0.19 & 0.8972 & 0.7927 & 0.6226\\
\;PRE & 17.598 & 17.631 & 17.46 & 28.27 & 31.04 & 37.52 & 7.51 & 0.03 & 0.8983 & 0.7954 & 0.6378\\
\;\textbf{PrefixCP} & 16.506 & 16.558 & 16.25 & 26.91 & 30.31 & 36.23 & 6.74 & 0.02 & 0.8983 & 0.7977 & 0.6308\\
\;HPS-Refresh & 15.242 & 15.315 & 14.76 & 26.13 & 29.03 & 35.89 & 11.33 & 0.02 & 0.8987 & 0.7906 & 0.6258\\
\;HPS-Frozen$^{\ddagger}$ & 15.083 & 15.153 & 14.60 & 25.92 & 28.60 & 35.67 & 11.27 & 0.01 & 0.8983 & 0.7908 & 0.5790\\
\midrule
\multicolumn{12}{@{}l@{}}{\emph{Stanford Cars / CLIP} ($K{=}196$)}\\
\;raw cosine & 44.854 & 44.957 & 46.19 & 63.97 & 69.75 & 90.12 & 0.29 & 0.15 & 0.9068 & 0.8090 & 0.6451\\
\;PRE & 30.090 & 30.158 & 30.83 & 40.56 & 43.01 & 53.39 & 0.54 & 0.00 & 0.9020 & 0.7979 & 0.6450\\
\;\textbf{PrefixCP} & 29.555 & 29.629 & 29.81 & 40.21 & 42.40 & 53.20 & 0.42 & 0.00 & 0.9023 & 0.7970 & 0.6539\\
\;HPS-Refresh & 28.780 & 28.860 & 28.90 & 39.90 & 42.03 & 52.80 & 0.58 & 0.00 & 0.9031 & 0.7993 & 0.6640\\
\;HPS-Frozen$^{\ddagger}$ & 29.021 & 29.101 & 29.13 & 40.27 & 42.23 & 53.43 & 0.55 & 0.00 & 0.9035 & 0.8030 & 0.6304\\
\midrule
\multicolumn{12}{@{}l@{}}{\emph{ImageNet-R / CLIP} ($K{=}200$)}\\
\;raw cosine & 14.159 & 14.244 & 13.79 & 19.86 & 20.98 & 29.08 & 3.64 & 1.02 & 0.9037 & 0.8153 & 0.5813\\
\;PRE & 7.033 & 6.892 & 6.12 & 12.45 & 13.66 & 16.34 & 17.73 & 0.04 & 0.9012 & 0.8001 & 0.5922\\
\;\textbf{PrefixCP} & 5.833 & 5.728 & 5.59 & 8.70 & 9.54 & 11.46 & 15.43 & 0.03 & 0.9024 & 0.8066 & 0.5929\\
\;HPS-Refresh & 4.861 & 4.773 & 4.61 & 7.06 & 7.68 & 10.41 & 25.23 & 0.02 & 0.9039 & 0.8090 & 0.5714\\
\;HPS-Frozen$^{\ddagger}$ & 4.541 & 4.459 & 4.30 & 6.57 & 7.27 & 9.74 & 27.00 & 0.02 & 0.9027 & 0.8112 & 0.5996\\
\midrule
\multicolumn{12}{@{}l@{}}{\emph{ImageNet-R / SigLIP} ($K{=}200$)}\\
\;raw cosine & 5.674 & 5.752 & 5.42 & 8.50 & 9.31 & 11.45 & 14.76 & 2.96 & 0.9036 & 0.8123 & 0.5550\\
\;PRE & 3.300 & 3.293 & 2.54 & 6.31 & 6.94 & 9.07 & 36.85 & 0.79 & 0.9036 & 0.8124 & 0.6508\\
\;\textbf{PrefixCP} & 2.626 & 2.635 & 2.41 & 4.40 & 4.75 & 5.90 & 36.67 & 0.69 & 0.9037 & 0.8136 & 0.6500\\
\;HPS-Refresh & 1.824 & 1.844 & 1.72 & 2.74 & 3.03 & 3.94 & 58.89 & 0.67 & 0.9025 & 0.8051 & 0.6460\\
\;HPS-Frozen$^{\ddagger}$ & 1.767 & 1.785 & 1.65 & 2.66 & 2.90 & 3.77 & 62.05 & 0.55 & 0.9127 & 0.8262 & 0.6349\\
\midrule
\multicolumn{12}{@{}l@{}}{\emph{CUB-200 / CLIP} ($K{=}200$)}\\
\;raw cosine & 60.291 & 60.163 & 63.45 & 81.07 & 83.57 & 88.80 & 0.05 & 0.00 & 0.8984 & 0.7733 & 0.5933\\
\;PRE & 30.250 & 30.247 & 30.03 & 43.96 & 45.96 & 57.25 & 0.24 & 0.00 & 0.8945 & 0.7800 & 0.6000\\
\;\textbf{PrefixCP} & 29.721 & 29.721 & 29.46 & 43.25 & 45.27 & 58.19 & 0.14 & 0.00 & 0.8957 & 0.7733 & 0.6108\\
\;HPS-Refresh & 29.234 & 29.237 & 29.19 & 42.80 & 45.10 & 58.75 & 0.16 & 0.00 & 0.8969 & 0.7793 & 0.6067\\
\;HPS-Frozen$^{\ddagger}$ & 28.849 & 28.853 & 28.62 & 42.42 & 44.59 & 58.22 & 0.17 & 0.00 & 0.8922 & 0.7733 & 0.5634\\
\midrule
\multicolumn{12}{@{}l@{}}{\emph{CUB-200 / SigLIP} ($K{=}200$)}\\
\;raw cosine & 44.448 & 44.345 & 46.57 & 67.46 & 70.96 & 81.87 & 0.49 & 0.48 & 0.8948 & 0.7865 & 0.5267\\
\;PRE & 21.630 & 21.613 & 19.12 & 37.35 & 39.77 & 44.72 & 2.33 & 0.00 & 0.8955 & 0.7533 & 0.6200\\
\;\textbf{PrefixCP} & 21.014 & 21.001 & 18.22 & 36.87 & 38.94 & 43.94 & 1.84 & 0.00 & 0.8955 & 0.7533 & 0.5975\\
\;HPS-Refresh & 20.320 & 20.312 & 17.72 & 36.51 & 38.51 & 43.46 & 2.72 & 0.00 & 0.8960 & 0.7659 & 0.5800\\
\;HPS-Frozen$^{\ddagger}$ & 20.248 & 20.239 & 17.64 & 36.33 & 38.16 & 43.23 & 2.70 & 0.00 & 0.8943 & 0.7653 & 0.5533\\
\midrule
\multicolumn{12}{@{}l@{}}{\emph{ImageNet-Sketch / CLIP} ($K{=}930$)}\\
\;raw cosine & 45.477 & 45.431 & 40.15 & 75.18 & 92.00 & 175.81 & 0.17 & 0.08 & 0.9041 & 0.7905 & 0.5090\\
\;PRE & 30.157 & 30.100 & 27.62 & 50.82 & 58.17 & 87.37 & 1.52 & 0.00 & 0.9034 & 0.7891 & 0.5000\\
\;\textbf{PrefixCP} & 27.014 & 26.982 & 25.53 & 42.67 & 46.86 & 65.68 & 1.10 & 0.00 & 0.9037 & 0.7918 & 0.4833\\
\;HPS-Refresh & 24.314 & 24.308 & 22.63 & 39.55 & 43.36 & 64.43 & 1.79 & 0.00 & 0.9045 & 0.7912 & 0.5043\\
\;HPS-Frozen$^{\ddagger}$ & 23.638 & 23.632 & 22.09 & 38.34 & 42.07 & 62.21 & 1.84 & 0.01 & 0.9010 & 0.7789 & 0.3077\\
\midrule
\multicolumn{12}{@{}l@{}}{\emph{ImageNet-Sketch / SigLIP} ($K{=}930$)}\\
\;raw cosine & 22.199 & 22.243 & 19.42 & 38.53 & 45.30 & 88.60 & 2.55 & 1.26 & 0.9015 & 0.7742 & 0.5034\\
\;PRE & 14.049 & 14.025 & 12.10 & 27.13 & 32.06 & 50.12 & 5.50 & 0.03 & 0.9031 & 0.7905 & 0.5294\\
\;\textbf{PrefixCP} & 11.677 & 11.678 & 11.11 & 19.84 & 22.95 & 37.14 & 4.50 & 0.02 & 0.9039 & 0.7852 & 0.5314\\
\;HPS-Refresh & 8.929 & 8.956 & 7.91 & 15.59 & 19.22 & 33.80 & 7.44 & 0.03 & 0.9032 & 0.7726 & 0.5160\\
\;HPS-Frozen$^{\ddagger}$ & 8.660 & 8.686 & 7.68 & 15.06 & 18.56 & 32.29 & 7.29 & 0.02 & 0.9054 & 0.7789 & 0.1528\\}
\pcpdef{tab.sup.arms.a}{%
\multicolumn{12}{@{}l@{}}{\emph{Food-101 / SigLIP $m{=}20$} ($K{=}101$)}\\
\;raw cosine & 17.736 & 17.735 & 18.06 & 24.68 & 25.79 & 27.80 & 1.08 & 0.16 & 0.9334 & 0.8692 & 0.7600\\
\;PRE & 13.727 & 13.725 & 13.89 & 21.07 & 22.60 & 26.74 & 8.44 & 0.00 & 0.9269 & 0.8575 & 0.7440\\
\;\textbf{PrefixCP} & 12.148 & 12.147 & 12.33 & 17.44 & 18.53 & 20.77 & 6.78 & 0.00 & 0.9267 & 0.8597 & 0.7424\\
\;HPS-Refresh & 11.029 & 11.027 & 11.04 & 16.60 & 17.72 & 20.02 & 10.89 & 0.00 & 0.9283 & 0.8536 & 0.7448\\
\;HPS-Frozen$^{\ddagger}$ & 10.330 & 10.328 & 10.38 & 15.70 & 16.60 & 18.85 & 11.66 & 0.00 & 0.9239 & 0.8512 & 0.7088\\
\midrule
\multicolumn{12}{@{}l@{}}{\emph{Food-101 / SigLIP $m{=}40$} ($K{=}101$)}\\
\;raw cosine & 15.199 & 15.198 & 15.52 & 21.75 & 22.68 & 24.35 & 2.25 & 0.35 & 0.9314 & 0.8751 & 0.7918\\
\;PRE & 11.781 & 11.779 & 11.85 & 18.33 & 19.72 & 23.75 & 10.36 & 0.00 & 0.9253 & 0.8693 & 0.7984\\
\;\textbf{PrefixCP} & 10.475 & 10.473 & 10.66 & 15.15 & 16.14 & 18.29 & 8.39 & 0.00 & 0.9256 & 0.8678 & 0.8046\\
\;HPS-Refresh & 9.308 & 9.306 & 9.26 & 14.34 & 15.17 & 17.27 & 13.86 & 0.00 & 0.9263 & 0.8758 & 0.7976\\
\;HPS-Frozen$^{\ddagger}$ & 8.832 & 8.831 & 8.79 & 13.63 & 14.54 & 16.50 & 14.63 & 0.00 & 0.9214 & 0.8653 & 0.7488\\
\midrule
\multicolumn{12}{@{}l@{}}{\emph{Food-101 / SigLIP $m{=}80$} ($K{=}101$)}\\
\;raw cosine & 14.307 & 14.305 & 14.68 & 20.85 & 21.64 & 23.24 & 2.67 & 0.47 & 0.9342 & 0.8903 & 0.8228\\
\;PRE & 11.316 & 11.315 & 11.25 & 17.69 & 18.93 & 22.71 & 10.84 & 0.00 & 0.9292 & 0.8880 & 0.8332\\
\;\textbf{PrefixCP} & 10.051 & 10.050 & 10.13 & 14.66 & 15.59 & 17.70 & 8.75 & 0.00 & 0.9288 & 0.8880 & 0.8428\\
\;HPS-Refresh & 8.784 & 8.783 & 8.64 & 13.48 & 14.32 & 16.52 & 14.77 & 0.00 & 0.9289 & 0.8871 & 0.8333\\
\;HPS-Frozen$^{\ddagger}$ & 8.425 & 8.424 & 8.28 & 13.04 & 13.87 & 15.84 & 15.43 & 0.00 & 0.9241 & 0.8839 & 0.7887\\
\midrule
\multicolumn{12}{@{}l@{}}{\emph{Stanford Cars / SigLIP} ($K{=}196$)}\\
\;raw cosine & 30.133 & 30.225 & 30.84 & 45.70 & 49.85 & 57.86 & 0.51 & 0.19 & 0.8972 & 0.7927 & 0.6226\\
\;PRE & 17.598 & 17.631 & 17.46 & 28.27 & 31.04 & 37.52 & 7.51 & 0.03 & 0.8983 & 0.7954 & 0.6378\\
\;\textbf{PrefixCP} & 16.506 & 16.558 & 16.25 & 26.91 & 30.31 & 36.23 & 6.74 & 0.02 & 0.8983 & 0.7977 & 0.6308\\
\;HPS-Refresh & 15.242 & 15.315 & 14.76 & 26.13 & 29.03 & 35.89 & 11.33 & 0.02 & 0.8987 & 0.7906 & 0.6258\\
\;HPS-Frozen$^{\ddagger}$ & 15.083 & 15.153 & 14.60 & 25.92 & 28.60 & 35.67 & 11.27 & 0.01 & 0.8983 & 0.7908 & 0.5790\\
\midrule
\multicolumn{12}{@{}l@{}}{\emph{Stanford Cars / CLIP} ($K{=}196$)}\\
\;raw cosine & 44.854 & 44.957 & 46.19 & 63.97 & 69.75 & 90.12 & 0.29 & 0.15 & 0.9068 & 0.8090 & 0.6451\\
\;PRE & 30.090 & 30.158 & 30.83 & 40.56 & 43.01 & 53.39 & 0.54 & 0.00 & 0.9020 & 0.7979 & 0.6450\\
\;\textbf{PrefixCP} & 29.555 & 29.629 & 29.81 & 40.21 & 42.40 & 53.20 & 0.42 & 0.00 & 0.9023 & 0.7970 & 0.6539\\
\;HPS-Refresh & 28.780 & 28.860 & 28.90 & 39.90 & 42.03 & 52.80 & 0.58 & 0.00 & 0.9031 & 0.7993 & 0.6640\\
\;HPS-Frozen$^{\ddagger}$ & 29.021 & 29.101 & 29.13 & 40.27 & 42.23 & 53.43 & 0.55 & 0.00 & 0.9035 & 0.8030 & 0.6304\\}
\pcpdef{tab.sup.arms.b}{%
\multicolumn{12}{@{}l@{}}{\emph{ImageNet-R / CLIP} ($K{=}200$)}\\
\;raw cosine & 14.159 & 14.244 & 13.79 & 19.86 & 20.98 & 29.08 & 3.64 & 1.02 & 0.9037 & 0.8153 & 0.5813\\
\;PRE & 7.033 & 6.892 & 6.12 & 12.45 & 13.66 & 16.34 & 17.73 & 0.04 & 0.9012 & 0.8001 & 0.5922\\
\;\textbf{PrefixCP} & 5.833 & 5.728 & 5.59 & 8.70 & 9.54 & 11.46 & 15.43 & 0.03 & 0.9024 & 0.8066 & 0.5929\\
\;HPS-Refresh & 4.861 & 4.773 & 4.61 & 7.06 & 7.68 & 10.41 & 25.23 & 0.02 & 0.9039 & 0.8090 & 0.5714\\
\;HPS-Frozen$^{\ddagger}$ & 4.541 & 4.459 & 4.30 & 6.57 & 7.27 & 9.74 & 27.00 & 0.02 & 0.9027 & 0.8112 & 0.5996\\
\midrule
\multicolumn{12}{@{}l@{}}{\emph{ImageNet-R / SigLIP} ($K{=}200$)}\\
\;raw cosine & 5.674 & 5.752 & 5.42 & 8.50 & 9.31 & 11.45 & 14.76 & 2.96 & 0.9036 & 0.8123 & 0.5550\\
\;PRE & 3.300 & 3.293 & 2.54 & 6.31 & 6.94 & 9.07 & 36.85 & 0.79 & 0.9036 & 0.8124 & 0.6508\\
\;\textbf{PrefixCP} & 2.626 & 2.635 & 2.41 & 4.40 & 4.75 & 5.90 & 36.67 & 0.69 & 0.9037 & 0.8136 & 0.6500\\
\;HPS-Refresh & 1.824 & 1.844 & 1.72 & 2.74 & 3.03 & 3.94 & 58.89 & 0.67 & 0.9025 & 0.8051 & 0.6460\\
\;HPS-Frozen$^{\ddagger}$ & 1.767 & 1.785 & 1.65 & 2.66 & 2.90 & 3.77 & 62.05 & 0.55 & 0.9127 & 0.8262 & 0.6349\\
\midrule
\multicolumn{12}{@{}l@{}}{\emph{CUB-200 / CLIP} ($K{=}200$)}\\
\;raw cosine & 60.291 & 60.163 & 63.45 & 81.07 & 83.57 & 88.80 & 0.05 & 0.00 & 0.8984 & 0.7733 & 0.5933\\
\;PRE & 30.250 & 30.247 & 30.03 & 43.96 & 45.96 & 57.25 & 0.24 & 0.00 & 0.8945 & 0.7800 & 0.6000\\
\;\textbf{PrefixCP} & 29.721 & 29.721 & 29.46 & 43.25 & 45.27 & 58.19 & 0.14 & 0.00 & 0.8957 & 0.7733 & 0.6108\\
\;HPS-Refresh & 29.234 & 29.237 & 29.19 & 42.80 & 45.10 & 58.75 & 0.16 & 0.00 & 0.8969 & 0.7793 & 0.6067\\
\;HPS-Frozen$^{\ddagger}$ & 28.849 & 28.853 & 28.62 & 42.42 & 44.59 & 58.22 & 0.17 & 0.00 & 0.8922 & 0.7733 & 0.5634\\
\midrule
\multicolumn{12}{@{}l@{}}{\emph{CUB-200 / SigLIP} ($K{=}200$)}\\
\;raw cosine & 44.448 & 44.345 & 46.57 & 67.46 & 70.96 & 81.87 & 0.49 & 0.48 & 0.8948 & 0.7865 & 0.5267\\
\;PRE & 21.630 & 21.613 & 19.12 & 37.35 & 39.77 & 44.72 & 2.33 & 0.00 & 0.8955 & 0.7533 & 0.6200\\
\;\textbf{PrefixCP} & 21.014 & 21.001 & 18.22 & 36.87 & 38.94 & 43.94 & 1.84 & 0.00 & 0.8955 & 0.7533 & 0.5975\\
\;HPS-Refresh & 20.320 & 20.312 & 17.72 & 36.51 & 38.51 & 43.46 & 2.72 & 0.00 & 0.8960 & 0.7659 & 0.5800\\
\;HPS-Frozen$^{\ddagger}$ & 20.248 & 20.239 & 17.64 & 36.33 & 38.16 & 43.23 & 2.70 & 0.00 & 0.8943 & 0.7653 & 0.5533\\
\midrule
\multicolumn{12}{@{}l@{}}{\emph{ImageNet-Sketch / CLIP} ($K{=}930$)}\\
\;raw cosine & 45.477 & 45.431 & 40.15 & 75.18 & 92.00 & 175.81 & 0.17 & 0.08 & 0.9041 & 0.7905 & 0.5090\\
\;PRE & 30.157 & 30.100 & 27.62 & 50.82 & 58.17 & 87.37 & 1.52 & 0.00 & 0.9034 & 0.7891 & 0.5000\\
\;\textbf{PrefixCP} & 27.014 & 26.982 & 25.53 & 42.67 & 46.86 & 65.68 & 1.10 & 0.00 & 0.9037 & 0.7918 & 0.4833\\
\;HPS-Refresh & 24.314 & 24.308 & 22.63 & 39.55 & 43.36 & 64.43 & 1.79 & 0.00 & 0.9045 & 0.7912 & 0.5043\\
\;HPS-Frozen$^{\ddagger}$ & 23.638 & 23.632 & 22.09 & 38.34 & 42.07 & 62.21 & 1.84 & 0.01 & 0.9010 & 0.7789 & 0.3077\\
\midrule
\multicolumn{12}{@{}l@{}}{\emph{ImageNet-Sketch / SigLIP} ($K{=}930$)}\\
\;raw cosine & 22.199 & 22.243 & 19.42 & 38.53 & 45.30 & 88.60 & 2.55 & 1.26 & 0.9015 & 0.7742 & 0.5034\\
\;PRE & 14.049 & 14.025 & 12.10 & 27.13 & 32.06 & 50.12 & 5.50 & 0.03 & 0.9031 & 0.7905 & 0.5294\\
\;\textbf{PrefixCP} & 11.677 & 11.678 & 11.11 & 19.84 & 22.95 & 37.14 & 4.50 & 0.02 & 0.9039 & 0.7852 & 0.5314\\
\;HPS-Refresh & 8.929 & 8.956 & 7.91 & 15.59 & 19.22 & 33.80 & 7.44 & 0.03 & 0.9032 & 0.7726 & 0.5160\\
\;HPS-Frozen$^{\ddagger}$ & 8.660 & 8.686 & 7.68 & 15.06 & 18.56 & 32.29 & 7.29 & 0.02 & 0.9054 & 0.7789 & 0.1528\\}
\pcpdef{tab.sup.cadadv}{%
Food-101 / SigLIP & 13.727 & 12.148 & 1.578 & high-confusion-first & 12.999 & 0.728 & 53.9 & 40.5\\
Stanford Cars / CLIP & 30.090 & 29.555 & 0.535 & near-base-first & 29.848 & 0.241 & 54.9 & 37.3\\
Stanford Cars / SigLIP & 17.598 & 16.506 & 1.092 & near-base-first & 17.205 & 0.393 & 64.0 & 31.7\\}
\pcpdef{tab.sup.cadcov}{%
Food-101 / SigLIP & 0.9266\pcpdash{}0.9274 & 0.9267 & 0.0008 & 0.7424\pcpdash{}0.7472\\
Stanford Cars / CLIP & 0.9017\pcpdash{}0.9025 & 0.9023 & 0.0008 & 0.6494\pcpdash{}0.6539\\
Stanford Cars / SigLIP & 0.8975\pcpdash{}0.8989 & 0.8983 & 0.0015 & 0.6245\pcpdash{}0.6426\\}
\pcpdef{tab.sup.caddiff}{%
Food-101 / SigLIP & random & +0.640 & 1.346 & 5.26 & 18.1 & 68.8 & 1.55 & \textbf{61.8} & 1.56\\
Food-101 / SigLIP & near-base first & +0.783 & 1.298 & 5.38 & 16.7 & 65.9 & 1.43 & \textbf{57.9} & 1.44\\
Food-101 / SigLIP & far-base first & +0.658 & 1.373 & 5.02 & 15.4 & 72.7 & 1.59 & \textbf{64.8} & 1.61\\
Food-101 / SigLIP & high-confusion first & +0.851 & 1.400 & 5.80 & 16.7 & 68.4 & 1.54 & \textbf{59.9} & 1.55\\
Stanford Cars / CLIP & random & +0.200 & 1.017 & 3.26 & 16.1 & 74.6 & 0.66 & \textbf{63.5} & 1.29\\
Stanford Cars / CLIP & near-base first & +0.294 & 1.241 & 4.42 & 19.6 & 75.6 & 0.78 & \textbf{65.4} & 1.53\\
Stanford Cars / CLIP & far-base first & +0.220 & 1.012 & 3.14 & 10.3 & 76.7 & 0.66 & \textbf{65.5} & 1.29\\
Stanford Cars / CLIP & high-confusion first & +0.275 & 1.207 & 4.15 & 20.4 & 75.7 & 0.78 & \textbf{65.8} & 1.52\\
Stanford Cars / SigLIP & random & +0.412 & 0.975 & 3.92 & 16.4 & 57.9 & 0.59 & \textbf{50.5} & 1.16\\
Stanford Cars / SigLIP & near-base first & +0.699 & 1.315 & 5.55 & 19.8 & 61.8 & 0.79 & \textbf{55.3} & 1.54\\
Stanford Cars / SigLIP & far-base first & +0.346 & 0.843 & 3.18 & 12.6 & 57.2 & 0.52 & \textbf{49.6} & 1.02\\
Stanford Cars / SigLIP & high-confusion first & +0.662 & 1.209 & 5.20 & 19.2 & 60.1 & 0.71 & \textbf{53.1} & 1.39\\}
\pcpdef{tab.sup.cadence}{%
\multicolumn{6}{@{}l@{}}{\emph{Food-101 / SigLIP} ($K{=}101$)}\\
\;T2 & 2 & 12.148 & 12.148 & 12.148 & 12.148\\
\;T6 & 6 & 12.651 & 12.788 & 12.662 & 12.815\\
\;T20 & 20 & 12.757 & 12.898 & 12.777 & 12.958\\
\;Tper\_class & 52 & 12.788 & 12.932 & 12.806 & 12.999\\
\midrule
\multicolumn{6}{@{}l@{}}{\emph{Stanford Cars / CLIP} ($K{=}196$)}\\
\;T2 & 2 & 29.555 & 29.555 & 29.555 & 29.555\\
\;T6 & 6 & 29.670 & 29.777 & 29.707 & 29.770\\
\;T20 & 20 & 29.727 & 29.821 & 29.765 & 29.821\\
\;Tper\_class & 99 & 29.755 & 29.848 & 29.774 & 29.830\\
\midrule
\multicolumn{6}{@{}l@{}}{\emph{Stanford Cars / SigLIP} ($K{=}196$)}\\
\;T2 & 2 & 16.506 & 16.506 & 16.506 & 16.506\\
\;T6 & 6 & 16.847 & 17.051 & 16.767 & 17.010\\
\;T20 & 20 & 16.902 & 17.175 & 16.842 & 17.122\\
\;Tper\_class & 99 & 16.918 & 17.205 & 16.852 & 17.168\\}
\pcpdef{tab.sup.cadnovel}{%
Food-101 / SigLIP & 6.745 & 5.403 & 6.254 & +15.7\\
Stanford Cars / CLIP & 15.362 & 14.193 & 14.486 & +2.1\\
Stanford Cars / SigLIP & 9.107 & 7.399 & 8.098 & +9.4\\}
\pcpdef{tab.sup.cadtail}{%
Food-101 / SigLIP & 17.44 & 19.37 & +11.1 & 18.53 & 20.87 & +12.7 & 20.77 & 24.11 & +16.1 & 12.15 & 13.00 & +7.0\\
Stanford Cars / CLIP & 40.21 & 40.38 & +0.4 & 42.40 & 42.92 & +1.2 & 53.20 & 53.64 & +0.8 & 29.55 & 29.85 & +1.0\\
Stanford Cars / SigLIP & 26.91 & 27.59 & +2.5 & 30.31 & 30.82 & +1.7 & 36.23 & 37.06 & +2.3 & 16.51 & 17.20 & +4.2\\}
\pcpdef{tab.sup.cadwork}{%
Food-101 / SigLIP & T2 & 2 & 151 & 109 & 1.4 & 8.4 & 11.6 & 1.4\\
Food-101 / SigLIP & T6 & 6 & 455 & 141 & 3.2 & 10.4 & 36.6 & 3.5\\
Food-101 / SigLIP & T20 & 20 & 1{,}549 & 253 & 6.1 & 15.2 & 124.9 & 8.2\\
Food-101 / SigLIP & per-class & 52 & 3{,}926 & 509 & 7.7 & 20.7 & 318.6 & 15.4\\
Stanford Cars / CLIP & T2 & 2 & 294 & 204 & 1.4 & 5.0 & 6.2 & 1.2\\
Stanford Cars / CLIP & T6 & 6 & 885 & 236 & 3.8 & 4.9 & 19.4 & 4.0\\
Stanford Cars / CLIP & T20 & 20 & 2{,}964 & 348 & 8.5 & 6.8 & 66.1 & 9.7\\
Stanford Cars / CLIP & per-class & 99 & 14{,}553 & 980 & 14.9 & 11.1 & 320.5 & 28.9\\
Stanford Cars / SigLIP & T2 & 2 & 294 & 204 & 1.4 & 4.3 & 6.6 & 1.5\\
Stanford Cars / SigLIP & T6 & 6 & 885 & 236 & 3.8 & 4.8 & 19.3 & 4.0\\
Stanford Cars / SigLIP & T20 & 20 & 2{,}964 & 348 & 8.5 & 6.9 & 65.2 & 9.4\\
Stanford Cars / SigLIP & per-class & 99 & 14{,}553 & 980 & 14.9 & 14.4 & 340.2 & 23.6\\}
\pcpdef{tab.sup.classical}{%
raw & 0.2530 & 17.74 & 0.2237 & 15.20 & 0.2142 & 14.31 & 0.2521 & 30.13 & 0.3672 & 44.85\\
uniform PRE & 0.2414 & \textbf{13.73} & 0.2122 & \textbf{11.78} & 0.2049 & \textbf{11.32} & \textbf{0.1716} & \textbf{17.60} & \textbf{0.2293} & \textbf{30.09}\\
DN (Mondrian) & 0.2158 & 16.70 & 0.1907 & 14.44 & 0.1837 & 13.72 & 0.2439 & 34.91 & 0.3031 & 44.71\\
CSLS, top-5 & 0.2060 & 14.36 & 0.1832 & 12.49 & 0.1766 & 11.90 & 0.1965 & 22.34 & 0.2781 & 34.39\\
CSLS, top-10 & \textbf{0.2056} & 15.05 & \textbf{0.1831} & 13.09 & \textbf{0.1754} & 12.37 & 0.2008 & 23.96 & 0.2759 & 35.32\\
mean-centring & 0.2283 & 19.46 & 0.2042 & 17.05 & 0.2006 & 16.47 & 0.2842 & 48.62 & 0.3463 & 56.81\\
max-centring & 0.2861 & 15.02 & 0.2549 & 12.96 & 0.2462 & 12.41 & 0.1967 & 18.46 & 0.2759 & 31.40\\
\midrule
active HPS (full-vocab.\ oracle) & 0.1743 & 11.03 & 0.1499 & 9.31 & 0.1422 & 8.78 & 0.1523 & 15.24 & 0.2194 & 28.78\\}
\pcpdef{tab.sup.classical.cars}{%
raw & 0.2521 & 30.13 & 0.3672 & 44.85\\
uniform PRE & \textbf{0.1716} & \textbf{17.60} & \textbf{0.2293} & \textbf{30.09}\\
DN (Mondrian) & 0.2439 & 34.91 & 0.3031 & 44.71\\
CSLS, top-5 & 0.1965 & 22.34 & 0.2781 & 34.39\\
CSLS, top-10 & 0.2008 & 23.96 & 0.2759 & 35.32\\
mean-centring & 0.2842 & 48.62 & 0.3463 & 56.81\\
max-centring & 0.1967 & 18.46 & 0.2759 & 31.40\\
\midrule
active HPS (oracle) & 0.1523 & 15.24 & 0.2194 & 28.78\\}
\pcpdef{tab.sup.classical.food}{%
raw & 0.2530 & 17.74 & 0.2237 & 15.20 & 0.2142 & 14.31\\
uniform PRE & 0.2414 & \textbf{13.73} & 0.2122 & \textbf{11.78} & 0.2049 & \textbf{11.32}\\
DN (Mondrian) & 0.2158 & 16.70 & 0.1907 & 14.44 & 0.1837 & 13.72\\
CSLS, top-5 & 0.2060 & 14.36 & 0.1832 & 12.49 & 0.1766 & 11.90\\
CSLS, top-10 & \textbf{0.2056} & 15.05 & \textbf{0.1831} & 13.09 & \textbf{0.1754} & 12.37\\
mean-centring & 0.2283 & 19.46 & 0.2042 & 17.05 & 0.2006 & 16.47\\
max-centring & 0.2861 & 15.02 & 0.2549 & 12.96 & 0.2462 & 12.41\\
\midrule
active HPS (oracle) & 0.1743 & 11.03 & 0.1499 & 9.31 & 0.1422 & 8.78\\}
\pcpdef{tab.sup.closure}{%
Food-101 / SigLIP $m{=}20$ & 0.2414 & 0.1932 & 0.1743 & 71.8 & 17.510 & 13.996 & 10.975 & 53.8\\
Food-101 / SigLIP $m{=}40$ & 0.2122 & 0.1695 & 0.1499 & 68.6 & 15.135 & 12.103 & 9.262 & 51.6\\
Food-101 / SigLIP $m{=}80$ & 0.2049 & 0.1640 & 0.1422 & 65.3 & 14.607 & 11.679 & 8.756 & 50.0\\
Stanford Cars / SigLIP & 0.1716 & 0.1608 & 0.1523 & 55.8 & 20.282 & 17.815 & 15.071 & 47.3\\
Stanford Cars / CLIP & 0.2293 & 0.2241 & 0.2194 & 52.5 & 32.005 & 30.507 & 28.717 & 45.6\\
ImageNet-R / CLIP & 0.0735 & 0.0517 & 0.0394 & 64.0 & 10.051 & 7.295 & 4.800 & 52.5\\
ImageNet-R / SigLIP & 0.0385 & 0.0256 & 0.0156 & 56.6 & 5.085 & 3.631 & 1.840 & 44.8\\
CUB-200 / CLIP & 0.2426 & 0.2389 & 0.2374 & 71.3 & 32.276 & 30.900 & 29.546 & 50.4\\
CUB-200 / SigLIP & 0.2031 & 0.1981 & 0.1940 & 55.0 & 23.828 & 22.273 & 20.682 & 49.4\\
ImageNet-Sketch / CLIP & 0.0706 & 0.0541 & 0.0480 & 72.8 & 37.292 & 30.405 & 24.046 & 52.0\\
ImageNet-Sketch / SigLIP & 0.0398 & 0.0275 & 0.0216 & 67.6 & 19.873 & 14.677 & 8.747 & 46.7\\
\midrule
range / mean &  &  &  & \makecell[c]{52.5\pcpdash{}72.8\\(63.7)} &  &  &  & \makecell[c]{44.8\pcpdash{}53.8\\(49.5)}\\}
\pcpdef{tab.sup.confot.q1}{%
Food $m{=}20$ & 0.9305 & 7.82 & 4.72 & 9.54 & 9.55\\
Food $m{=}40$ & 0.9277 & 7.50 & 4.59 & 9.01 & 9.03\\
Food $m{=}80$ & 0.9290 & 7.63 & 4.63 & 9.23 & 9.35\\
Cars/SigLIP & 0.8974 & 13.21 & 8.30 & 16.56 & 15.73\\
Cars/CLIP & 0.9049 & 25.55 & 6.68 & 27.72 & 30.73\\}
\pcpdef{tab.sup.confot.q3}{%
Food $m{=}20$ & 7.36 & 49.0 & 0.149 & 13.3\\
Food $m{=}40$ & 6.42 & 50.7 & 0.123 & 11.1\\
Food $m{=}80$ & 6.23 & 51.6 & 0.102 & 9.3\\
Cars/SigLIP & 3.49 & 35.3 & 0.165 & 22.3\\
Cars/CLIP & 4.57 & 29.4 & 0.137 & 21.7\\}
\pcpdef{tab.sup.geo.clip}{%
\;raw cosine & 2.286 & 2.278 & 2.192 & 3.841 & 4.878 & 37.7 & 4.8 & 0.8971 & 0.8000 & 2.235 & 1.071\\
\;PRE & 1.871 & 1.865 & 1.693 & 3.342 & 4.134 & 47.4 & 2.7 & 0.8991 & 0.8003 & 2.537 & 0.835\\
\;\textbf{PrefixCP} & 1.643 & 1.632 & 1.575 & 2.506 & 2.812 & 50.2 & 2.4 & 0.9006 & 0.8053 & 2.040 & 0.607\\
\;HPS-Refresh & 1.268 & 1.258 & 1.195 & 1.800 & 2.028 & 74.7 & 2.7 & 0.9008 & 0.8109 & 1.248 & 0.607\\
\;HPS-Frozen$^{\ddagger}$ & 1.286 & 1.276 & 1.205 & 1.834 & 2.061 & 75.0 & 1.8 & 0.9137 & 0.8255 & 1.242 & 0.607\\}
\pcpdef{tab.sup.geo.siglip}{%
\;raw cosine & 1.397 & 1.400 & 1.219 & 2.312 & 2.815 & 61.9 & 7.0 & 0.8966 & 0.7956 & 1.395 & 0.659\\
\;PRE & 1.425 & 1.426 & 1.270 & 2.389 & 2.788 & 61.3 & 4.5 & 0.9042 & 0.8178 & 1.823 & 0.664\\
\;\textbf{PrefixCP} & 1.284 & 1.279 & 1.165 & 1.992 & 2.225 & 67.4 & 4.6 & 0.9056 & 0.8210 & 1.525 & 0.523\\
\;HPS-Refresh & 1.064 & 1.059 & 0.990 & 1.440 & 1.815 & 84.8 & 5.1 & 0.9071 & 0.8281 & 1.061 & 0.523\\
\;HPS-Frozen$^{\ddagger}$ & 1.081 & 1.078 & 1.018 & 1.435 & 1.746 & 86.9 & 3.1 & 0.9269 & 0.8418 & 1.051 & 0.523\\}
\pcpdef{tab.sup.geochurn}{%
CLIP ViT-B/16 & HPS-Frozen$^{\ddagger}$ & 2.46 & 37.96 & \textbf{40.14} & 0.35\\
CLIP ViT-B/16 & HPS-Refresh & 2.49 & 38.12 & \textbf{41.02} & 0.32\\
SigLIP B/16 & HPS-Frozen$^{\ddagger}$ & 1.51 & 26.69 & \textbf{31.82} & 0.26\\
SigLIP B/16 & HPS-Refresh & 1.47 & 25.82 & \textbf{32.30} & 0.19\\}
\pcpdef{tab.sup.geochurn.grouped}{%
\multicolumn{5}{@{}l}{\emph{CLIP ViT-B/16}}\\
\quad HPS-Frozen$^{\ddagger}$ & 2.46 & 37.96 & \textbf{40.14} & 0.35\\
\quad HPS-Refresh & 2.49 & 38.12 & \textbf{41.02} & 0.32\\
\multicolumn{5}{@{}l}{\emph{SigLIP B/16}}\\
\quad HPS-Frozen$^{\ddagger}$ & 1.51 & 26.69 & \textbf{31.82} & 0.26\\
\quad HPS-Refresh & 1.47 & 25.82 & \textbf{32.30} & 0.19\\}
\pcpdef{tab.sup.geoclosed}{%
CLIP ViT-B/16 & all-row $\mathbb E|\Gamma|$ & 1.871 & 1.643 & 1.268 & 37.9\\
CLIP ViT-B/16 & class-macro $\mathbb E|\Gamma|$ & 1.865 & 1.632 & 1.258 & 38.4\\
CLIP ViT-B/16 & novel-row $\mathbb E|\Gamma|$ & 2.537 & 2.040 & 1.248 & 38.5\\
CLIP ViT-B/16 & novel-candidate block & 0.835 & 0.607 & 0.607 & 100.0\\
SigLIP B/16 & all-row $\mathbb E|\Gamma|$ & 1.425 & 1.284 & 1.064 & 39.1\\
SigLIP B/16 & class-macro $\mathbb E|\Gamma|$ & 1.426 & 1.279 & 1.059 & 39.9\\
SigLIP B/16 & novel-row $\mathbb E|\Gamma|$ & 1.823 & 1.525 & 1.061 & 39.1\\
SigLIP B/16 & novel-candidate block & 0.664 & 0.523 & 0.523 & 100.0\\}
\pcpdef{tab.sup.georegion}{%
CLIP ViT-B/16 & PRE & 2.261 & 1.920 & 1.830 & 1.932 & 1.570 & 1.713\\
CLIP ViT-B/16 & \textbf{PrefixCP} & 1.930 & 1.648 & 1.609 & 1.692 & 1.374 & 1.602\\
CLIP ViT-B/16 & \emph{gain} & 0.330 & 0.272 & 0.220 & 0.240 & 0.197 & 0.111\\
SigLIP B/16 & PRE & 1.570 & 1.459 & 1.452 & 1.432 & 1.442 & 1.198\\
SigLIP B/16 & \textbf{PrefixCP} & 1.411 & 1.306 & 1.300 & 1.276 & 1.310 & 1.102\\
SigLIP B/16 & \emph{gain} & 0.159 & 0.153 & 0.152 & 0.156 & 0.132 & 0.095\\}
\pcpdef{tab.sup.m4080}{%
\multicolumn{5}{@{}l}{\emph{Base-class rows}}\\
raw cosine & 0.2257 & 15.251 & 0.2153 & 14.336\\
uniform PRE & 0.1431 & 8.359 & 0.1376 & 7.959\\
\textbf{PrefixCP} & 0.1470 & 8.813 & 0.1409 & 8.391\\
HPS-Refresh$^{\dagger}$ & 0.1529 & 9.354 & 0.1450 & 8.813\\
\midrule
\multicolumn{5}{@{}l}{\emph{Novel-class rows}}\\
raw cosine & 0.2237 & 15.149 & 0.2142 & 14.277\\
uniform PRE & 0.2122 & 15.135 & 0.2049 & 14.607\\
\textbf{PrefixCP} & 0.1695 & 12.103 & 0.1640 & 11.679\\
HPS-Refresh$^{\dagger}$ & 0.1499 & 9.262 & 0.1422 & 8.756\\
\midrule
\multicolumn{5}{@{}l}{\emph{All rows}}\\
raw cosine & 0.2331 & 15.199 & 0.2228 & 14.307\\
uniform PRE & 0.2122 & 11.781 & 0.2049 & 11.316\\
\textbf{PrefixCP} & 0.1711 & 10.475 & 0.1651 & 10.051\\
HPS-Refresh$^{\dagger}$ & 0.1620 & 9.308 & 0.1537 & 8.784\\
\midrule
\multicolumn{5}{@{}l}{\emph{Novel candidates per novel-class image}}\\
uniform PRE & \multicolumn{2}{c}{8.122} & \multicolumn{2}{c}{7.779}\\
\textbf{PrefixCP} $=$ HPS-Refresh$^{\dagger}$ & \multicolumn{2}{c}{5.090} & \multicolumn{2}{c}{4.851}\\
\emph{change} & \multicolumn{2}{c}{$-37.3\%$} & \multicolumn{2}{c}{$-37.6\%$}\\}
\pcpdef{tab.sup.oracle}{%
Food/SigLIP & 0.010597 & 0.011305 & 0.5351 & 0.4105\\
Cars/SigLIP & 0.014556 & 0.015872 & 0.6691 & 0.6302\\
Cars/CLIP & 0.005783 & 0.009911 & 0.9079 & 0.7796\\}
\pcpdef{tab.sup.paired}{%
Food-101 / SigLIP $m{=}20$ & +1.578 & [+1.073,\,+2.084] & 5/5 & -1.120 & [-1.663,\,-0.577] & 0/5 & -1.818 & [-2.188,\,-1.449] & 0/5\\
Food-101 / SigLIP $m{=}40$ & +1.306 & [+0.981,\,+1.632] & 5/5 & -1.167 & [-1.514,\,-0.820] & 0/5 & -1.642 & [-1.784,\,-1.500] & 0/5\\
Food-101 / SigLIP $m{=}80$ & +1.265 & [+0.993,\,+1.538] & 5/5 & -1.267 & [-1.530,\,-1.004] & 0/5 & -1.626 & [-1.766,\,-1.487] & 0/5\\
Stanford Cars / SigLIP & +1.092 & [+0.893,\,+1.291] & 5/5 & -1.264 & [-1.525,\,-1.002] & 0/5 & -1.423 & [-1.769,\,-1.077] & 0/5\\
Stanford Cars / CLIP & +0.535 & [+0.263,\,+0.807] & 5/5 & -0.775 & [-0.999,\,-0.550] & 0/5 & -0.534 & [-0.787,\,-0.281] & 0/5\\
ImageNet-R / CLIP & +1.200 & [+0.880,\,+1.520] & 5/5 & -0.971 & [-1.384,\,-0.559] & 0/5 & -1.292 & [-1.459,\,-1.125] & 0/5\\
ImageNet-R / SigLIP & +0.674 & [+0.422,\,+0.926] & 5/5 & -0.802 & [-1.257,\,-0.347] & 0/5 & -0.859 & [-1.058,\,-0.660] & 0/5\\
CUB-200 / CLIP & +0.529 & [+0.203,\,+0.856] & 5/5 & -0.487 & [-0.724,\,-0.250] & 0/5 & -0.872 & [-1.412,\,-0.331] & 0/5\\
CUB-200 / SigLIP & +0.617 & [+0.134,\,+1.099] & 4/5 & -0.694 & [-0.924,\,-0.464] & 0/5 & -0.766 & [-0.958,\,-0.574] & 0/5\\
ImageNet-Sketch / CLIP & +3.143 & [+2.709,\,+3.577] & 5/5 & -2.700 & [-3.391,\,-2.008] & 0/5 & -3.376 & [-3.738,\,-3.014] & 0/5\\
ImageNet-Sketch / SigLIP & +2.372 & [+1.888,\,+2.855] & 5/5 & -2.749 & [-3.335,\,-2.163] & 0/5 & -3.017 & [-3.340,\,-2.694] & 0/5\\}
\pcpdef{tab.sup.rbbreak}{%
1{,}000 & 2 & 1.17 & 1.18 & 1.45 & 1.59\\
1{,}000 & 10 & 2.06 & 3.66 & 4.65 & 4.12\\
1{,}000 & 50 & 2.16 & 7.70 & 15.39 & 15.25\\
1{,}000 & 100 & 2.03 & 11.30 & 22.87 & 22.66\\
1{,}000 & 500 & 2.05 & 12.40 & 57.34 & 64.53\\
10{,}000 & 2 & 1.47 & 1.51 & 1.51 & 1.49\\
10{,}000 & 10 & 2.65 & 5.02 & 4.85 & 4.76\\
10{,}000 & 50 & 4.04 & 18.45 & 18.62 & 18.48\\
10{,}000 & 100 & 4.83 & 29.79 & 64.93 & 67.61\\
10{,}000 & 500 & 5.54 & 65.48 & 272.81 & 209.43\\
100{,}000 & 2 & 1.53 & 1.53 & 1.54 & \emph{skipped}\\
100{,}000 & 10 & 5.08 & 5.41 & 5.45 & \emph{skipped}\\
100{,}000 & 50 & 16.19 & 21.74 & 52.87 & \emph{skipped}\\
100{,}000 & 100 & 25.71 & 47.77 & 96.22 & \emph{skipped}\\
100{,}000 & 500 & 43.87 & 189.75 & 370.49 & \emph{skipped}\\}
\pcpdef{tab.sup.rbcurve}{%
1{,}000 & 2 & 1.5 & 1.00$\times$ & 1.00$\times$ & 1.00\\
1{,}000 & 10 & 5.5 & 1.21$\times$ & 3.87$\times$ & 1.05\\
1{,}000 & 25 & 13.0 & 1.34$\times$ & 9.05$\times$ & 1.04\\
1{,}000 & 100 & 50.5 & 2.22$\times$ & 34.95$\times$ & 1.04\\
1{,}000 & 500 & 250.5 & 7.00$\times$ & 276.14$\times$ & 1.65\\
10{,}000 & 2 & 1.5 & 1.00$\times$ & 1.00$\times$ & 1.00\\
10{,}000 & 10 & 5.5 & 1.19$\times$ & 3.83$\times$ & 1.05\\
10{,}000 & 25 & 13.0 & 1.33$\times$ & 9.04$\times$ & 1.04\\
10{,}000 & 100 & 50.5 & 1.44$\times$ & 62.07$\times$ & 1.84\\
10{,}000 & 500 & 250.5 & 1.80$\times$ & 325.41$\times$ & 1.95\\
100{,}000 & 2 & 1.5 & 1.00$\times$ & 1.00$\times$ & 1.00\\
100{,}000 & 10 & 5.5 & 1.02$\times$ & 3.61$\times$ & 0.98\\
100{,}000 & 25 & 13.0 & 1.13$\times$ & 8.23$\times$ & 0.95\\
100{,}000 & 100 & 50.5 & 1.22$\times$ & 75.81$\times$ & 2.25\\
100{,}000 & 500 & 250.5 & 1.46$\times$ & 351.06$\times$ & 2.10\\}
\pcpdef{tab.sup.rbmem}{%
\multicolumn{3}{@{}l@{}}{\emph{$K{=}1{,}000$, batch 4{,}096, $\nver{=}50$}}\\
\;one active partition & 31 & 98\\
\;direct & 31 & 1{,}402\\
\;sequential (superseded) & 31 & 66\\
\;\textbf{blocked} & 31 & 9\\
\multicolumn{3}{@{}l@{}}{\emph{$K{=}10{,}000$, batch 1{,}024, $\nver{=}10$}}\\
\;one active partition & 78 & 244\\
\;direct & 78 & 1{,}300\\
\;sequential (superseded) & 78 & 166\\
\;\textbf{blocked} & 78 & 25\\
\multicolumn{3}{@{}l@{}}{\emph{$K{=}10{,}000$, batch 4{,}096, $\nver{=}100$}}\\
\;one active partition & 313 & 977\\
\;direct & 313 & 8{,}994\\
\;sequential (superseded) & 313 & 664\\
\;\textbf{blocked} & 313 & 52\\
\multicolumn{3}{@{}l@{}}{\emph{$K{=}100{,}000$, batch 1{,}024, $\nver{=}10$}}\\
\;one active partition & 781 & 2{,}441\\
\;direct & 781 & 10{,}229\\
\;sequential (superseded) & 781 & 1{,}660\\
\;\textbf{blocked} & 781 & 254\\
\multicolumn{3}{@{}l@{}}{\emph{$K{=}100{,}000$, batch 1{,}024, $\nver{=}100$}}\\
\;one active partition & 781 & 2{,}442\\
\;direct & 781 & 14{,}645\\
\;sequential (superseded) & 781 & 1{,}660\\
\;\textbf{blocked} & 781 & 122\\}
\pcpdef{tab.sup.rbreal}{%
Food-101 & 101 & 2 & 25{,}250 & 10.26\,ms & 6.28\,ms & 10.53\,ms & 55.85\,ms & \textbf{7.64\,ms} & 1.22 & 1.38 & 0.74\\
Food-101 & 101 & 6 & 25{,}250 & 10.26\,ms & 6.28\,ms & 28.87\,ms & 56.06\,ms & \textbf{12.28\,ms} & 1.96 & 2.35 & 1.20\\
Stanford Cars & 196 & 2 & 8{,}041 & 5.69\,ms & 3.86\,ms & 6.18\,ms & 33.74\,ms & \textbf{4.40\,ms} & 1.14 & 1.40 & 0.77\\
Stanford Cars & 196 & 6 & 8{,}041 & 5.69\,ms & 3.86\,ms & 16.44\,ms & 33.77\,ms & \textbf{5.60\,ms} & 1.45 & 2.93 & 0.98\\
ImageNet-Sketch & 930 & 2 & 24{,}003 & 70.68\,ms & 52.38\,ms & 85.47\,ms & 494.20\,ms & \textbf{58.18\,ms} & 1.11 & 1.47 & 0.82\\
ImageNet-Sketch & 930 & 6 & 24{,}003 & 70.68\,ms & 52.38\,ms & 226.08\,ms & 494.40\,ms & \textbf{74.95\,ms} & 1.43 & 3.02 & 1.06\\}
\pcpdef{tab.sup.rbskew}{%
1 & 54.88\,$\mu$s & 610.92\,$\mu$s & 123.17\,$\mu$s & 4.96$\times$ & 2.24$\times$\\
32 & 954.75\,$\mu$s & 16.95\,ms & 1.04\,ms & 16.22$\times$ & 1.09$\times$\\
1{,}024 & 29.00\,ms & 540.47\,ms & 31.18\,ms & 17.34$\times$ & 1.08$\times$\\
4{,}096 & 112.37\,ms & 2.17\,s & 121.62\,ms & 17.84$\times$ & 1.08$\times$\\}
\pcpdef{tab.sup.rbstack}{%
Food-101 & 101 & 2 & 1.50 & 1.45 & 1.50\\
Food-101 & 101 & 6 & 3.75 & 2.99 & 3.75\\
Stanford Cars & 196 & 2 & 1.50 & 1.46 & 1.50\\
Stanford Cars & 196 & 6 & 3.75 & 3.35 & 3.75\\
ImageNet-Sketch & 930 & 2 & 1.50 & 1.49 & 1.50\\
ImageNet-Sketch & 930 & 6 & 3.75 & 3.37 & 3.75\\}
\pcpdef{tab.sup.rbsyn}{%
\multicolumn{5}{@{}l@{}}{\emph{1 BLAS thread}}\\
\;one active partition (the floor) & 3.17\,ms & 29.00\,ms & 263.95\,ms & 263.95\,ms\\
\;direct historical reductions & 19.64\,ms & 807.73\,ms & 31.22\,s & 144.57\,s\\
\;sequential PrefixCP (superseded) & 22.62\,ms & 181.30\,ms & 1.48\,s & 1.48\,s\\
\;\textbf{blocked PrefixCP} & 4.22\,ms & 43.37\,ms & 324.43\,ms & 390.22\,ms\\
\;dispatch baseline (names blocked) & 4.04\,ms & 44.91\,ms & 368.60\,ms & 381.13\,ms\\
\;direct, validation-free & 17.96\,ms & 731.30\,ms & 28.80\,s & 146.30\,s\\
\;blocked, validation-free & 4.05\,ms & 42.13\,ms & 299.48\,ms & 331.48\,ms\\
\;\textbf{blocked / one active partition} & 1.33$\times$ & 1.50$\times$ & 1.23$\times$ & 1.48$\times$\\
\;direct / blocked & 4.65$\times$ & 18.62$\times$ & 96.22$\times$ & 370.49$\times$\\
\;sequential / blocked & 5.36$\times$ & 4.18$\times$ & 4.55$\times$ & 3.79$\times$\\
\midrule
\multicolumn{5}{@{}l@{}}{\emph{8 BLAS threads}}\\
\;one active partition (the floor) & 2.45\,ms & 23.45\,ms & 231.92\,ms & 231.92\,ms\\
\;direct historical reductions & 14.92\,ms & 640.50\,ms & 24.67\,s & 130.85\,s\\
\;sequential PrefixCP (superseded) & 19.98\,ms & 162.85\,ms & 1.36\,s & 1.36\,s\\
\;\textbf{blocked PrefixCP} & 2.86\,ms & 32.09\,ms & 288.74\,ms & 316.36\,ms\\
\;dispatch baseline (names blocked) & 2.86\,ms & 31.97\,ms & 289.55\,ms & 313.15\,ms\\
\;direct, validation-free & 13.69\,ms & 593.92\,ms & 26.62\,s & 128.73\,s\\
\;blocked, validation-free & 2.89\,ms & 30.72\,ms & 316.99\,ms & 299.66\,ms\\
\;\textbf{blocked / one active partition} & 1.17$\times$ & 1.37$\times$ & 1.24$\times$ & 1.36$\times$\\
\;direct / blocked & 5.22$\times$ & 19.96$\times$ & 85.44$\times$ & 413.60$\times$\\
\;sequential / blocked & 6.99$\times$ & 5.07$\times$ & 4.71$\times$ & 4.31$\times$\\}
\pcpdef{tab.sup.rc3p}{%
\texttt{rc3p\_default} ($t_{\mathrm{gap}}{=}0.9$) & 1.0000 / 101.00 & 1.0000 / 101.00 & 1.0000 / 101.00 & 1.0000 / 196.00 & 1.0000 / 196.00\\
\texttt{rc3p\_gap0} (sensitivity) & 0.3330 / 21.62 & 0.3322 / 21.70 & 0.1988 / 11.66 & 0.3192 / 30.06 & 0.4029 / 49.04\\
classwise APS (author) & 0.2336 / 14.24 & 0.2073 / 12.29 & 0.1988 / 11.66 & 0.2383 / 20.87 & 0.2988 / 34.36\\
\midrule
our active HPS $+$ randomized ranks & 0.1743 / 11.03 & 0.1499 / 9.31 & 0.1422 / 8.78 & 0.1523 / 15.24 & 0.2194 / 28.78\\
our uniform PRE (stable) & 0.2414 / 13.73 & 0.2122 / 11.78 & 0.2049 / 11.32 & 0.1716 / 17.60 & 0.2293 / 30.09\\}
\pcpdef{tab.sup.refit}{%
Food-101 / SigLIP $m{=}20$ & 50 & 1{,}000 & 50 & 0.005 & 0.005\\
Food-101 / SigLIP $m{=}40$ & 50 & 2{,}000 & 50 & 0.010 & 0.009\\
Food-101 / SigLIP $m{=}80$ & 50 & 4{,}000 & 50 & 0.020 & 0.018\\
Stanford Cars / SigLIP & 98 & 1{,}960 & 98 & 0.011 & 0.010\\
Stanford Cars / CLIP & 98 & 1{,}960 & 98 & 0.011 & 0.010\\
ImageNet-R / CLIP & 100 & 2{,}000 & 100 & 0.013 & 0.011\\
ImageNet-R / SigLIP & 100 & 2{,}000 & 100 & 0.012 & 0.011\\
CUB-200 / CLIP & 100 & 2{,}000 & 100 & 0.012 & 0.010\\
CUB-200 / SigLIP & 100 & 2{,}000 & 100 & 0.012 & 0.011\\
ImageNet-Sketch / CLIP & 465 & 9{,}300 & 465 & 0.327 & 0.080\\
ImageNet-Sketch / SigLIP & 465 & 9{,}300 & 465 & 0.104 & 0.097\\}
\pcpdef{tab.sup.rowgroup.all}{%
raw cosine & 0.2631 & 17.736 & 0.2695 & 30.133 & 0.3840 & 44.854 & 0.1155 & 14.159 & 0.0504 & 5.674\\
uniform PRE & 0.2414 & 13.727 & 0.1754 & 17.598 & 0.2360 & 30.090 & 0.0735 & 7.033 & 0.0385 & 3.300\\
\textbf{PrefixCP} & 0.1944 & 12.148 & 0.1687 & 16.506 & 0.2342 & 29.555 & 0.0519 & 5.833 & 0.0256 & 2.626\\
HPS-Refresh$^{\dagger}$ & 0.1882 & 11.029 & 0.1649 & 15.242 & 0.2340 & 28.780 & 0.0434 & 4.861 & 0.0168 & 1.824\\}
\pcpdef{tab.sup.rowgroup.base}{%
raw cosine & 0.2553 & 17.782 & 0.2568 & 30.301 & 0.3571 & 45.135 & 0.1100 & 14.298 & 0.0472 & 5.680\\
uniform PRE & 0.1662 & 9.867 & 0.1486 & 14.915 & 0.2171 & 28.174 & 0.0360 & 4.129 & 0.0136 & 1.588\\
\textbf{PrefixCP} & 0.1684 & 10.263 & 0.1502 & 15.198 & 0.2202 & 28.602 & 0.0375 & 4.430 & 0.0144 & 1.664\\
HPS-Refresh$^{\dagger}$ & 0.1785 & 11.083 & 0.1513 & 15.416 & 0.2222 & 28.842 & 0.0404 & 4.921 & 0.0155 & 1.810\\
\midrule
PrefixCP $-$ PRE & $+0.0022$ & $+0.396$ & $+0.0016$ & $+0.283$ & $+0.0031$ & $+0.428$ & $+0.0016$ & $+0.301$ & $+0.0008$ & $+0.076$\\}
\pcpdef{tab.sup.rvf}{%
Food-101 / SigLIP $m{=}20$ & +0.698 & [+0.364,\,+1.033] & 0.7088 & 0.7448 & \textbf{0.7424} & 0.9239 & 0.9283\\
Food-101 / SigLIP $m{=}40$ & +0.475 & [+0.191,\,+0.760] & 0.7488 & 0.7976 & \textbf{0.8046} & 0.9214 & 0.9263\\
Food-101 / SigLIP $m{=}80$ & +0.359 & [+0.229,\,+0.489] & 0.7887 & 0.8333 & \textbf{0.8428} & 0.9241 & 0.9289\\
Stanford Cars / SigLIP & +0.159 & [-0.045,\,+0.363] & 0.5790 & 0.6258 & \textbf{0.6308} & 0.8983 & 0.8987\\
Stanford Cars / CLIP & -0.241 & [-0.562,\,+0.081] & 0.6304 & 0.6640 & \textbf{0.6539} & 0.9035 & 0.9031\\
ImageNet-R / CLIP & +0.321 & [-0.015,\,+0.656] & 0.5996 & 0.5714 & \textbf{0.5929} & 0.9027 & 0.9039\\
ImageNet-R / SigLIP & +0.057 & [-0.211,\,+0.326] & 0.6349 & 0.6460 & \textbf{0.6500} & 0.9127 & 0.9025\\
CUB-200 / CLIP & +0.384 & [+0.047,\,+0.722] & 0.5634 & 0.6067 & \textbf{0.6108} & 0.8922 & 0.8969\\
CUB-200 / SigLIP & +0.072 & [-0.108,\,+0.252] & 0.5533 & 0.5800 & \textbf{0.5975} & 0.8943 & 0.8960\\
ImageNet-Sketch / CLIP & +0.676 & [+0.276,\,+1.077] & 0.3077 & 0.5043 & \textbf{0.4833} & 0.9010 & 0.9045\\
ImageNet-Sketch / SigLIP & +0.269 & [-0.064,\,+0.602] & 0.1528 & 0.5160 & \textbf{0.5314} & 0.9054 & 0.9032\\}
\pcpdef{tab.sup.scores}{%
\multicolumn{9}{@{}l@{}}{\emph{Food-101 / SigLIP} ($50\to101$ labels)}\\
\;LAC (sum-form HPS), frozen & 50.12 & 0.00 & 6.72 & 66.9 & 0.9256$\to$0.8737 & 0.7640$\to$0.5624 & 0.8611$\to$0.7723 & 6.74$\to$3.38\\
\;LAC ($1-p$), frozen & 50.12 & 0.00 & 6.72 & 66.9 & 0.9256$\to$0.8737 & 0.7640$\to$0.5624 & 0.8611$\to$0.7723 & 6.74$\to$3.38\\
\;LAC (mean-form HPS), frozen & 32.46 & 0.84 & 5.08 & 64.8 & 0.9256$\to$0.9199 & 0.7640$\to$0.7360 & 0.8611$\to$0.8484 & 6.74$\to$4.93\\
\;LAC (mean-form HPS), \emph{refresh} & 31.83 & 2.37 & 6.34 & 79.3 & 0.9256$\to$0.9287 & 0.7640$\to$0.7678 & 0.8611$\to$0.8507 & 6.74$\to$5.62\\
\;APS (randomized), frozen & 35.29 & 1.29 & 7.63 & 72.3 & 0.9181$\to$0.9157 & 0.7570$\to$0.7500 & 0.8446$\to$0.8381 & 9.35$\to$6.58\\
\;APS (randomized), \emph{refresh} & 36.21 & 2.62 & 8.88 & 81.4 & 0.9181$\to$0.9238 & 0.7570$\to$0.7720 & 0.8446$\to$0.8469 & 9.35$\to$7.04\\
\;APS (deterministic), frozen & 23.34 & 4.12 & 12.58 & 88.0 & 0.8964$\to$0.9535 & 0.6819$\to$0.7686 & 0.7878$\to$0.8828 & 22.11$\to$18.11\\
\;RAPS, $\lambda{=}0.01$, frozen & 51.93 & 0.23 & 10.22 & 96.9 & 0.9214$\to$0.8959 & 0.7530$\to$0.7348 & 0.8372$\to$0.8256 & 9.65$\to$4.73\\
\;RAPS, $\lambda{=}0.01$, \emph{refresh} & 20.62 & 2.49 & 5.92 & 90.8 & 0.9214$\to$0.9257 & 0.7530$\to$0.7644 & 0.8372$\to$0.8535 & 9.65$\to$8.70\\
\;RAPS, $\lambda{=}0.1$, frozen & 55.57 & 0.11 & 11.09 & 99.3 & 0.9240$\to$0.8815 & 0.7491$\to$0.6868 & 0.8511$\to$0.8046 & 9.90$\to$4.44\\
\midrule
\multicolumn{9}{@{}l@{}}{\emph{Stanford Cars / SigLIP} ($98\to196$ labels)}\\
\;LAC (sum-form HPS), frozen & 36.50 & 0.00 & 3.39 & 62.7 & 0.8989$\to$0.8388 & 0.6426$\to$0.3414 & 0.8005$\to$0.7093 & 9.11$\to$5.78\\
\;LAC ($1-p$), frozen & 36.50 & 0.00 & 3.39 & 62.7 & 0.8989$\to$0.8388 & 0.6426$\to$0.3414 & 0.8005$\to$0.7093 & 9.11$\to$5.78\\
\;LAC (mean-form HPS), frozen & 19.74 & 0.43 & 2.23 & 59.9 & 0.8989$\to$0.8988 & 0.6426$\to$0.5790 & 0.8005$\to$0.7910 & 9.11$\to$7.68\\
\;LAC (mean-form HPS), \emph{refresh} & 20.86 & 0.72 & 2.60 & 67.7 & 0.8989$\to$0.8987 & 0.6426$\to$0.6286 & 0.8005$\to$0.7933 & 9.11$\to$7.84\\
\;APS (randomized), frozen & 18.77 & 0.57 & 2.80 & 63.8 & 0.8939$\to$0.8925 & 0.6085$\to$0.5589 & 0.7811$\to$0.7800 & 11.97$\to$10.21\\
\;APS (randomized), \emph{refresh} & 20.70 & 1.02 & 3.43 & 69.7 & 0.8939$\to$0.8959 & 0.6085$\to$0.6521 & 0.7811$\to$0.7818 & 11.97$\to$10.37\\
\;APS (deterministic), frozen & 16.69 & 1.31 & 5.77 & 81.2 & 0.8993$\to$0.9464 & 0.6374$\to$0.7142 & 0.7880$\to$0.8546 & 28.47$\to$24.64\\
\;RAPS, $\lambda{=}0.01$, frozen & 49.39 & 0.06 & 11.12 & 100.0 & 0.8938$\to$0.8448 & 0.6225$\to$0.5665 & 0.7826$\to$0.7276 & 21.96$\to$11.16\\
\;RAPS, $\lambda{=}0.01$, \emph{refresh} & 7.49 & 1.79 & 3.06 & 92.5 & 0.8938$\to$0.8962 & 0.6225$\to$0.6556 & 0.7826$\to$0.7902 & 21.96$\to$21.67\\
\;RAPS, $\lambda{=}0.1$, frozen & 51.88 & 0.03 & 12.15 & 100.0 & 0.8931$\to$0.8312 & 0.6411$\to$0.5473 & 0.7853$\to$0.7101 & 22.91$\to$11.05\\
\midrule
\multicolumn{9}{@{}l@{}}{\emph{Stanford Cars / CLIP} ($98\to196$ labels)}\\
\;LAC (sum-form HPS), frozen & 34.57 & 0.00 & 5.41 & 85.1 & 0.9014$\to$0.8252 & 0.6539$\to$0.3291 & 0.8023$\to$0.7024 & 15.36$\to$10.05\\
\;LAC ($1-p$), frozen & 34.57 & 0.00 & 5.41 & 85.1 & 0.9014$\to$0.8252 & 0.6539$\to$0.3291 & 0.8023$\to$0.7024 & 15.36$\to$10.05\\
\;LAC (mean-form HPS), frozen & 8.48 & 0.93 & 2.11 & 67.3 & 0.9014$\to$0.9039 & 0.6539$\to$0.6304 & 0.8023$\to$0.8090 & 15.36$\to$14.82\\
\;LAC (mean-form HPS), \emph{refresh} & 10.50 & 1.01 & 2.50 & 77.1 & 0.9014$\to$0.9023 & 0.6539$\to$0.6645 & 0.8023$\to$0.8049 & 15.36$\to$14.58\\
\;APS (randomized), frozen & 9.10 & 1.24 & 2.70 & 77.3 & 0.8984$\to$0.9087 & 0.6452$\to$0.6763 & 0.7863$\to$0.8111 & 18.16$\to$17.50\\
\;APS (randomized), \emph{refresh} & 13.35 & 1.38 & 3.60 & 85.4 & 0.8984$\to$0.8979 & 0.6452$\to$0.6641 & 0.7863$\to$0.7923 & 18.16$\to$16.84\\
\;APS (deterministic), frozen & 7.68 & 1.70 & 3.14 & 83.2 & 0.8976$\to$0.9312 & 0.6464$\to$0.7032 & 0.7976$\to$0.8425 & 23.55$\to$23.01\\
\;RAPS, $\lambda{=}0.01$, frozen & 45.68 & 0.11 & 12.10 & 100.0 & 0.8986$\to$0.8494 & 0.6505$\to$0.4825 & 0.7927$\to$0.7204 & 25.74$\to$14.04\\
\;RAPS, $\lambda{=}0.01$, \emph{refresh} & 6.74 & 2.38 & 3.51 & 96.0 & 0.8986$\to$0.8999 & 0.6505$\to$0.6360 & 0.7927$\to$0.7948 & 25.74$\to$25.73\\
\;RAPS, $\lambda{=}0.1$, frozen & 51.72 & 0.03 & 14.77 & 100.0 & 0.9012$\to$0.8228 & 0.6215$\to$0.4139 & 0.7897$\to$0.6917 & 27.94$\to$13.50\\}
\pcpdef{tab.sup.scrvf}{%
Food-101 / SigLIP & LAC (mean-form HPS) & 2.85 & 60.5 & 9{,}179 & 26{,}801 & 4.93$\to$5.62 & 49.8/50 & 0.4036\\
Food-101 / SigLIP & APS (randomized) & 3.28 & 62.6 & 14{,}955 & 26{,}466 & 6.58$\to$7.04 & 50.0/50 & 0.0336\\
Food-101 / SigLIP & RAPS, $\lambda{=}0.01$ & 8.31 & 98.6 & 2{,}331 & 102{,}503 & 4.73$\to$8.70 & 50.0/50 & 0.0911\\
Stanford Cars / SigLIP & LAC (mean-form HPS) & 0.95 & 42.4 & 3{,}089 & 4{,}360 & 7.68$\to$7.84 & 96.6/98 & 0.3399\\
Stanford Cars / SigLIP & APS (randomized) & 1.36 & 45.2 & 4{,}729 & 5{,}960 & 10.21$\to$10.37 & 96.4/98 & 0.0208\\
Stanford Cars / SigLIP & RAPS, $\lambda{=}0.01$ & 10.83 & 100.0 & 398 & 84{,}610 & 11.16$\to$21.67 & 96.2/98 & 0.1832\\
Stanford Cars / CLIP & LAC (mean-form HPS) & 1.24 & 59.2 & 5{,}849 & 3{,}903 & 14.82$\to$14.58 & 97.6/98 & 0.1810\\
Stanford Cars / CLIP & APS (randomized) & 2.11 & 72.5 & 10{,}918 & 5{,}654 & 17.50$\to$16.84 & 98.0/98 & 0.0354\\
Stanford Cars / CLIP & RAPS, $\lambda{=}0.01$ & 12.18 & 100.0 & 980 & 94{,}580 & 14.04$\to$25.73 & 98.0/98 & 0.1943\\}
\pcpdef{tab.sup.seqdetail}{%
old-candidate bits under audit & 2{,}169{,}264 & 2{,}169{,}264 & 1{,}337{,}169 & 1{,}337{,}169\\
cumulative revocations & 104{,}244 & 110{,}439 & 27{,}570 & 30{,}064\\
cumulative additions & 16{,}215 & 32{,}458 & 4{,}998 & 10{,}016\\
revocation rate among onboarded positives & 33.1 & 35.0 & 22.6 & 24.5\\
addition rate among onboarded negatives & 0.87 & 1.75 & 0.41 & 0.82\\
path-cumulative flips (summed step to step) & 158{,}220 & 236{,}238 & 43{,}633 & 65{,}663\\
\;\emph{as a rate} & 7.29 & 10.89 & 3.26 & 4.91\\
images with $\ge1$ old-label change & 86.6 & 92.2 & 81.2 & 85.3\\
old-class images with $\ge1$ old-label change & 85.6 & 91.3 & 80.0 & 84.1\\
max per-class flip rate & 26.4 & 32.0 & 15.2 & 17.3\\
old-class macro cov.\ at onboarding & 0.9287 & 0.9287 & 0.9008 & 0.9008\\
old-class macro cov.\ at $V_6$ & 0.9238 & 0.9272 & 0.8967 & 0.8989\\
worst old-class cov.\ at $V_6$ & 0.7588 & 0.7536 & 0.4642 & 0.6258\\
\emph{all-class} macro cov.\ at $V_6$ & 0.9255 & 0.9284 & 0.8967 & 0.8986\\
\midrule
\emph{raw / PRE / \textbf{PrefixCP}}, every row above & \textbf{0} & \textbf{0} & \textbf{0} & \textbf{0}\\
worst / macro old-class / \emph{all-class} macro cov.\ at $V_6$, raw cosine & \multicolumn{2}{c}{0.7600 / 0.9318 / 0.9334} & \multicolumn{2}{c}{0.6282 / 0.8963 / 0.8972}\\
worst / macro old-class / \emph{all-class} macro cov.\ at $V_6$, uniform PRE & \multicolumn{2}{c}{0.7459 / 0.9270 / 0.9273} & \multicolumn{2}{c}{0.6360 / 0.8995 / 0.9000}\\
worst / macro old-class / \emph{all-class} macro cov.\ at $V_6$, \textbf{PrefixCP} & \multicolumn{2}{c}{0.7642 / 0.9287 / 0.9296} & \multicolumn{2}{c}{0.6301 / 0.9008 / 0.9002}\\}
\pcpdef{tab.sup.seqeff}{%
\multicolumn{11}{@{}l@{}}{\emph{Food-101 / SigLIP}}\\
1 & 25 & 25 & 2.848 & 2.848 & 2.848 & 2.848 & 4.195 & 4.195 & 4.195 & 4.195\\
2 & 40 & 15 & 8.398 & 6.837 & 4.542 & 4.390 & 7.289 & 6.878 & 6.337 & 6.141\\
3 & 56 & 16 & 11.775 & 9.993 & 5.834 & 5.434 & 9.765 & 8.729 & 7.599 & 7.076\\
4 & 71 & 15 & 14.588 & 12.254 & 7.309 & 6.763 & 12.142 & 10.494 & 8.816 & 8.094\\
5 & 86 & 15 & 17.774 & 15.232 & 9.588 & 8.966 & 14.972 & 12.478 & 10.098 & 9.489\\
6 & 101 & 15 & 19.590 & 17.228 & 11.278 & 10.897 & 17.425 & 14.122 & 11.029 & 10.632\\
\midrule
\multicolumn{11}{@{}l@{}}{\emph{Stanford Cars / SigLIP}}\\
1 & 49 & 49 & 4.069 & 4.069 & 4.069 & 4.069 & 5.038 & 5.038 & 5.038 & 5.038\\
2 & 78 & 29 & 9.685 & 8.370 & 6.136 & 6.077 & 8.104 & 7.763 & 7.174 & 7.131\\
3 & 108 & 30 & 12.762 & 11.192 & 8.105 & 7.934 & 11.400 & 10.481 & 9.345 & 9.180\\
4 & 137 & 29 & 16.726 & 14.835 & 11.233 & 11.078 & 14.517 & 12.915 & 11.292 & 11.123\\
5 & 167 & 30 & 18.834 & 16.570 & 11.928 & 11.789 & 17.684 & 15.309 & 13.123 & 12.988\\
6 & 196 & 29 & 22.665 & 19.906 & 15.186 & 14.872 & 21.014 & 17.753 & 15.242 & 14.933\\}
\pcpdef{tab.sup.seqfrontier}{%
\multicolumn{9}{@{}l@{}}{\emph{Food-101 / SigLIP}}\\
1 & 25 & 4.195 & \textbf{4.195} & 4.195 & 4.195 & +0.000 & +0.000\,[+0.000,+0.000] & ---\\
2 & 40 & 7.289 & \textbf{6.878} & 6.141 & 6.337 & +0.195 & -0.541\,[-0.760,-0.323] & 43.2\\
3 & 56 & 9.765 & \textbf{8.729} & 7.076 & 7.599 & +0.522 & -1.131\,[-1.462,-0.800] & 47.8\\
4 & 71 & 12.142 & \textbf{10.494} & 8.094 & 8.816 & +0.722 & -1.678\,[-2.125,-1.231] & 49.5\\
5 & 86 & 14.972 & \textbf{12.478} & 9.489 & 10.098 & +0.609 & -2.379\,[-2.943,-1.816] & 51.2\\
6 & 101 & 17.425 & \textbf{14.122} & 10.632 & 11.029 & +0.396 & -3.094\,[-3.842,-2.346] & 51.6\\
\midrule
\multicolumn{9}{@{}l@{}}{\emph{Stanford Cars / SigLIP}}\\
1 & 49 & 5.038 & \textbf{5.038} & 5.038 & 5.038 & +0.000 & +0.000\,[+0.000,+0.000] & ---\\
2 & 78 & 8.104 & \textbf{7.763} & 7.131 & 7.174 & +0.043 & -0.589\,[-0.896,-0.282] & 36.7\\
3 & 108 & 11.400 & \textbf{10.481} & 9.180 & 9.345 & +0.165 & -1.136\,[-1.671,-0.600] & 44.7\\
4 & 137 & 14.517 & \textbf{12.915} & 11.123 & 11.292 & +0.169 & -1.624\,[-2.112,-1.135] & 49.7\\
5 & 167 & 17.684 & \textbf{15.309} & 12.988 & 13.123 & +0.135 & -2.186\,[-2.772,-1.600] & 52.1\\
6 & 196 & 21.014 & \textbf{17.753} & 14.933 & 15.242 & +0.310 & -2.510\,[-3.168,-1.853] & 56.5\\}
\pcpdef{tab.sup.seqstab}{%
\multicolumn{8}{@{}l@{}}{\emph{Food-101 / SigLIP} ($K{=}101$)}\\
2 & 40 & \;HPS-Frozen$^{\ddagger}$ & 23{,}992 & 5{,}426 & 4.66 & 48.9 & 0.9200\\
2 & 40 & \;HPS-Refresh & 26{,}491 & 12{,}824 & 6.23 & 66.1 & 0.9223\\
3 & 56 & \;HPS-Frozen$^{\ddagger}$ & 50{,}310 & 8{,}581 & 5.84 & 68.3 & 0.9222\\
3 & 56 & \;HPS-Refresh & 51{,}750 & 23{,}292 & 7.44 & 82.9 & 0.9312\\
4 & 71 & \;HPS-Frozen$^{\ddagger}$ & 71{,}332 & 10{,}770 & 5.81 & 77.2 & 0.9195\\
4 & 71 & \;HPS-Refresh & 72{,}109 & 29{,}798 & 7.21 & 88.1 & 0.9281\\
5 & 86 & \;HPS-Frozen$^{\ddagger}$ & 88{,}579 & 13{,}187 & 5.68 & 82.6 & 0.9202\\
5 & 86 & \;HPS-Refresh & 91{,}512 & 31{,}600 & 6.87 & 90.7 & 0.9255\\
6 & 101 & \;HPS-Frozen$^{\ddagger}$ & 104{,}244 & 16{,}215 & 5.55 & 86.6 & 0.9238\\
6 & 101 & \;HPS-Refresh & 110{,}439 & 32{,}458 & 6.59 & 92.2 & 0.9272\\
\multicolumn{2}{@{}l}{all $t$} & \;raw / PRE / \textbf{PrefixCP} & \textbf{0} & \textbf{0} & \textbf{0.00} & \textbf{0.0} & ---\\
\midrule
\multicolumn{8}{@{}l@{}}{\emph{Stanford Cars / SigLIP} ($K{=}196$)}\\
2 & 78 & \;HPS-Frozen$^{\ddagger}$ & 6{,}948 & 1{,}879 & 2.25 & 45.6 & 0.9013\\
2 & 78 & \;HPS-Refresh & 7{,}929 & 3{,}244 & 2.85 & 56.5 & 0.9012\\
3 & 108 & \;HPS-Frozen$^{\ddagger}$ & 13{,}248 & 2{,}807 & 2.57 & 62.8 & 0.8986\\
3 & 108 & \;HPS-Refresh & 14{,}223 & 5{,}161 & 3.10 & 71.3 & 0.8959\\
4 & 137 & \;HPS-Frozen$^{\ddagger}$ & 18{,}125 & 3{,}760 & 2.53 & 71.7 & 0.8986\\
4 & 137 & \;HPS-Refresh & 19{,}750 & 6{,}793 & 3.07 & 78.1 & 0.9012\\
5 & 167 & \;HPS-Frozen$^{\ddagger}$ & 23{,}212 & 4{,}626 & 2.54 & 78.0 & 0.8970\\
5 & 167 & \;HPS-Refresh & 25{,}527 & 8{,}071 & 3.06 & 82.6 & 0.8970\\
6 & 196 & \;HPS-Frozen$^{\ddagger}$ & 27{,}570 & 4{,}998 & 2.44 & 81.2 & 0.8967\\
6 & 196 & \;HPS-Refresh & 30{,}064 & 10{,}016 & 3.00 & 85.3 & 0.8989\\
\multicolumn{2}{@{}l}{all $t$} & \;raw / PRE / \textbf{PrefixCP} & \textbf{0} & \textbf{0} & \textbf{0.00} & \textbf{0.0} & ---\\}
\pcpdef{tab.sup.tau}{%
Food-101 / SigLIP $m{=}20$ & 0.8 & 7.652 & 6.353 & \textbf{5.709} & 4.977 & 4.812$^{\ddagger}$ & +0.165 & 46.8 & 0.8445 & 0.8447\\
Food-101 / SigLIP $m{=}20$ & 0.9 & 17.736 & 13.727 & \textbf{12.148} & 11.029 & 10.330$^{\ddagger}$ & +0.698 & 58.5 & 0.9267 & 0.9283\\
Food-101 / SigLIP $m{=}20$ & 0.95 & 34.267 & 24.975 & \textbf{22.788} & 20.241 & 19.854$^{\ddagger}$ & +0.387 & 46.2 & 0.9651 & 0.9651\\
Stanford Cars / CLIP & 0.8 & 24.861 & 17.600 & \textbf{17.102} & 16.582 & 16.747$^{\ddagger}$ & -0.165 & 48.9 & 0.8023 & 0.8015\\
Stanford Cars / CLIP & 0.9 & 44.854 & 30.090 & \textbf{29.555} & 28.780 & 29.021$^{\ddagger}$ & -0.241 & 40.9 & 0.9023 & 0.9031\\
Stanford Cars / CLIP & 0.95 & 65.465 & 45.928 & \textbf{45.117} & 44.103 & 44.373$^{\ddagger}$ & -0.270 & 44.5 & 0.9534 & 0.9527\\}
\pcpdef{tab.sup.template}{%
Food/SigLIP & novel tail & 71.8 & 71.8 & 72.4 & 68.9\\
 & novel size & 53.8 & 53.8 & 53.7 & 53.9\\
Cars/SigLIP & novel tail & 55.8 & 55.8 & 55.8 & 58.9\\
 & novel size & 47.3 & 47.3 & 47.4 & 46.9\\
Cars/CLIP & novel tail & 52.5 & 52.5 & 57.1 & 52.8\\
 & novel size & 45.6 & 45.6 & 48.3 & 46.5\\}
\pcpdef{tab.sup.topq}{%
Food-101 / SigLIP $m{=}20$ & 0.2450 & 0.1957 & 0.1778 & 0.1670 & 0.2414 & 0.1932 & 0.1743 & 0.1637 & 0.2286 & 0.1840 & 0.1625 & 0.1527\\
Food-101 / SigLIP $m{=}40$ & 0.2156 & 0.1723 & 0.1534 & 0.1459 & 0.2122 & 0.1695 & 0.1499 & 0.1427 & 0.2001 & 0.1608 & 0.1392 & 0.1323\\
Food-101 / SigLIP $m{=}80$ & 0.2084 & 0.1664 & 0.1453 & 0.1402 & 0.2049 & 0.1640 & 0.1422 & 0.1372 & 0.1933 & 0.1554 & 0.1319 & 0.1270\\
Stanford Cars / SigLIP & 0.1829 & 0.1732 & 0.1647 & 0.1633 & 0.1716 & 0.1608 & 0.1523 & 0.1509 & 0.1574 & 0.1463 & 0.1373 & 0.1359\\
Stanford Cars / CLIP & 0.2397 & 0.2352 & 0.2313 & 0.2330 & 0.2293 & 0.2241 & 0.2194 & 0.2210 & 0.2165 & 0.2115 & 0.2060 & 0.2074\\
ImageNet-R / CLIP & 0.0765 & 0.0540 & 0.0426 & 0.0397 & 0.0735 & 0.0517 & 0.0394 & 0.0367 & 0.0694 & 0.0486 & 0.0361 & 0.0335\\
ImageNet-R / SigLIP & 0.0409 & 0.0267 & 0.0166 & 0.0160 & 0.0385 & 0.0256 & 0.0156 & 0.0150 & 0.0359 & 0.0242 & 0.0142 & 0.0137\\
CUB-200 / CLIP & 0.2552 & 0.2530 & 0.2528 & 0.2502 & 0.2426 & 0.2389 & 0.2374 & 0.2348 & 0.2299 & 0.2255 & 0.2231 & 0.2208\\
CUB-200 / SigLIP & 0.2106 & 0.2057 & 0.2013 & 0.2003 & 0.2031 & 0.1981 & 0.1940 & 0.1931 & 0.1927 & 0.1878 & 0.1837 & 0.1827\\
ImageNet-Sketch / CLIP & 0.0762 & 0.0577 & 0.0521 & 0.0505 & 0.0706 & 0.0541 & 0.0480 & 0.0466 & 0.0640 & 0.0498 & 0.0433 & 0.0420\\
ImageNet-Sketch / SigLIP & 0.0431 & 0.0300 & 0.0249 & 0.0238 & 0.0398 & 0.0275 & 0.0216 & 0.0208 & 0.0357 & 0.0246 & 0.0182 & 0.0175\\}
\pcpdef{tau08.cars-clip-m20.frozen}{16.747}
\pcpdef{tau08.cars-clip-m20.pre}{17.600}
\pcpdef{tau08.cars-clip-m20.prefix}{17.102}
\pcpdef{tau08.cars-clip-m20.raw}{24.861}
\pcpdef{tau08.cars-clip-m20.refresh}{16.582}
\pcpdef{tau08.food-siglip-m20.frozen}{4.812}
\pcpdef{tau08.food-siglip-m20.pre}{6.353}
\pcpdef{tau08.food-siglip-m20.prefix}{5.709}
\pcpdef{tau08.food-siglip-m20.raw}{7.652}
\pcpdef{tau08.food-siglip-m20.refresh}{4.977}
\pcpdef{tau095.cars-clip-m20.frozen}{44.373}
\pcpdef{tau095.cars-clip-m20.pre}{45.928}
\pcpdef{tau095.cars-clip-m20.prefix}{45.117}
\pcpdef{tau095.cars-clip-m20.raw}{65.465}
\pcpdef{tau095.cars-clip-m20.refresh}{44.103}
\pcpdef{tau095.food-siglip-m20.frozen}{19.854}
\pcpdef{tau095.food-siglip-m20.pre}{24.975}
\pcpdef{tau095.food-siglip-m20.prefix}{22.788}
\pcpdef{tau095.food-siglip-m20.raw}{34.267}
\pcpdef{tau095.food-siglip-m20.refresh}{20.241}
\pcpdef{ts.add.frozen.range}{0.1\pcpdash{}0.9}
\pcpdef{ts.add.refresh.range}{0.2\pcpdash{}2.5}
\pcpdef{ts.cars-clip-m20.K}{196}
\pcpdef{ts.cars-clip-m20.add.frozen}{0.92}
\pcpdef{ts.cars-clip-m20.add.refresh}{1.25}
\pcpdef{ts.cars-clip-m20.ci.frozen}{[-0.79,\,-0.28]}
\pcpdef{ts.cars-clip-m20.ci.pre}{[+0.26,\,+0.81]}
\pcpdef{ts.cars-clip-m20.ci.refresh}{[-1.00,\,-0.55]}
\pcpdef{ts.cars-clip-m20.closed.frozen}{50.1}
\pcpdef{ts.cars-clip-m20.closed.refresh}{40.9}
\pcpdef{ts.cars-clip-m20.delta.frozen}{-0.534}
\pcpdef{ts.cars-clip-m20.delta.pre}{+0.535}
\pcpdef{ts.cars-clip-m20.delta.refresh}{-0.775}
\pcpdef{ts.cars-clip-m20.frozen}{29.021}
\pcpdef{ts.cars-clip-m20.macro.frozen}{0.9035}
\pcpdef{ts.cars-clip-m20.macro.pre}{0.9020}
\pcpdef{ts.cars-clip-m20.macro.prefix}{0.9023}
\pcpdef{ts.cars-clip-m20.macro.raw}{0.9068}
\pcpdef{ts.cars-clip-m20.macro.refresh}{0.9031}
\pcpdef{ts.cars-clip-m20.pre}{30.090}
\pcpdef{ts.cars-clip-m20.prefix}{29.555}
\pcpdef{ts.cars-clip-m20.raw}{44.854}
\pcpdef{ts.cars-clip-m20.reach.frozen}{67.3}
\pcpdef{ts.cars-clip-m20.reach.refresh}{81.1}
\pcpdef{ts.cars-clip-m20.refit}{98}
\pcpdef{ts.cars-clip-m20.refresh}{28.780}
\pcpdef{ts.cars-clip-m20.refreshminusfrozen}{-0.241}
\pcpdef{ts.cars-clip-m20.rescored}{1{,}960}
\pcpdef{ts.cars-clip-m20.rev.frozen}{8.4}
\pcpdef{ts.cars-clip-m20.rev.refresh}{11.6}
\pcpdef{ts.cars-clip-m20.rf.ci}{[-0.562,\,+0.081]}
\pcpdef{ts.cars-clip-m20.rf.mean}{-0.241}
\pcpdef{ts.cars-clip-m20.wins.frozen}{0/5}
\pcpdef{ts.cars-clip-m20.wins.pre}{5/5}
\pcpdef{ts.cars-clip-m20.wins.refresh}{0/5}
\pcpdef{ts.cars-clip-m20.work.blocked}{204}
\pcpdef{ts.cars-clip-m20.work.direct}{294}
\pcpdef{ts.cars-clip-m20.work.route}{blocked}
\pcpdef{ts.cars-clip-m20.worst.frozen}{0.6304}
\pcpdef{ts.cars-clip-m20.worst.pre}{0.6450}
\pcpdef{ts.cars-clip-m20.worst.prefix}{0.6539}
\pcpdef{ts.cars-clip-m20.worst.raw}{0.6451}
\pcpdef{ts.cars-clip-m20.worst.refresh}{0.6640}
\pcpdef{ts.cars-siglip-m20.K}{196}
\pcpdef{ts.cars-siglip-m20.add.frozen}{0.42}
\pcpdef{ts.cars-siglip-m20.add.refresh}{0.85}
\pcpdef{ts.cars-siglip-m20.ci.frozen}{[-1.77,\,-1.08]}
\pcpdef{ts.cars-siglip-m20.ci.pre}{[+0.89,\,+1.29]}
\pcpdef{ts.cars-siglip-m20.ci.refresh}{[-1.52,\,-1.00]}
\pcpdef{ts.cars-siglip-m20.closed.frozen}{43.4}
\pcpdef{ts.cars-siglip-m20.closed.refresh}{46.4}
\pcpdef{ts.cars-siglip-m20.delta.frozen}{-1.423}
\pcpdef{ts.cars-siglip-m20.delta.pre}{+1.092}
\pcpdef{ts.cars-siglip-m20.delta.refresh}{-1.264}
\pcpdef{ts.cars-siglip-m20.frozen}{15.083}
\pcpdef{ts.cars-siglip-m20.macro.frozen}{0.8983}
\pcpdef{ts.cars-siglip-m20.macro.pre}{0.8983}
\pcpdef{ts.cars-siglip-m20.macro.prefix}{0.8983}
\pcpdef{ts.cars-siglip-m20.macro.raw}{0.8972}
\pcpdef{ts.cars-siglip-m20.macro.refresh}{0.8987}
\pcpdef{ts.cars-siglip-m20.pre}{17.598}
\pcpdef{ts.cars-siglip-m20.prefix}{16.506}
\pcpdef{ts.cars-siglip-m20.raw}{30.133}
\pcpdef{ts.cars-siglip-m20.reach.frozen}{59.4}
\pcpdef{ts.cars-siglip-m20.reach.refresh}{69.1}
\pcpdef{ts.cars-siglip-m20.refit}{98}
\pcpdef{ts.cars-siglip-m20.refresh}{15.242}
\pcpdef{ts.cars-siglip-m20.refreshminusfrozen}{+0.159}
\pcpdef{ts.cars-siglip-m20.rescored}{1{,}960}
\pcpdef{ts.cars-siglip-m20.rev.frozen}{19.6}
\pcpdef{ts.cars-siglip-m20.rev.refresh}{22.0}
\pcpdef{ts.cars-siglip-m20.rf.ci}{[-0.045,\,+0.363]}
\pcpdef{ts.cars-siglip-m20.rf.mean}{+0.159}
\pcpdef{ts.cars-siglip-m20.wins.frozen}{0/5}
\pcpdef{ts.cars-siglip-m20.wins.pre}{5/5}
\pcpdef{ts.cars-siglip-m20.wins.refresh}{0/5}
\pcpdef{ts.cars-siglip-m20.work.blocked}{204}
\pcpdef{ts.cars-siglip-m20.work.direct}{294}
\pcpdef{ts.cars-siglip-m20.work.route}{blocked}
\pcpdef{ts.cars-siglip-m20.worst.frozen}{0.5790}
\pcpdef{ts.cars-siglip-m20.worst.pre}{0.6378}
\pcpdef{ts.cars-siglip-m20.worst.prefix}{0.6308}
\pcpdef{ts.cars-siglip-m20.worst.raw}{0.6226}
\pcpdef{ts.cars-siglip-m20.worst.refresh}{0.6258}
\pcpdef{ts.cells}{55}
\pcpdef{ts.closed.frozen.range}{37.8\pcpdash{}50.1}
\pcpdef{ts.closed.refresh.max}{58.5}
\pcpdef{ts.closed.refresh.min}{40.9}
\pcpdef{ts.closed.refresh.range}{40.9\pcpdash{}58.5}
\pcpdef{ts.corpora}{5}
\pcpdef{ts.cub200-clip-m20.K}{200}
\pcpdef{ts.cub200-clip-m20.add.frozen}{0.75}
\pcpdef{ts.cub200-clip-m20.add.refresh}{1.47}
\pcpdef{ts.cub200-clip-m20.ci.frozen}{[-1.41,\,-0.33]}
\pcpdef{ts.cub200-clip-m20.ci.pre}{[+0.20,\,+0.86]}
\pcpdef{ts.cub200-clip-m20.ci.refresh}{[-0.72,\,-0.25]}
\pcpdef{ts.cub200-clip-m20.closed.frozen}{37.8}
\pcpdef{ts.cub200-clip-m20.closed.refresh}{52.1}
\pcpdef{ts.cub200-clip-m20.delta.frozen}{-0.872}
\pcpdef{ts.cub200-clip-m20.delta.pre}{+0.529}
\pcpdef{ts.cub200-clip-m20.delta.refresh}{-0.487}
\pcpdef{ts.cub200-clip-m20.frozen}{28.849}
\pcpdef{ts.cub200-clip-m20.macro.frozen}{0.8922}
\pcpdef{ts.cub200-clip-m20.macro.pre}{0.8945}
\pcpdef{ts.cub200-clip-m20.macro.prefix}{0.8957}
\pcpdef{ts.cub200-clip-m20.macro.raw}{0.8984}
\pcpdef{ts.cub200-clip-m20.macro.refresh}{0.8969}
\pcpdef{ts.cub200-clip-m20.pre}{30.250}
\pcpdef{ts.cub200-clip-m20.prefix}{29.721}
\pcpdef{ts.cub200-clip-m20.raw}{60.291}
\pcpdef{ts.cub200-clip-m20.reach.frozen}{69.2}
\pcpdef{ts.cub200-clip-m20.reach.refresh}{87.0}
\pcpdef{ts.cub200-clip-m20.refit}{100}
\pcpdef{ts.cub200-clip-m20.refresh}{29.234}
\pcpdef{ts.cub200-clip-m20.refreshminusfrozen}{+0.384}
\pcpdef{ts.cub200-clip-m20.rescored}{2{,}000}
\pcpdef{ts.cub200-clip-m20.rev.frozen}{10.0}
\pcpdef{ts.cub200-clip-m20.rev.refresh}{11.6}
\pcpdef{ts.cub200-clip-m20.rf.ci}{[+0.047,\,+0.722]}
\pcpdef{ts.cub200-clip-m20.rf.mean}{+0.384}
\pcpdef{ts.cub200-clip-m20.wins.frozen}{0/5}
\pcpdef{ts.cub200-clip-m20.wins.pre}{5/5}
\pcpdef{ts.cub200-clip-m20.wins.refresh}{0/5}
\pcpdef{ts.cub200-clip-m20.work.blocked}{208}
\pcpdef{ts.cub200-clip-m20.work.direct}{300}
\pcpdef{ts.cub200-clip-m20.work.route}{blocked}
\pcpdef{ts.cub200-clip-m20.worst.frozen}{0.5634}
\pcpdef{ts.cub200-clip-m20.worst.pre}{0.6000}
\pcpdef{ts.cub200-clip-m20.worst.prefix}{0.6108}
\pcpdef{ts.cub200-clip-m20.worst.raw}{0.5933}
\pcpdef{ts.cub200-clip-m20.worst.refresh}{0.6067}
\pcpdef{ts.cub200-siglip-m20.K}{200}
\pcpdef{ts.cub200-siglip-m20.add.frozen}{0.58}
\pcpdef{ts.cub200-siglip-m20.add.refresh}{0.98}
\pcpdef{ts.cub200-siglip-m20.ci.frozen}{[-0.96,\,-0.57]}
\pcpdef{ts.cub200-siglip-m20.ci.pre}{[+0.13,\,+1.10]}
\pcpdef{ts.cub200-siglip-m20.ci.refresh}{[-0.92,\,-0.46]}
\pcpdef{ts.cub200-siglip-m20.closed.frozen}{44.6}
\pcpdef{ts.cub200-siglip-m20.closed.refresh}{47.1}
\pcpdef{ts.cub200-siglip-m20.delta.frozen}{-0.766}
\pcpdef{ts.cub200-siglip-m20.delta.pre}{+0.617}
\pcpdef{ts.cub200-siglip-m20.delta.refresh}{-0.694}
\pcpdef{ts.cub200-siglip-m20.frozen}{20.248}
\pcpdef{ts.cub200-siglip-m20.macro.frozen}{0.8943}
\pcpdef{ts.cub200-siglip-m20.macro.pre}{0.8955}
\pcpdef{ts.cub200-siglip-m20.macro.prefix}{0.8955}
\pcpdef{ts.cub200-siglip-m20.macro.raw}{0.8948}
\pcpdef{ts.cub200-siglip-m20.macro.refresh}{0.8960}
\pcpdef{ts.cub200-siglip-m20.pre}{21.630}
\pcpdef{ts.cub200-siglip-m20.prefix}{21.014}
\pcpdef{ts.cub200-siglip-m20.raw}{44.448}
\pcpdef{ts.cub200-siglip-m20.reach.frozen}{62.0}
\pcpdef{ts.cub200-siglip-m20.reach.refresh}{79.0}
\pcpdef{ts.cub200-siglip-m20.refit}{100}
\pcpdef{ts.cub200-siglip-m20.refresh}{20.320}
\pcpdef{ts.cub200-siglip-m20.refreshminusfrozen}{+0.072}
\pcpdef{ts.cub200-siglip-m20.rescored}{2{,}000}
\pcpdef{ts.cub200-siglip-m20.rev.frozen}{12.2}
\pcpdef{ts.cub200-siglip-m20.rev.refresh}{15.0}
\pcpdef{ts.cub200-siglip-m20.rf.ci}{[-0.108,\,+0.252]}
\pcpdef{ts.cub200-siglip-m20.rf.mean}{+0.072}
\pcpdef{ts.cub200-siglip-m20.wins.frozen}{0/5}
\pcpdef{ts.cub200-siglip-m20.wins.pre}{4/5}
\pcpdef{ts.cub200-siglip-m20.wins.refresh}{0/5}
\pcpdef{ts.cub200-siglip-m20.work.blocked}{208}
\pcpdef{ts.cub200-siglip-m20.work.direct}{300}
\pcpdef{ts.cub200-siglip-m20.work.route}{blocked}
\pcpdef{ts.cub200-siglip-m20.worst.frozen}{0.5533}
\pcpdef{ts.cub200-siglip-m20.worst.pre}{0.6200}
\pcpdef{ts.cub200-siglip-m20.worst.prefix}{0.5975}
\pcpdef{ts.cub200-siglip-m20.worst.raw}{0.5267}
\pcpdef{ts.cub200-siglip-m20.worst.refresh}{0.5800}
\pcpdef{ts.food-siglip-m20.K}{101}
\pcpdef{ts.food-siglip-m20.add.frozen}{0.83}
\pcpdef{ts.food-siglip-m20.add.refresh}{2.53}
\pcpdef{ts.food-siglip-m20.ci.frozen}{[-2.19,\,-1.45]}
\pcpdef{ts.food-siglip-m20.ci.pre}{[+1.07,\,+2.08]}
\pcpdef{ts.food-siglip-m20.ci.refresh}{[-1.66,\,-0.58]}
\pcpdef{ts.food-siglip-m20.closed.frozen}{46.5}
\pcpdef{ts.food-siglip-m20.closed.refresh}{58.5}
\pcpdef{ts.food-siglip-m20.delta.frozen}{-1.818}
\pcpdef{ts.food-siglip-m20.delta.pre}{+1.578}
\pcpdef{ts.food-siglip-m20.delta.refresh}{-1.120}
\pcpdef{ts.food-siglip-m20.frozen}{10.330}
\pcpdef{ts.food-siglip-m20.macro.frozen}{0.9239}
\pcpdef{ts.food-siglip-m20.macro.pre}{0.9269}
\pcpdef{ts.food-siglip-m20.macro.prefix}{0.9267}
\pcpdef{ts.food-siglip-m20.macro.raw}{0.9334}
\pcpdef{ts.food-siglip-m20.macro.refresh}{0.9283}
\pcpdef{ts.food-siglip-m20.pre}{13.727}
\pcpdef{ts.food-siglip-m20.prefix}{12.148}
\pcpdef{ts.food-siglip-m20.raw}{17.736}
\pcpdef{ts.food-siglip-m20.reach.frozen}{64.9}
\pcpdef{ts.food-siglip-m20.reach.refresh}{80.3}
\pcpdef{ts.food-siglip-m20.refit}{50}
\pcpdef{ts.food-siglip-m20.refresh}{11.029}
\pcpdef{ts.food-siglip-m20.refreshminusfrozen}{+0.698}
\pcpdef{ts.food-siglip-m20.rescored}{1{,}000}
\pcpdef{ts.food-siglip-m20.rev.frozen}{32.5}
\pcpdef{ts.food-siglip-m20.rev.refresh}{32.9}
\pcpdef{ts.food-siglip-m20.rf.ci}{[+0.364,\,+1.033]}
\pcpdef{ts.food-siglip-m20.rf.mean}{+0.698}
\pcpdef{ts.food-siglip-m20.wins.frozen}{0/5}
\pcpdef{ts.food-siglip-m20.wins.pre}{5/5}
\pcpdef{ts.food-siglip-m20.wins.refresh}{0/5}
\pcpdef{ts.food-siglip-m20.work.blocked}{109}
\pcpdef{ts.food-siglip-m20.work.direct}{151}
\pcpdef{ts.food-siglip-m20.work.route}{blocked}
\pcpdef{ts.food-siglip-m20.worst.frozen}{0.7088}
\pcpdef{ts.food-siglip-m20.worst.pre}{0.7440}
\pcpdef{ts.food-siglip-m20.worst.prefix}{0.7424}
\pcpdef{ts.food-siglip-m20.worst.raw}{0.7600}
\pcpdef{ts.food-siglip-m20.worst.refresh}{0.7448}
\pcpdef{ts.food-siglip-m40.K}{101}
\pcpdef{ts.food-siglip-m40.add.frozen}{0.72}
\pcpdef{ts.food-siglip-m40.add.refresh}{1.63}
\pcpdef{ts.food-siglip-m40.ci.frozen}{[-1.78,\,-1.50]}
\pcpdef{ts.food-siglip-m40.ci.pre}{[+0.98,\,+1.63]}
\pcpdef{ts.food-siglip-m40.ci.refresh}{[-1.51,\,-0.82]}
\pcpdef{ts.food-siglip-m40.closed.frozen}{44.3}
\pcpdef{ts.food-siglip-m40.closed.refresh}{52.8}
\pcpdef{ts.food-siglip-m40.delta.frozen}{-1.642}
\pcpdef{ts.food-siglip-m40.delta.pre}{+1.306}
\pcpdef{ts.food-siglip-m40.delta.refresh}{-1.167}
\pcpdef{ts.food-siglip-m40.frozen}{8.832}
\pcpdef{ts.food-siglip-m40.macro.frozen}{0.9214}
\pcpdef{ts.food-siglip-m40.macro.pre}{0.9253}
\pcpdef{ts.food-siglip-m40.macro.prefix}{0.9256}
\pcpdef{ts.food-siglip-m40.macro.raw}{0.9314}
\pcpdef{ts.food-siglip-m40.macro.refresh}{0.9263}
\pcpdef{ts.food-siglip-m40.pre}{11.781}
\pcpdef{ts.food-siglip-m40.prefix}{10.475}
\pcpdef{ts.food-siglip-m40.raw}{15.199}
\pcpdef{ts.food-siglip-m40.reach.frozen}{62.4}
\pcpdef{ts.food-siglip-m40.reach.refresh}{73.8}
\pcpdef{ts.food-siglip-m40.refit}{50}
\pcpdef{ts.food-siglip-m40.refresh}{9.308}
\pcpdef{ts.food-siglip-m40.refreshminusfrozen}{+0.475}
\pcpdef{ts.food-siglip-m40.rescored}{2{,}000}
\pcpdef{ts.food-siglip-m40.rev.frozen}{34.2}
\pcpdef{ts.food-siglip-m40.rev.refresh}{32.9}
\pcpdef{ts.food-siglip-m40.rf.ci}{[+0.191,\,+0.760]}
\pcpdef{ts.food-siglip-m40.rf.mean}{+0.475}
\pcpdef{ts.food-siglip-m40.wins.frozen}{0/5}
\pcpdef{ts.food-siglip-m40.wins.pre}{5/5}
\pcpdef{ts.food-siglip-m40.wins.refresh}{0/5}
\pcpdef{ts.food-siglip-m40.work.blocked}{109}
\pcpdef{ts.food-siglip-m40.work.direct}{151}
\pcpdef{ts.food-siglip-m40.work.route}{blocked}
\pcpdef{ts.food-siglip-m40.worst.frozen}{0.7488}
\pcpdef{ts.food-siglip-m40.worst.pre}{0.7984}
\pcpdef{ts.food-siglip-m40.worst.prefix}{0.8046}
\pcpdef{ts.food-siglip-m40.worst.raw}{0.7918}
\pcpdef{ts.food-siglip-m40.worst.refresh}{0.7976}
\pcpdef{ts.food-siglip-m80.K}{101}
\pcpdef{ts.food-siglip-m80.add.frozen}{0.69}
\pcpdef{ts.food-siglip-m80.add.refresh}{1.25}
\pcpdef{ts.food-siglip-m80.ci.frozen}{[-1.77,\,-1.49]}
\pcpdef{ts.food-siglip-m80.ci.pre}{[+0.99,\,+1.54]}
\pcpdef{ts.food-siglip-m80.ci.refresh}{[-1.53,\,-1.00]}
\pcpdef{ts.food-siglip-m80.closed.frozen}{43.8}
\pcpdef{ts.food-siglip-m80.closed.refresh}{50.0}
\pcpdef{ts.food-siglip-m80.delta.frozen}{-1.626}
\pcpdef{ts.food-siglip-m80.delta.pre}{+1.265}
\pcpdef{ts.food-siglip-m80.delta.refresh}{-1.267}
\pcpdef{ts.food-siglip-m80.frozen}{8.425}
\pcpdef{ts.food-siglip-m80.macro.frozen}{0.9241}
\pcpdef{ts.food-siglip-m80.macro.pre}{0.9292}
\pcpdef{ts.food-siglip-m80.macro.prefix}{0.9288}
\pcpdef{ts.food-siglip-m80.macro.raw}{0.9342}
\pcpdef{ts.food-siglip-m80.macro.refresh}{0.9289}
\pcpdef{ts.food-siglip-m80.pre}{11.316}
\pcpdef{ts.food-siglip-m80.prefix}{10.051}
\pcpdef{ts.food-siglip-m80.raw}{14.307}
\pcpdef{ts.food-siglip-m80.reach.frozen}{61.9}
\pcpdef{ts.food-siglip-m80.reach.refresh}{69.7}
\pcpdef{ts.food-siglip-m80.refit}{50}
\pcpdef{ts.food-siglip-m80.refresh}{8.784}
\pcpdef{ts.food-siglip-m80.refreshminusfrozen}{+0.359}
\pcpdef{ts.food-siglip-m80.rescored}{4{,}000}
\pcpdef{ts.food-siglip-m80.rev.frozen}{34.7}
\pcpdef{ts.food-siglip-m80.rev.refresh}{32.6}
\pcpdef{ts.food-siglip-m80.rf.ci}{[+0.229,\,+0.489]}
\pcpdef{ts.food-siglip-m80.rf.mean}{+0.359}
\pcpdef{ts.food-siglip-m80.wins.frozen}{0/5}
\pcpdef{ts.food-siglip-m80.wins.pre}{5/5}
\pcpdef{ts.food-siglip-m80.wins.refresh}{0/5}
\pcpdef{ts.food-siglip-m80.work.blocked}{109}
\pcpdef{ts.food-siglip-m80.work.direct}{151}
\pcpdef{ts.food-siglip-m80.work.route}{blocked}
\pcpdef{ts.food-siglip-m80.worst.frozen}{0.7887}
\pcpdef{ts.food-siglip-m80.worst.pre}{0.8332}
\pcpdef{ts.food-siglip-m80.worst.prefix}{0.8428}
\pcpdef{ts.food-siglip-m80.worst.raw}{0.8228}
\pcpdef{ts.food-siglip-m80.worst.refresh}{0.8333}
\pcpdef{ts.gate.entries}{132}
\pcpdef{ts.grids}{11}
\pcpdef{ts.grids.word}{eleven}
\pcpdef{ts.harness.entries}{132}
\pcpdef{ts.imagenet-r-clip-m20.K}{200}
\pcpdef{ts.imagenet-r-clip-m20.add.frozen}{0.22}
\pcpdef{ts.imagenet-r-clip-m20.add.refresh}{0.59}
\pcpdef{ts.imagenet-r-clip-m20.ci.frozen}{[-1.46,\,-1.13]}
\pcpdef{ts.imagenet-r-clip-m20.ci.pre}{[+0.88,\,+1.52]}
\pcpdef{ts.imagenet-r-clip-m20.ci.refresh}{[-1.38,\,-0.56]}
\pcpdef{ts.imagenet-r-clip-m20.closed.frozen}{48.2}
\pcpdef{ts.imagenet-r-clip-m20.closed.refresh}{55.3}
\pcpdef{ts.imagenet-r-clip-m20.delta.frozen}{-1.292}
\pcpdef{ts.imagenet-r-clip-m20.delta.pre}{+1.200}
\pcpdef{ts.imagenet-r-clip-m20.delta.refresh}{-0.971}
\pcpdef{ts.imagenet-r-clip-m20.frozen}{4.541}
\pcpdef{ts.imagenet-r-clip-m20.macro.frozen}{0.9027}
\pcpdef{ts.imagenet-r-clip-m20.macro.pre}{0.9012}
\pcpdef{ts.imagenet-r-clip-m20.macro.prefix}{0.9024}
\pcpdef{ts.imagenet-r-clip-m20.macro.raw}{0.9037}
\pcpdef{ts.imagenet-r-clip-m20.macro.refresh}{0.9039}
\pcpdef{ts.imagenet-r-clip-m20.pre}{7.033}
\pcpdef{ts.imagenet-r-clip-m20.prefix}{5.833}
\pcpdef{ts.imagenet-r-clip-m20.raw}{14.159}
\pcpdef{ts.imagenet-r-clip-m20.reach.frozen}{55.6}
\pcpdef{ts.imagenet-r-clip-m20.reach.refresh}{68.8}
\pcpdef{ts.imagenet-r-clip-m20.refit}{100}
\pcpdef{ts.imagenet-r-clip-m20.refresh}{4.861}
\pcpdef{ts.imagenet-r-clip-m20.refreshminusfrozen}{+0.321}
\pcpdef{ts.imagenet-r-clip-m20.rescored}{2{,}000}
\pcpdef{ts.imagenet-r-clip-m20.rev.frozen}{43.2}
\pcpdef{ts.imagenet-r-clip-m20.rev.refresh}{44.2}
\pcpdef{ts.imagenet-r-clip-m20.rf.ci}{[-0.015,\,+0.656]}
\pcpdef{ts.imagenet-r-clip-m20.rf.mean}{+0.321}
\pcpdef{ts.imagenet-r-clip-m20.wins.frozen}{0/5}
\pcpdef{ts.imagenet-r-clip-m20.wins.pre}{5/5}
\pcpdef{ts.imagenet-r-clip-m20.wins.refresh}{0/5}
\pcpdef{ts.imagenet-r-clip-m20.work.blocked}{208}
\pcpdef{ts.imagenet-r-clip-m20.work.direct}{300}
\pcpdef{ts.imagenet-r-clip-m20.work.route}{blocked}
\pcpdef{ts.imagenet-r-clip-m20.worst.frozen}{0.5996}
\pcpdef{ts.imagenet-r-clip-m20.worst.pre}{0.5922}
\pcpdef{ts.imagenet-r-clip-m20.worst.prefix}{0.5929}
\pcpdef{ts.imagenet-r-clip-m20.worst.raw}{0.5813}
\pcpdef{ts.imagenet-r-clip-m20.worst.refresh}{0.5714}
\pcpdef{ts.imagenet-r-siglip-m20.K}{200}
\pcpdef{ts.imagenet-r-siglip-m20.add.frozen}{0.08}
\pcpdef{ts.imagenet-r-siglip-m20.add.refresh}{0.17}
\pcpdef{ts.imagenet-r-siglip-m20.ci.frozen}{[-1.06,\,-0.66]}
\pcpdef{ts.imagenet-r-siglip-m20.ci.pre}{[+0.42,\,+0.93]}
\pcpdef{ts.imagenet-r-siglip-m20.ci.refresh}{[-1.26,\,-0.35]}
\pcpdef{ts.imagenet-r-siglip-m20.closed.frozen}{44.0}
\pcpdef{ts.imagenet-r-siglip-m20.closed.refresh}{45.7}
\pcpdef{ts.imagenet-r-siglip-m20.delta.frozen}{-0.859}
\pcpdef{ts.imagenet-r-siglip-m20.delta.pre}{+0.674}
\pcpdef{ts.imagenet-r-siglip-m20.delta.refresh}{-0.802}
\pcpdef{ts.imagenet-r-siglip-m20.frozen}{1.767}
\pcpdef{ts.imagenet-r-siglip-m20.macro.frozen}{0.9127}
\pcpdef{ts.imagenet-r-siglip-m20.macro.pre}{0.9036}
\pcpdef{ts.imagenet-r-siglip-m20.macro.prefix}{0.9037}
\pcpdef{ts.imagenet-r-siglip-m20.macro.raw}{0.9036}
\pcpdef{ts.imagenet-r-siglip-m20.macro.refresh}{0.9025}
\pcpdef{ts.imagenet-r-siglip-m20.pre}{3.300}
\pcpdef{ts.imagenet-r-siglip-m20.prefix}{2.626}
\pcpdef{ts.imagenet-r-siglip-m20.raw}{5.674}
\pcpdef{ts.imagenet-r-siglip-m20.reach.frozen}{48.0}
\pcpdef{ts.imagenet-r-siglip-m20.reach.refresh}{53.4}
\pcpdef{ts.imagenet-r-siglip-m20.refit}{100}
\pcpdef{ts.imagenet-r-siglip-m20.refresh}{1.824}
\pcpdef{ts.imagenet-r-siglip-m20.refreshminusfrozen}{+0.057}
\pcpdef{ts.imagenet-r-siglip-m20.rescored}{2{,}000}
\pcpdef{ts.imagenet-r-siglip-m20.rev.frozen}{51.8}
\pcpdef{ts.imagenet-r-siglip-m20.rev.refresh}{52.7}
\pcpdef{ts.imagenet-r-siglip-m20.rf.ci}{[-0.211,\,+0.326]}
\pcpdef{ts.imagenet-r-siglip-m20.rf.mean}{+0.057}
\pcpdef{ts.imagenet-r-siglip-m20.wins.frozen}{0/5}
\pcpdef{ts.imagenet-r-siglip-m20.wins.pre}{5/5}
\pcpdef{ts.imagenet-r-siglip-m20.wins.refresh}{0/5}
\pcpdef{ts.imagenet-r-siglip-m20.work.blocked}{208}
\pcpdef{ts.imagenet-r-siglip-m20.work.direct}{300}
\pcpdef{ts.imagenet-r-siglip-m20.work.route}{blocked}
\pcpdef{ts.imagenet-r-siglip-m20.worst.frozen}{0.6349}
\pcpdef{ts.imagenet-r-siglip-m20.worst.pre}{0.6508}
\pcpdef{ts.imagenet-r-siglip-m20.worst.prefix}{0.6500}
\pcpdef{ts.imagenet-r-siglip-m20.worst.raw}{0.5550}
\pcpdef{ts.imagenet-r-siglip-m20.worst.refresh}{0.6460}
\pcpdef{ts.imagenet-sketch-clip-m20.K}{930}
\pcpdef{ts.imagenet-sketch-clip-m20.add.frozen}{0.19}
\pcpdef{ts.imagenet-sketch-clip-m20.add.refresh}{0.44}
\pcpdef{ts.imagenet-sketch-clip-m20.ci.frozen}{[-3.74,\,-3.01]}
\pcpdef{ts.imagenet-sketch-clip-m20.ci.pre}{[+2.71,\,+3.58]}
\pcpdef{ts.imagenet-sketch-clip-m20.ci.refresh}{[-3.39,\,-2.01]}
\pcpdef{ts.imagenet-sketch-clip-m20.closed.frozen}{48.2}
\pcpdef{ts.imagenet-sketch-clip-m20.closed.refresh}{53.8}
\pcpdef{ts.imagenet-sketch-clip-m20.delta.frozen}{-3.376}
\pcpdef{ts.imagenet-sketch-clip-m20.delta.pre}{+3.143}
\pcpdef{ts.imagenet-sketch-clip-m20.delta.refresh}{-2.700}
\pcpdef{ts.imagenet-sketch-clip-m20.frozen}{23.638}
\pcpdef{ts.imagenet-sketch-clip-m20.macro.frozen}{0.9010}
\pcpdef{ts.imagenet-sketch-clip-m20.macro.pre}{0.9034}
\pcpdef{ts.imagenet-sketch-clip-m20.macro.prefix}{0.9037}
\pcpdef{ts.imagenet-sketch-clip-m20.macro.raw}{0.9041}
\pcpdef{ts.imagenet-sketch-clip-m20.macro.refresh}{0.9045}
\pcpdef{ts.imagenet-sketch-clip-m20.pre}{30.157}
\pcpdef{ts.imagenet-sketch-clip-m20.prefix}{27.014}
\pcpdef{ts.imagenet-sketch-clip-m20.raw}{45.477}
\pcpdef{ts.imagenet-sketch-clip-m20.reach.frozen}{78.4}
\pcpdef{ts.imagenet-sketch-clip-m20.reach.refresh}{90.5}
\pcpdef{ts.imagenet-sketch-clip-m20.refit}{465}
\pcpdef{ts.imagenet-sketch-clip-m20.refresh}{24.314}
\pcpdef{ts.imagenet-sketch-clip-m20.refreshminusfrozen}{+0.676}
\pcpdef{ts.imagenet-sketch-clip-m20.rescored}{9{,}300}
\pcpdef{ts.imagenet-sketch-clip-m20.rev.frozen}{28.6}
\pcpdef{ts.imagenet-sketch-clip-m20.rev.refresh}{31.6}
\pcpdef{ts.imagenet-sketch-clip-m20.rf.ci}{[+0.276,\,+1.077]}
\pcpdef{ts.imagenet-sketch-clip-m20.rf.mean}{+0.676}
\pcpdef{ts.imagenet-sketch-clip-m20.wins.frozen}{0/5}
\pcpdef{ts.imagenet-sketch-clip-m20.wins.pre}{5/5}
\pcpdef{ts.imagenet-sketch-clip-m20.wins.refresh}{0/5}
\pcpdef{ts.imagenet-sketch-clip-m20.work.blocked}{938}
\pcpdef{ts.imagenet-sketch-clip-m20.work.direct}{1{,}395}
\pcpdef{ts.imagenet-sketch-clip-m20.work.route}{blocked}
\pcpdef{ts.imagenet-sketch-clip-m20.worst.frozen}{0.3077}
\pcpdef{ts.imagenet-sketch-clip-m20.worst.pre}{0.5000}
\pcpdef{ts.imagenet-sketch-clip-m20.worst.prefix}{0.4833}
\pcpdef{ts.imagenet-sketch-clip-m20.worst.raw}{0.5090}
\pcpdef{ts.imagenet-sketch-clip-m20.worst.refresh}{0.5043}
\pcpdef{ts.imagenet-sketch-siglip-m20.K}{930}
\pcpdef{ts.imagenet-sketch-siglip-m20.add.frozen}{0.08}
\pcpdef{ts.imagenet-sketch-siglip-m20.add.refresh}{0.18}
\pcpdef{ts.imagenet-sketch-siglip-m20.ci.frozen}{[-3.34,\,-2.69]}
\pcpdef{ts.imagenet-sketch-siglip-m20.ci.pre}{[+1.89,\,+2.86]}
\pcpdef{ts.imagenet-sketch-siglip-m20.ci.refresh}{[-3.33,\,-2.16]}
\pcpdef{ts.imagenet-sketch-siglip-m20.closed.frozen}{44.0}
\pcpdef{ts.imagenet-sketch-siglip-m20.closed.refresh}{46.3}
\pcpdef{ts.imagenet-sketch-siglip-m20.delta.frozen}{-3.017}
\pcpdef{ts.imagenet-sketch-siglip-m20.delta.pre}{+2.372}
\pcpdef{ts.imagenet-sketch-siglip-m20.delta.refresh}{-2.749}
\pcpdef{ts.imagenet-sketch-siglip-m20.frozen}{8.660}
\pcpdef{ts.imagenet-sketch-siglip-m20.macro.frozen}{0.9054}
\pcpdef{ts.imagenet-sketch-siglip-m20.macro.pre}{0.9031}
\pcpdef{ts.imagenet-sketch-siglip-m20.macro.prefix}{0.9039}
\pcpdef{ts.imagenet-sketch-siglip-m20.macro.raw}{0.9015}
\pcpdef{ts.imagenet-sketch-siglip-m20.macro.refresh}{0.9032}
\pcpdef{ts.imagenet-sketch-siglip-m20.pre}{14.049}
\pcpdef{ts.imagenet-sketch-siglip-m20.prefix}{11.677}
\pcpdef{ts.imagenet-sketch-siglip-m20.raw}{22.199}
\pcpdef{ts.imagenet-sketch-siglip-m20.reach.frozen}{66.7}
\pcpdef{ts.imagenet-sketch-siglip-m20.reach.refresh}{77.4}
\pcpdef{ts.imagenet-sketch-siglip-m20.refit}{465}
\pcpdef{ts.imagenet-sketch-siglip-m20.refresh}{8.929}
\pcpdef{ts.imagenet-sketch-siglip-m20.refreshminusfrozen}{+0.269}
\pcpdef{ts.imagenet-sketch-siglip-m20.rescored}{9{,}300}
\pcpdef{ts.imagenet-sketch-siglip-m20.rev.frozen}{45.4}
\pcpdef{ts.imagenet-sketch-siglip-m20.rev.refresh}{47.9}
\pcpdef{ts.imagenet-sketch-siglip-m20.rf.ci}{[-0.064,\,+0.602]}
\pcpdef{ts.imagenet-sketch-siglip-m20.rf.mean}{+0.269}
\pcpdef{ts.imagenet-sketch-siglip-m20.wins.frozen}{0/5}
\pcpdef{ts.imagenet-sketch-siglip-m20.wins.pre}{5/5}
\pcpdef{ts.imagenet-sketch-siglip-m20.wins.refresh}{0/5}
\pcpdef{ts.imagenet-sketch-siglip-m20.work.blocked}{938}
\pcpdef{ts.imagenet-sketch-siglip-m20.work.direct}{1{,}395}
\pcpdef{ts.imagenet-sketch-siglip-m20.work.route}{blocked}
\pcpdef{ts.imagenet-sketch-siglip-m20.worst.frozen}{0.1528}
\pcpdef{ts.imagenet-sketch-siglip-m20.worst.pre}{0.5294}
\pcpdef{ts.imagenet-sketch-siglip-m20.worst.prefix}{0.5314}
\pcpdef{ts.imagenet-sketch-siglip-m20.worst.raw}{0.5034}
\pcpdef{ts.imagenet-sketch-siglip-m20.worst.refresh}{0.5160}
\pcpdef{ts.kernel.bits}{313{,}896{,}080}
\pcpdef{ts.kernel.bitschanged}{0}
\pcpdef{ts.kernel.worstdelta}{1.83\times10^{-15}}
\pcpdef{ts.macro.range}{0.892\pcpdash{}0.934}
\pcpdef{ts.onboarding.maxdelta}{1.11\times10^{-16}}
\pcpdef{ts.parity.range}{12.5\pcpdash{}54.9}
\pcpdef{ts.planner.lambda}{8}
\pcpdef{ts.reach.frozen.range}{48\pcpdash{}78}
\pcpdef{ts.reach.refresh.range}{53\pcpdash{}90}
\pcpdef{ts.refreshminusfrozen.poscount}{10/11}
\pcpdef{ts.refreshminusfrozen.range}{-0.241\text{ to }+0.698}
\pcpdef{ts.refreshminusfrozen.significant}{5/11}
\pcpdef{ts.regression.armpairs}{5}
\pcpdef{ts.regression.compared}{8{,}250}
\pcpdef{ts.regression.differing}{0}
\pcpdef{ts.regression.worstdelta}{0.0}
\pcpdef{ts.rev.frozen.range}{8.4\pcpdash{}51.8}
\pcpdef{ts.rev.refresh.range}{11.6\pcpdash{}52.7}
\pcpdef{ts.route.bits}{941{,}688{,}240}
\pcpdef{ts.route.cells}{55}
\pcpdef{ts.route.differing}{0}
\pcpdef{ts.route.worstdelta}{1.89\times10^{-15}}
\pcpdef{ts.seeds}{5}
\pcpdef{ts.tau}{0.9}
\pcpdef{ts.worst.frozen.range}{0.153\pcpdash{}0.789}
\pcpdef{ts.worst.prefix.range}{0.483\pcpdash{}0.843}
\pcpdef{ts.worst.refresh.range}{0.504\pcpdash{}0.833}
\pcpdef{tz.K}{101}
\pcpdef{tz.base}{50}
\pcpdef{tz.cap}{4}
\pcpdef{tz.gate.leaves}{256}
\pcpdef{tz.novel}{51}
\pcpdef{tz.rows}{25{,}224}
\pcpdef{tz.seed.cal}{400}
\pcpdef{tz.seed.partition}{100}
\pcpdef{tz.seed.refit}{10400}
\pcpdef{tz.set.prefix}{cup cakes, macarons, red velvet cake}
\pcpdef{tz.set.refresh}{cup cakes}
\pcpdef{tz.set.v1}{macarons, red velvet cake}
\pcpdef{tz.truelabel}{macarons}

\input{presentation_overrides} %

\def\wacvPaperID{1546}
\def\confName{WACV}
\def\confYear{2027}

\title{Coverage Hides Churn: Auditing and Preserving\\
Conformal Decisions as VLM Label Registries Grow}

\author{Jai Kumar Sharma\\
Virginia Tech\\
{\tt\small jaisharma@vt.edu}
\and
Peeyush Tapadiya\\
Accenture\\
{\tt\small peeyush.tapadiya@accenture.com}}

\AtBeginDocument{\AddToShipoutPicture*{\AtPageLowerLeft{%
\raisebox{12pt}[0pt][0pt]{\hspace*{0.65in}\parbox{7.2in}{%
\fontsize{6}{7}\selectfont\normalfont
Accepted for publication at WACV 2027. Author's accepted manuscript.\\
\textcopyright{} 2027 IEEE. Personal use of this material is permitted. Permission from IEEE must be obtained for all other uses, in any current or future media, including reprinting/republishing this material for advertising or promotional purposes, creating new collective works, for resale or redistribution to servers or lists, or reuse of any copyrighted component of this work in other works.
}}}}}

\begin{document}
\raggedbottom
\makeatletter
\twocolumn[%
  \@maketitle
  \vspace*{-8pt}%
  \begin{center}%
    \providecolor{pcpwarm}{HTML}{9E2A2B}
\providecolor{pcpcool}{HTML}{1F7A8C}
\providecommand{\pcptruemark}{\,$\star$}
\providecommand{\pcpvar}[2]{\ensuremath{\mathsf{#1}_{\mathsf{#2}}}}

\begingroup\frenchspacing
\begin{tikzpicture}[
  x=1cm, y=1cm, font=\sffamily,
  every node/.style={inner sep=0pt, outer sep=0pt},
  pcpchip/.style={anchor=north west, rounded corners=0.6mm,
    line width=0.4pt, inner xsep=0.11cm, inner ysep=0.055cm,
    text height=0.185cm, text depth=0.058cm, minimum height=0.34cm,
    font=\sffamily\scriptsize},
  pcpstable/.style={pcpchip, fill=black!8, draw=black!45, text=black!88},
  pcpdrop/.style={pcpchip, fill=pcpwarm!6, draw=pcpwarm!55, text=black!38},
  pcpadd/.style={pcpchip, fill=pcpwarm!22, draw=pcpwarm!85, text=black!88,
    font=\sffamily\scriptsize\bfseries},
  pcpnew/.style={pcpchip, fill=white, draw=black!35,
    dash pattern=on 0.7mm off 0.4mm, text=black!48,
    font=\sffamily\scriptsize\itshape},
  pcpglyph/.style={anchor=center, font=\sffamily\scriptsize},
  pcphdr/.style={anchor=center, font=\sffamily\scriptsize\bfseries},
  pcpnote/.style={anchor=north west, font=\sffamily\scriptsize,
    text=black!62, align=flush left,
    execute at begin node={\hyphenpenalty=10000\exhyphenpenalty=10000}},
]
\node[anchor=west, font=\sffamily\scriptsize, text=black!70] (pcptl) at (3.45,-0.15) {\pcpvar{V}{1} registry: 50 labels};
\node[anchor=east, font=\sffamily\scriptsize\bfseries, text=black!70] (pcptr) at (17.35,-0.15) {\pcpvar{V}{2}: 101 labels};
\draw[->, >=stealth, draw=black!45, line width=0.5pt] ([xshift=0.18cm]pcptl.east) -- ([xshift=-0.18cm]pcptr.west);
\node[anchor=center, fill=white, inner xsep=0.12cm, text=black!70, font=\sffamily\scriptsize] at (8.068,-0.15) {$+$51 labels appended};
\draw[draw=black!28, line width=0.4pt, dash pattern=on 0.5mm off 0.5mm] (8.068,-0.42) -- (8.068,-2.64);
\draw[rounded corners=0.7mm, line width=0.4pt, fill=black!7, draw=black!30] (3.45,-0.40) rectangle (7.79,-0.78);
\node[pcphdr, text=black!72] at (5.622,-0.590) {before update (\pcpvar{V}{1} set)};
\draw[rounded corners=0.7mm, line width=0.4pt, fill=pcpwarm!12, draw=pcpwarm!45] (8.34,-0.40) rectangle (12.69,-0.78);
\node[pcphdr, text=pcpwarm] at (10.515,-0.590) {HPS-Refresh (refit)};
\draw[rounded corners=0.7mm, line width=0.4pt, fill=pcpcool!12, draw=pcpcool!45] (13.01,-0.40) rectangle (17.35,-0.78);
\node[pcphdr, text=pcpcool] at (15.178,-0.590) {PrefixCP (ours)};
\node[pcpstable] (pcp00) at (3.85,-0.900) {macarons\pcptruemark};
\node[pcpdrop] (pcp10) at (8.74,-0.900) {macarons\pcptruemark};
\draw[pcpwarm!85, line width=0.5pt] ([xshift=0.6mm]pcp10.west) -- ([xshift=-0.6mm]pcp10.east);
\node[pcpglyph, text=pcpwarm] at (8.543,-1.070) {$\times$};
\node[pcpstable] (pcp20) at (13.41,-0.900) {macarons\pcptruemark};
\node[pcpglyph, text=pcpcool] at (13.207,-1.070) {$\checkmark$};
\node[pcpstable] (pcp01) at (3.85,-1.320) {red velvet cake};
\node[pcpdrop] (pcp11) at (8.74,-1.320) {red velvet cake};
\draw[pcpwarm!85, line width=0.5pt] ([xshift=0.6mm]pcp11.west) -- ([xshift=-0.6mm]pcp11.east);
\node[pcpglyph, text=pcpwarm] at (8.543,-1.490) {$\times$};
\node[pcpstable] (pcp21) at (13.41,-1.320) {red velvet cake};
\node[pcpglyph, text=pcpcool] at (13.207,-1.490) {$\checkmark$};
\node[pcpnew] (pcp12) at (8.74,-1.740) {cup cakes};
\node[pcpnew] (pcp22) at (13.41,-1.740) {cup cakes};
\draw[rounded corners=0.7mm, line width=0.4pt, fill=pcpwarm!8, draw=pcpwarm!35] (8.34,-2.69) rectangle (12.69,-3.05);
\node[anchor=center, font=\sffamily\small\bfseries, text=pcpwarm] at (10.515,-2.870) {$\times$\ old decisions revoked};
\draw[rounded corners=0.7mm, line width=0.4pt, fill=pcpcool!8, draw=pcpcool!35] (13.01,-2.69) rectangle (17.35,-3.05);
\node[anchor=center, font=\sffamily\small\bfseries, text=pcpcool] at (15.178,-2.870) {$\checkmark$\ old decisions kept};
\node[anchor=north west] at (0,0.00) {\includegraphics[width=3.05cm]{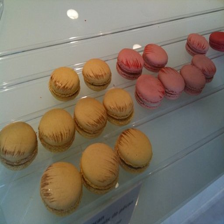}};
\fill[black, opacity=0.60] (0,-2.690) rectangle (3.05,-3.05);
\node[anchor=west, font=\sffamily\scriptsize, text=white] at (0.13,-2.870) {true label: macarons$\,\star$};
\draw[draw=black!55, line width=0.4pt] (0,0.00) rectangle (3.05,-3.05);
\end{tikzpicture}%
\endgroup
  \end{center}%
  \refstepcounter{figure}\label{fig:teaser}%
  \@makecaption{\figurename~\thefigure}{\textbf{Recalibrating can revoke old
  decisions.} For this Food-101/SigLIP image, HPS-Refresh rescores and refits
  the old classes after 51 labels are added, and drops both previously
  included labels, including the true class. PrefixCP lets each class keep the
  registry it was added with, so both old decisions survive. Both methods also
  include the new label, shown dashed. It is the first evaluation image, in
  identifier order, whose sets are small enough to show and whose true old label
  HPS-Refresh drops (supplement Sec.~M); it was scored blurred, as in all
  Food-101 settings, and is shown clean for legibility.}%
  \vspace*{8pt}%
]
\thispagestyle{plain}\@thanks
\makeatother

\begin{abstract}
\looseness=-1
Vision--language classifiers can learn new labels from text alone, so the label
lists they serve keep growing. We show that this growth silently changes
conformal decisions about old labels. Common softmax-based conformal scores
compare each label with an average over all labels; adding labels changes that
average, and with it the score of every old label, even when the model, the
image, and the old calibration data stay the same. Recalibrating after the
update keeps the coverage guarantee, yet across \N{ts.grids.word} settings it
changes at least one old decision on $\N{ts.reach.refresh.range}\%$ of new
images. Coverage cannot reveal this, because it checks only the true label: in
\N{cd.grids.word} controlled settings that keep the old calibration, so that
only the normalizer changes, old-class coverage moves by at most
$\N{cd.mean.dcov.absmax.r2}$ while $\N{cd.mean.reach.range}\%$ of images
change. We propose a lifecycle audit that measures these changes alongside
coverage and set size, and a simple fix, PrefixCP: each class keeps the label
registry that existed when it was added. Its old decisions then never change,
each class keeps exact class-conditional coverage under exchangeability, and
one prefix scan computes all historical normalizers at once. PrefixCP recovers
$\N{ts.closed.refresh.range}\%$ of the set-size gap between never updating the
reference and recalibrating, which gives smaller sets but changes old
decisions.
\end{abstract}

\section{Introduction}
\label{sec:intro}

Vision--language models (VLMs) classify an image by comparing it with a text
description of each label~\cite{radford2021clip,zhai2023siglip}. Adding a label
therefore needs no retraining: a product catalog can add a category by adding
its name. We call the current list of labels the \emph{label registry}.
Deployments often wrap such classifiers in conformal prediction, which returns
a \emph{prediction set} of plausible labels instead of a single guess. A small
set of labeled \emph{calibration examples} decides which labels enter the set,
and the method guarantees \emph{coverage}: the set contains the true label with
a chosen probability, such as
90\%~\cite{vovk2005algorithmic,angelopoulos2023gentle}.

Downstream systems often act on the labels in that set. A catalog might send
an image for manual review whenever its set contains ``damaged packaging.'' Now
suppose the catalog adds new product subcategories. The image has not changed,
and neither has its similarity to ``damaged packaging,'' yet that label can
drop out of the set and the review never happens. The cause is a shared
normalizer: common softmax-based conformal scores compare each label's
similarity with an average over all labels~\cite{sadinle2019least}, so adding
labels changes the score of every old label (Fig.~\ref{fig:mechanism}). We call
changes to old-label decisions \emph{churn}.

Recalibrating after the update does not prevent churn. It restores the coverage
guarantee, yet across \N{ts.grids.word} settings it changes at least one old
decision on $\N{ts.reach.refresh.range}\%$ of new images
(Sec.~\ref{sec:rq2}; Fig.~\ref{fig:teaser} shows one example). Standard
evaluation misses this
because coverage only checks the true label. In \N{cd.grids.word} controlled
settings on Food-101 and Stanford Cars that keep the old calibration
(HPS-Frozen), old-class coverage moves by at most
$\N{cd.mean.dcov.absmax.r2}$ while at least one old decision changes on
$\N{cd.mean.reach.range}\%$ of images (Sec.~\ref{sec:rq1}). Logging past
outputs does not help either: a log can replay decisions for images already
seen, but not for images that arrive after the update.

Our fix rests on one observation: conformal calibration can be done separately
for each class, so classes need not share a score. \textbf{PrefixCP} lets each
class keep the registry that existed when it was added. Its decisions then
never change, and it keeps exact class-conditional coverage under the usual
exchangeability assumption. Every earlier registry is a prefix of the current
label list, so one prefix scan computes all of them cheaply. Stability has a
price: recalibrating every class returns smaller sets in every setting we test.

We contribute (1) a \emph{lifecycle audit} for conformal classifiers with a
growing label registry, which measures churn on new images, coverage, set
size, and dependence on release history; (2) an evaluation on five datasets,
two encoders, and \N{ts.grids.word} settings; and (3) PrefixCP, with coverage
and stability guarantees, an efficient implementation, and a measured
trade-off between stability and set size.

\section{Related work}
\label{sec:related}

\noindent\textbf{Conformal prediction with a fixed label set.}
Conformal methods for VLMs improve the model, the score, or the calibration
procedure, but evaluate one fixed label
set~\cite{fillioux2024foundation,cact2026,dpsm2025,cadapter2024,%
attar2026softmax,fuchs2025localized,zeng2026ebcp,survey2026mmcp}.
We keep the model fixed and let the label set grow. The closest set-size
competitor, Conf-OT~\cite{silva2025confot}, is \emph{transductive}: it reads an
unlabeled batch of test images and guarantees marginal coverage. PrefixCP reads
no test batch and guarantees class-conditional coverage and unchanged old
decisions.

\noindent\textbf{New labels and open-set prediction.}
Other work handles labels that have no calibration
data~\cite{xie2025cgtc,xie2025openset,missingmass2025,ngnie2026cppol}. In our
setting every new class arrives with labeled examples, so we ask a different
question: does a class that is already deployed keep its rule when labels are
added after it?

\noindent\textbf{Model updates and fixed references.}
Backward-compatible learning reduces prediction changes when the \emph{model}
is updated~\cite{shen2020bct,yan2021positive,trauble2021bcpu,gao2025cpcl,milanifard2016churn};
we fix the model and grow the label list. Hubness corrections and frozen label
banks normalize every class against one shared
reference~\cite{smith2017offline,bogolin2022cross,zhou2023dn,jiang2024neglabel,fu2025clipscope,che2025law},
and separated softmax uses different normalizers during
training~\cite{ahn2021ssil}. PrefixCP instead gives each class the reference
that existed when it was added. Our earlier workshop
audit~\cite{uncv2026audit} found poor per-class coverage under distribution
shift; it did not study churn and supplies none of the evidence here.

\section{Setting: a label registry that grows}
\label{sec:setting}

Labels arrive in releases. The registry grows as
$V_1\subseteq\cdots\subseteq V_\nver$, with $K_t=|V_t|$ labels at version $t$
and $K=K_\nver$. Each release adds at least one label, and $\birth(c)$ is the
version at which class $c$ was added. At an update, \emph{old} classes are
those already in the registry and \emph{new} classes are those being added.
Each label has a fixed text \emph{descriptor} (its prompt). A frozen VLM maps
an image $x$ to a unit vector $z=\phi(x)\in\Sph$ and the descriptor of class
$c$ to a unit vector $e_c$; their inner product is the cosine similarity.

\begin{figure}[t]
\centering
\begin{tikzpicture}[x=0.51cm,y=0.72cm,font=\footnotesize]
\fill[wacvblue!7] (5,-0.25) rectangle (10.3,2.95);
\node[anchor=north east,font=\scriptsize,text=wacvblue!75!black] at (10.25,2.92) {in the set};
\node[anchor=north west,font=\scriptsize,text=black!45] at (0.05,2.92) {out of the set};
\draw[dashed,black!70,thick] (5,-0.35) -- (5,3.05);
\node[anchor=south,font=\scriptsize,text=black!70] at (5,3.05) {cutoff fitted on $c$'s calibration examples};
\draw[-{Latex[length=1.6mm]},black!55] (0,-0.35) -- (10.4,-0.35);
\node[anchor=north,font=\scriptsize,text=black!65] at (5.2,-0.42) {score of old class $c$ (higher is better)};
\node[anchor=east] at (-0.1,2.2) {before update ($V_1$)};
\node[anchor=east] at (-0.1,1.2) {HPS-Frozen, after ($V_2$)};
\node[anchor=east] at (-0.1,0.2) {\textbf{PrefixCP}, after ($V_2$)};
\draw[densely dotted,black!35] (7.4,2.05) -- (7.4,0.35);
\fill[black!50] (7.4,2.2) circle (3.2pt);
\draw[-{Latex[length=1.6mm]},diagorange,thick] (7.15,2.08) -- (3.25,1.3);
\fill[diagorange] (3.0,1.2) circle (3.2pt);
\node[anchor=east,font=\scriptsize,text=diagorange!85!black] at (2.7,1.2) {revoked};
\node[anchor=east,font=\scriptsize,text=diagorange!85!black] at (4.85,1.95) {normalizer rises};
\fill[wacvblue] (7.4,0.2) circle (3.2pt);
\node[anchor=west,font=\scriptsize,text=wacvblue!80!black] at (7.7,0.2) {kept};
\end{tikzpicture}
\caption{\textbf{How adding labels revokes an old label} (schematic). The
image's similarity to old class $c$ is unchanged, but the new labels in $V_2$
raise the shared normalizer, so $c$'s score falls below the cutoff fitted on
its own calibration examples and $c$ leaves the set. HPS-Frozen keeps that
cutoff; HPS-Refresh also refits it, which can move decisions either way.
PrefixCP keeps scoring $c$ against $V_1$, so its score and decision do not
move.}
\label{fig:mechanism}
\end{figure}
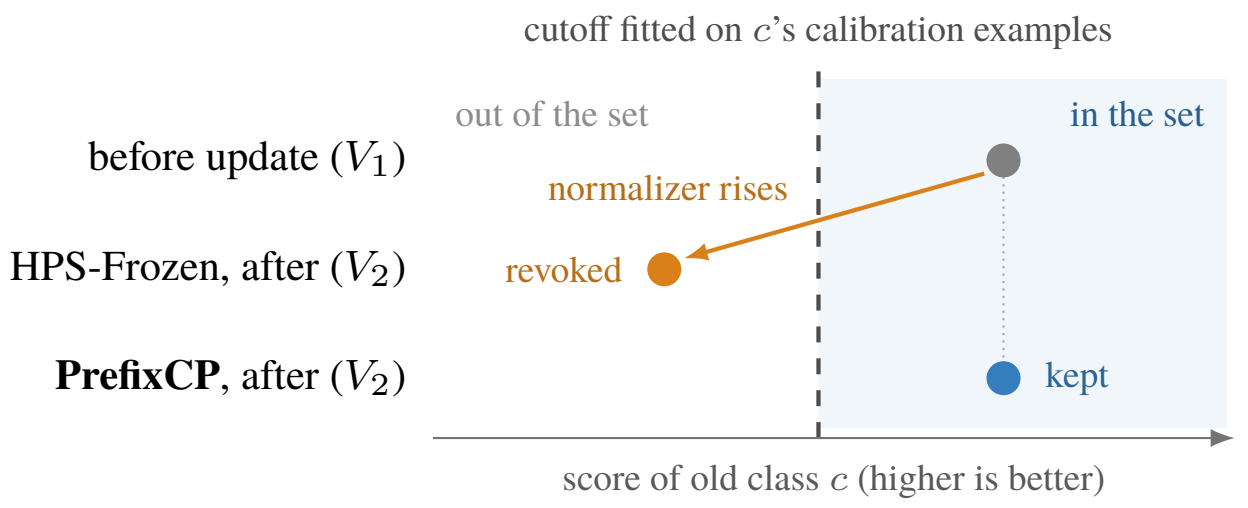

\paragraph{The shared normalizer.}
Homogeneous prediction sets (HPS)~\cite{sadinle2019least}, the standard
conformal method for softmax classifiers, subtract from every similarity one
image-dependent offset computed from all current labels:
\begin{equation}
  \energy_t(z)
  =
  T\log\!\left[
    \frac{1}{K_t}
    \sum_{j\in V_t}
    \exp\!\left(\frac{\ip{z}{e_j}}{T}\right)
  \right],
  \label{eq:active-normalizer}
\end{equation}
where $T>0$ is the model temperature. We call $\energy_t(z)$ the
\emph{normalizer}. The score of class $c$ is
\begin{equation}
  \score_t(z,c)=\ip{z}{e_c}-\energy_t(z).
  \label{eq:active-score}
\end{equation}
This is the temperature-scaled log-softmax probability $T\log p_t(c\mid z)$
plus the constant $T\log K_t$. Each class has its own similarity, but all
classes share one normalizer. Figure~\ref{fig:mechanism} shows what happens when labels are added: the
similarity of an old class stays the same, the normalizer changes, and the
score can cross the cutoff fitted on that class's calibration examples. The
factor $1/K_t$ removes the effect of label count alone, so only \emph{which}
labels are added matters. After an update, HPS can be applied in two ways:
\emph{HPS-Refresh} rescores the old calibration examples with the new registry
and refits every class, and \emph{HPS-Frozen} rescores but keeps the old
calibration values.

\paragraph{Classwise calibration.}
When class $c$ is added, it is calibrated with $m_c$ labeled examples; we call
this \emph{onboarding}, and the resulting rule its \emph{onboarding rule}. The
setting is few-shot, not zero-shot. To decide whether $c$ enters the set for a
new image, we rank the image's score for $c$ among the calibration scores of
class $c$ alone; higher scores mean better matches, and the true class is not
needed. This classwise calibration (Mondrian conformal
prediction~\cite{vovk2003mondrian}) is the design choice PrefixCP relies on: because each class is compared only with its own examples,
different classes need not share a normalizer. Its cost is that every class
needs labeled examples.

A little randomness makes coverage exact. With $m_c=20$ and target
$\cov=0.9$, the image takes one of 21 equally likely ranks, so a fixed cutoff
gives coverage $18/21$ or $19/21$, never exactly $0.9$. Instead, ranks 1--18
are included and rank 19 is included with probability 0.9, giving
$(18+0.9)/21=0.9$. A \emph{boundary variable} makes this draw and a \emph{tie
variable} breaks ties~\cite{vovk2003mondrian,sadinle2019least,ding2023class,shi2024rc3p};
we call them the class's \emph{random keys}. They are derived from a stable
record identifier $\qid(x)$, such as a digest of the image bytes with an
occurrence index, so the same record always gets the same draws and an unchanged rule reproduces its
decisions exactly (supplement Sec.~B). We
write $\Set_t(x,\qid)\subseteq V_t$ for the prediction set and
$D_t(x,c;\qid)=\mathbf1\{c\in\Set_t(x,\qid)\}$ for the decision about
class $c$.

\begin{definition}[Prospective old-label consistency]
\label{def:contract}
A system is prospectively old-label consistent if every old class keeps the same
decision after every later registry update. Formally, for every $t'\ge t$, every
$c\in V_t$, and every input record $(x,\qid(x))$,
\begin{equation}
  D_{t'}(x,c;\qid)=D_t(x,c;\qid).
  \label{eq:prospective}
\end{equation}
Here everything fixed when $c$ was added (its descriptor, score, calibration
values, and random keys) stays fixed; see supplement Sec.~B.4, \emph{What
must remain fixed after a class is added}.
\end{definition}

The definition concerns old labels only: new labels may still join the set. It
is also stronger than replaying a log, because it covers images first seen
after the update.

\section{PrefixCP}
\label{sec:method}

PrefixCP has two parts: a scoring rule that keeps old decisions fixed, and a
computation that makes the rule cheap to serve.

\paragraph{1. Each class keeps the registry it was added with.}
Class $c$ is scored against the registry $V_{\birth(c)}$ that existed when it
was added:
\begin{equation}
  \score_c(z)=\ip{z}{e_c}-\energy_{\birth(c)}(z).
  \label{eq:prefixscore}
\end{equation}
We call $V_{\birth(c)}$ the \emph{reference} of class $c$. The reference
\emph{labels} are fixed; the normalizer is still computed for each image. Class $c$ keeps its onboarding rule: its calibration values and
random keys never change, and labels added later never enter its score. A new
class uses the full registry at the time it is added.

For example, let $V_1=\{\text{cat},\text{dog}\}$,
$V_2=V_1\cup\{\text{fox}\}$, and $V_3=V_2\cup\{\text{wolf}\}$. Cat and dog
keep $\energy_1$, fox keeps $\energy_2$, and wolf uses $\energy_3$. Adding wolf
changes every HPS score but no old PrefixCP rule. Had fox and wolf arrived
together, both would use the four-label registry: a different release history
gives different, equally stable rules.

Two baselines bracket PrefixCP. \emph{PRE} (projective reference energy)
never updates the reference: every
class, including later ones, is scored against the first registry $V_1$. HPS
always uses the newest registry for every class. PrefixCP sits between them:
old classes keep their old reference, and each new class gets the newest one.

\paragraph{2. All historical normalizers in one pass.}
Computing every historical normalizer separately costs
$\Theta(\sum_{u\le t}K_u)$ work. Because labels are appended in blocks, every
historical registry is a prefix of one list, so PrefixCP sums each block once
and accumulates the sums. With $\mathcal I_u=\{K_{u-1}+1,\ldots,K_u\}$ ($K_0=0$)
the labels added at version $u$ and $q_i=\ip{z}{e_i}/T$,
\begin{align}
  \beta_u
  &=\log\sum_{i\in\mathcal I_u}e^{q_i},
  \label{eq:block-summary}\\
  \Lambda_1&=\beta_1,
  \qquad
  \Lambda_u
  =\operatorname{logaddexp}(\Lambda_{u-1},\beta_u),
  \label{eq:prefix-summary}\\
  \energy_u(z)&=T\bigl(\Lambda_u-\log K_u\bigr),
  \label{eq:blocked}
\end{align}
with $\operatorname{logaddexp}(a,b)=\log(e^a+e^b)$. All normalizers then cost
$\Theta(K+\nver)$ after the similarities are computed. The scan is
classical~\cite{blelloch1990prefix}; our contribution is the scoring rule that
makes every required registry a prefix of one list (pseudocode and storage
bound: supplement Secs.~C.2 and D.4).

\section{Guarantees}
\label{sec:theory}

PrefixCP makes two promises that answer different questions. Coverage asks
whether a future image's true class is in the set, on average over the
calibration data, the image, and the random keys. Stability asks whether an
old decision stays the same, for every single image. We state both in exact
arithmetic; proofs and conditions for numerical replay are in supplement
Sec.~D.

\paragraph{Coverage, one class at a time.}
Fix a class $c$, and let $\mathcal F_c$ collect everything chosen before its
calibration examples are seen: the model, prompts, registry history,
temperature, and score.

\begin{theorem}[Exact marginal-within-class coverage]
\label{thm:coverage}
Suppose that, conditional on $(Y=c,\mathcal F_c)$, the $m_c$ calibration
examples and one future class-$c$ example are exchangeable. Also suppose that,
under the same conditioning, all tie and boundary variables are mutually
independent continuous uniforms, independent of the calibration and test
examples. Then the randomized classwise rank rule satisfies
\begin{equation}
  \Pr\!\left\{c\in\Set(X,\qid)\mid Y=c,\mathcal F_c\right\}=\cov .
  \label{eq:coverage}
\end{equation}
\end{theorem}

Exchangeability makes the image's rank uniform, and the boundary variable
includes exactly a $\cov$ fraction of ranks. Only ranks within a class matter,
so classes may use different references, even though their scores then no
longer form one probability vector. We call this marginal-within-class
guarantee simply \emph{class-conditional coverage}: it holds for each class, on average over
calibration draws. It does not hold for each individual image or under
arbitrary distribution shift.

\paragraph{Stability, image by image.}
Let $H_c(x,\qid)\in\{0,1\}$ be the decision of class $c$'s onboarding rule,
that is, the decision HPS makes for $c$ at version $\birth(c)$ with the same
calibration examples and random keys as PrefixCP.

\begin{theorem}[Equality to the onboarding rule]
\label{thm:serving}
Assume an append-only registry and that, after onboarding, class $c$ keeps its
descriptor, score definition, calibration values, and random-key rule fixed.
In exact arithmetic, for every class $c$, every later version
$t'\ge\birth(c)$, and every input record $(x,\qid)$,
\begin{equation}
  \mathbf 1\!\left\{c\in\Set^{\mathrm{PrefixCP}}_{t'}(x,\qid)\right\}
  =H_c(x,\qid).
  \label{eq:candidate-serving}
\end{equation}
\end{theorem}

Later releases change none of the inputs to an old rule, so its decision is
fixed for every future image, and no exchangeability is needed: PrefixCP is
prospectively old-label consistent (Def.~\ref{def:contract}). The equality
holds class by class; the full PrefixCP set need not equal any single HPS set.

\paragraph{Is PrefixCP's rule the only option?}
Among rules that choose each class's reference before seeing its calibration
data, PrefixCP's is the only one that both keeps old references fixed and gives
every new class the current registry: PRE fixes everything at $V_1$, and HPS
always moves to the newest registry (supplement Sec.~D.5, Prop.~1).

\section{Lifecycle audit}
\label{sec:experiments}

\begin{figure*}[t]
\centering
\begin{tikzpicture}[x=0.98cm,y=1cm,
 box/.style={draw=black!35,rounded corners=2pt,align=left,inner sep=4pt,font=\footnotesize,anchor=south west},
 lab/.style={font=\footnotesize,text=black!70}]
\draw[-{Latex[length=2mm]},black!60,thick] (0,0) -- (16.9,0) node[right,lab] {time};
\fill[black!60] (1.2,0) circle (2.2pt);
\node[anchor=north,lab] at (1.2,-0.1) {release 1: registry $V_1$};
\fill[black!60] (8.0,0) circle (2.2pt);
\node[anchor=north,lab] at (8.0,-0.1) {release 2: $V_2 = V_1 \cup{}$new classes};
\node[box,fill=wacvblue!5] at (0.1,0.3) {\textbf{Calibrate the old classes}\\$m$ labeled images per class\\from the calibration pool};
\node[box,fill=diagorange!7] at (5.75,0.3) {\textbf{Add and calibrate new classes}\\Old classes, per method:\\
 \textcolor{wacvblue!85!black}{PrefixCP}, \textcolor{newgreen}{PRE}, \textcolor{black!60}{raw}: rules kept\\
 \textcolor{diagorange!85!black}{HPS-Frozen: scores change}\\
 \textcolor{evictred}{HPS-Refresh: rescored and refit}};
\node[box,fill=black!4] at (11.75,0.3) {\textbf{Audit old-label decisions}\\Each image, each old label:\\same decision before and after?\\$\Rightarrow$ churn; also coverage, set size};
\draw[decorate,decoration={brace,mirror,amplitude=4pt},black!45] (0.1,-0.62) -- (7.85,-0.62)
  node[midway,below=6pt,lab] {historical images: seen before the update};
\draw[decorate,decoration={brace,mirror,amplitude=4pt},black!45] (8.15,-0.62) -- (16.9,-0.62)
  node[midway,below=6pt,lab] {new images: first seen after the update};
\end{tikzpicture}
\caption{\textbf{Audit protocol.} The old classes are calibrated at release 1.
At release 2 the new classes are added and calibrated, and each method either
keeps or changes the old classes' rules. The audit compares every old-label
decision before and after the update, on historical images and on new images
first seen after it. Calibration and evaluation images are disjoint; repeated
releases repeat this audit.}
\label{fig:setup}
\end{figure*}
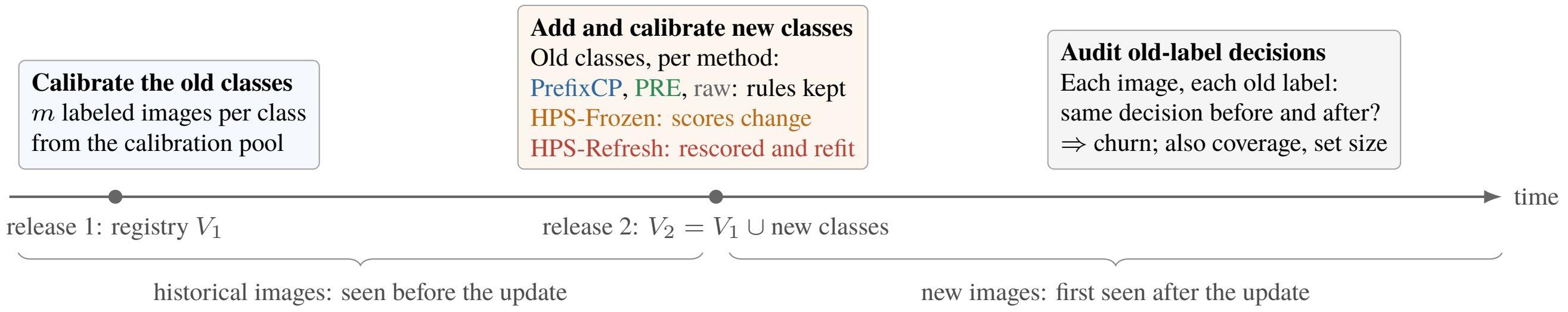

Our audit asks six questions, answered in
Secs.~\ref{sec:rq1}--\ref{sec:rq6}. Figure~\ref{fig:setup} shows the protocol,
and Table~\ref{tab:settings} lists the \N{ts.grids.word} \emph{development
settings} used while building the method; GeoDE (RQ6) was held out.

\begin{table*}[t]
\centering\small
\begin{tabular}{@{}lllll@{}}
\toprule
Dataset & Encoders & $m$ & Calibration / evaluation pools & Processing on both pools\\
\midrule
Food-101 & SigLIP & $20,40,80$ & Provided train / test & Blur severity 5 (radius 6)\\
Stanford Cars & CLIP, SigLIP & $20$ & Provided train / test & Blur severity 5 (radius 6)\\
CUB & CLIP, SigLIP & $20$ & Within-class identity-group halves & Blur severity 5 (radius 6)\\
ImageNet-R & CLIP, SigLIP & $20$ & Within-class identity-group halves & Natural renditions\\
ImageNet-Sketch & CLIP, SigLIP & $20$ & Within-class identity-group halves & Sketches\\
\bottomrule
\end{tabular}
\caption{\textbf{Settings.} CLIP is ViT-B/16; SigLIP is Base/16; $m$ is the
number of calibration images per class. Each setting uses five random half/half
class splits and target coverage $\cov=\N{ts.tau}$. Calibration and evaluation
pools are disjoint. For CUB, ImageNet-R, and Sketch we form them by splitting
each class's \emph{identity groups} (sets of pixel-identical images) in half,
so duplicates never straddle the two pools. Blur is a stress test applied to
both pools. Full protocol: supplement Sec.~N.}
\label{tab:settings}
\end{table*}

\paragraph{Data.}
We use Food-101~\cite{bossard2014food101}, Stanford Cars~\cite{krause2013cars},
CUB~\cite{wah2011cub}, ImageNet-R~\cite{hendrycks2021imagenetr}, and
ImageNet-Sketch~\cite{wang2019imagenetsketch}, with
CLIP~\cite{radford2021clip} and SigLIP~\cite{zhai2023siglip} encoders, both
ViT-B/16; for
SigLIP, softmax is a diagnostic choice rather than its native head. The first
three datasets are strongly blurred to make classification harder; the blur is
applied to calibration and evaluation images alike, so it is a stress test,
not a distribution shift. ImageNet-R and Sketch use their natural renditions
and sketches instead. Each
class has $m_c=m$ calibration images, with $m=20$ in most settings and also
$40$ and $80$ on Food-101. Calibration and evaluation images come from
disjoint pools of the same corpus (supplement Sec.~N).

\paragraph{Protocol.}
Seeds $100$--$104$ split each dataset's classes into an old half and a new
half, giving \N{ts.cells} runs. We onboard the old classes, add the new half,
and compare every old-label decision before and after the update. Churn is
measured on \emph{historical images}, seen before the update, and on
\emph{new images}, first seen after it; new images test what a log cannot.
Coverage is measured on the full evaluation split after the update. All methods share calibration examples within a run.

\paragraph{Methods.}
Three \emph{stable} methods keep old rules fixed: raw cosine (the similarity
$\ip{z}{e_c}$ with no normalizer), PRE, and PrefixCP. Two \emph{mutable}
methods update old classes (Sec.~\ref{sec:setting}). HPS-Refresh keeps a valid
coverage guarantee for the new registry but must keep the old calibration
images or their embeddings. HPS-Frozen's guarantee no longer holds, so we use
it only as a diagnostic, marked $^{\ddagger}$.
Conf-OT is compared separately in its own transductive setting
(Sec.~\ref{sec:rq3}).

\paragraph{Metrics.}
For each image and old label we compare the decision before and after the
update. The \emph{churn rate} is the fraction of these decisions that change;
the \emph{revocation rate} is the fraction of previously included old labels
that are excluded, and the \emph{addition rate} the fraction of previously
excluded ones that are included. Each image has many old labels, so even a
small churn rate can touch many images; the \emph{affected-image rate} is the
fraction of images with at least one changed old decision. \emph{Set size} is
the average number of labels returned, and \emph{macro coverage} averages
coverage over classes. Paired $95\%$ intervals vary the five class splits
of fixed data, not the datasets.

\subsection{RQ1: can coverage hide churn?}
\label{sec:rq1}

\begin{table*}[t]
\centering
\small
\setlength{\tabcolsep}{6pt}
\begin{tabular}{@{}l ccc c cc@{}}
\toprule
& \makecell[c]{revocation\\rate (\%)} & \makecell[c]{addition\\rate (\%)} &
\makecell[c]{affected-image\\rate (\%)} & \makecell[c]{worst\\coverage} &
\makecell[c]{valid\\coverage?} & \makecell[c]{old classes\\refit?}\\
\midrule
\N{tab.mutation.body}
\bottomrule
\end{tabular}
\caption{\textbf{Old-label decisions after one update.}
The first three columns measure changes on new images (lower is better). Ranges
span all \N{ts.grids.word} settings. The three stable methods change no old
decision in any of the $\N{ts.cells}$ runs. Rates are defined in Sec.~\ref{sec:experiments}
(Metrics). HPS-Frozen$^{\ddagger}$ rescores
old classes but keeps their old calibration, so its coverage guarantee no
longer applies; ``valid coverage?'' refers to that guarantee under
exchangeability, and ``by refit'' means HPS-Refresh keeps it only by refitting
every class. Worst coverage is the lowest realized coverage over all
classes on the post-update split; it is low for every method, including the
valid ones, because it picks the worst of many finite classes
(Sec.~\ref{sec:rq3}).}
\label{tab:mutation}
\end{table*}

\textbf{Yes.} The cleanest test keeps everything fixed except the normalizer.
In \N{cd.grids.word} settings, Food-101/SigLIP at three calibration sizes and
Stanford Cars with both encoders, we rescore old classes with the larger
registry but keep their calibration (HPS-Frozen), so only the normalizer
changes. On the full post-update split, the revocation rate is
$\N{cd.mean.rev.range}\%$ and the affected-image rate
$\N{cd.mean.reach.range}\%$, yet old-class coverage moves by at most
$\N{cd.mean.dcov.absmax.r2}$ (supplement Sec.~E). Coverage looks only at the true label; churn
looks at every old label. This shows that a small coverage change can hide
widespread churn; it does not make HPS-Frozen valid after the update.

Why are revocations so much more common than additions
(Table~\ref{tab:mutation})? Adding labels can lower the normalizer of
Eq.~\eqref{eq:active-normalizer} by at most $T\log(K_2/K_1)$, when every new
label is far from the image, but can raise it without bound when a new label is
close to the image. A higher normalizer lowers every old score, so old labels
mostly leave the set; without the $1/K_t$ factor (sum normalization) they can
only leave.

The \N{ts.grids.word}-setting audit (Table~\ref{tab:mutation}) measures new
images only, so its ranges differ, but the pattern is the same: HPS-Frozen's
revocation rate is $\N{ts.rev.frozen.range}\%$ and its affected-image rate
$\N{ts.reach.frozen.range}\%$. On three settings, two other conformal scores,
APS~\cite{romano2020adaptive} and RAPS~\cite{angelopoulos2021raps}, give the
same qualitative result; LAC~\cite{sadinle2019least} makes exactly the same
decisions as HPS without the $1/K_t$ factor (supplement Sec.~I).

\subsection{RQ2: which methods keep old decisions fixed?}
\label{sec:rq2}

\begin{figure*}[t]
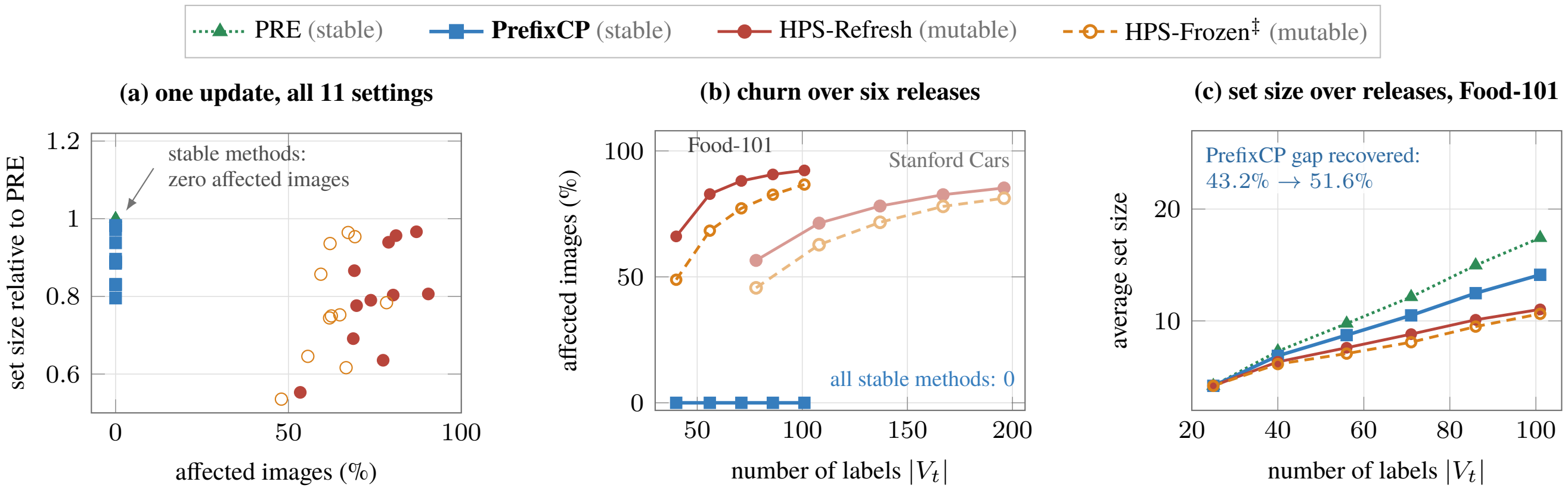

\centering
\pgfplotsset{
  gstyle/.style={
    width=0.235\textwidth, height=3.1cm, scale only axis,
    tick label style={font=\footnotesize}, label style={font=\footnotesize},
    title style={font=\footnotesize\bfseries, yshift=-1pt},
    grid=major, grid style={black!12, line width=0.3pt},
    axis line style={black!55}, tick style={black!55},
  }}
\begin{tikzpicture}
\begin{axis}[hide axis, scale only axis, width=1cm, height=0.1cm,
  xmin=0, xmax=1, ymin=0, ymax=1,
  legend columns=4, legend cell align=left,
  legend style={draw=black!25, fill=white, font=\footnotesize, inner sep=2pt,
    at={(0.5,0.5)}, anchor=center,
    /tikz/every even column/.append style={column sep=12pt}}]
\addlegendimage{newgreen, mark=triangle*, mark size=2pt, line width=0.9pt, densely dotted, mark options={solid}}
\addlegendentry{PRE \textcolor{black!55}{(stable)}}
\addlegendimage{wacvblue, mark=square*, mark size=1.8pt, line width=1.1pt}
\addlegendentry{\textbf{PrefixCP} \textcolor{black!55}{(stable)}}
\addlegendimage{evictred, mark=*, mark size=1.8pt, line width=0.9pt}
\addlegendentry{HPS-Refresh \textcolor{black!55}{(mutable)}}
\addlegendimage{diagorange, mark=o, mark size=1.9pt, line width=0.9pt, densely dashed, mark options={solid}}
\addlegendentry{HPS-Frozen$^{\ddagger}$ \textcolor{black!55}{(mutable)}}
\end{axis}
\end{tikzpicture}\\[3pt]
\begin{tikzpicture}
\begin{axis}[gstyle, title={(a) one update, all 11 settings},
  xlabel={affected images (\%)},
  ylabel={set size relative to PRE},
  xmin=-7, xmax=100, ymin=0.5, ymax=1.22]
\input{fig_pareto}
\draw[-{Latex[length=1.4mm]}, black!55] (axis cs:11,1.13) -- (axis cs:3,1.02);
\node[anchor=west, align=left, font=\scriptsize, text=black!70] at (axis cs:12,1.13)
  {stable methods:\\zero affected images};
\end{axis}
\end{tikzpicture}\hfill
\begin{tikzpicture}
\begin{axis}[gstyle, title={(b) churn over six releases},
  xlabel={number of labels $|V_t|$},
  ylabel={affected images (\%)},
  xmin=30, xmax=206, ymin=-3, ymax=108]
\input{fig_reach}
\node[anchor=south, font=\scriptsize, text=black!75] at (axis cs:66,95) {Food-101};
\node[anchor=south, font=\scriptsize, text=black!55] at (axis cs:170,89) {Stanford Cars};
\node[anchor=south west, font=\scriptsize, text=wacvblue] at (axis cs:108,2) {all stable methods: 0};
\end{axis}
\end{tikzpicture}\hfill
\begin{tikzpicture}
\begin{axis}[gstyle, title={(c) set size over releases, Food-101},
  xlabel={number of labels $|V_t|$}, ylabel={average set size},
  xmin=20, xmax=106, ymin=2, ymax=27]
\input{fig_size}
\node[anchor=north west,align=left,font=\scriptsize,text=wacvblue!85!black,fill=white,fill opacity=0.85,text opacity=1,inner sep=1.5pt]
  at (axis cs:22,26) {PrefixCP gap recovered:\\$\N{sg.food.closure.first}\%\rightarrow\N{sg.food.closure.last}\%$};
\end{axis}
\end{tikzpicture}
\caption{\textbf{Stability versus set size.}
\emph{(a)} Across all eleven settings, the stable methods change no old
decision, and the mutable methods return smaller sets by changing old
decisions. HPS-Frozen$^{\ddagger}$ has no valid coverage
guarantee; raw cosine lies off the plot at
$\N{fig.pareto.rawrel.range}\times$ PRE.
\emph{(b)} Over six releases, the share of affected images grows for the
mutable methods (lighter lines: Stanford Cars) and stays at zero for the
stable ones, including raw cosine (drawn for Food-101; also zero on Stanford
Cars).
\emph{(c)} Over the same releases, PrefixCP's set size stays between PRE and
HPS-Refresh and moves toward HPS-Refresh. In (b) and (c) the first registry
holds a quarter of the classes, not half as in Table~\ref{tab:frontier}, so
PRE's sets are larger.}
\label{fig:pareto}
\end{figure*}

\textbf{Only the three stable methods.} Raw cosine, PRE, and PrefixCP change no
old decision, on historical or new images, in any of the \N{ts.cells} runs
(new images: Table~\ref{tab:mutation}, Fig.~\ref{fig:pareto}a; historical:
supplement Sec.~F.5). PrefixCP also reproduces
each new class's independently fitted onboarding rule in every run (largest
score difference $\N{ts.onboarding.maxdelta}$). HPS-Refresh, despite valid
coverage, changes at least one old decision on $\N{ts.reach.refresh.range}\%$
of new images, more than HPS-Frozen, because it changes the cutoff as well as
the score and also draws new random keys. Fresh keys do not explain this away:
with every random draw replayed exactly, HPS-Refresh still changes at least
one old decision on $\N{sc.hpsmeanrefresh.reach}\%$ of images on three
settings (supplement Sec.~I). HPS-Frozen can return even smaller sets, but
only by losing valid coverage.

\subsection{RQ3: what does stability cost?}
\label{sec:rq3}

\begin{table*}[t]
\centering
\footnotesize
\setlength{\tabcolsep}{3.6pt}
\begin{tabular}{@{}l c ccccc c cc@{}}
\toprule
& & \multicolumn{5}{c}{average number of labels returned} & &
\multicolumn{2}{c}{paired set-size difference}\\
\cmidrule(lr){3-7}\cmidrule(lr){9-10}
& & \multicolumn{3}{c}{\emph{stable}} & \multicolumn{2}{c}{\emph{mutable}} & & &\\
\cmidrule(lr){3-5}\cmidrule(lr){6-7}
Setting ($\cov{=}\N{ts.tau}$) & $K$ & raw & PRE & \textbf{PrefixCP} &
HPS-Refresh & HPS-Frozen$^{\ddagger}$ & \makecell[c]{gap\\recovered (\%)} &
\makecell[c]{PrefixCP\\$-$ PRE} & \makecell[c]{PrefixCP $-$ HPS-Refresh\\($95\%$ CI)}\\
\midrule
\N{tab.frontier.body.m20}
\bottomrule
\end{tabular}
\caption{\textbf{Average number of labels returned after one update}
(smaller is better) for the nine $m{=}20$ settings. PrefixCP is smaller than
PRE and larger than HPS-Refresh in every setting. Gap recovery
(Eq.~\eqref{eq:closure-main}) uses HPS-Refresh, not HPS-Frozen$^{\ddagger}$,
as the small-set end. The last two columns are negative and positive in every
setting, so PrefixCP always lies between PRE and HPS-Refresh. Intervals vary the five class
splits of fixed data. The supplement
adds the $m{=}40,80$ Food-101 settings and coverage (Secs.~F.1--F.3).}
\label{tab:frontier}
\end{table*}

\textbf{Stability costs set size, and PrefixCP recovers about half of that
cost.} PRE is stable but returns large sets; HPS-Refresh returns smaller sets
but changes old decisions. We measure how far PrefixCP moves from PRE toward
HPS-Refresh. Let $M(A)$ be the average number of labels returned by method
$A$, over images and the five class splits, and define
\begin{equation}
  C
  =100\,
  \frac{M(\mathrm{PRE})-M(\mathrm{PrefixCP})}
       {M(\mathrm{PRE})-M(\mathrm{HPS\text{-}Refresh})}.
  \label{eq:closure-main}
\end{equation}
$C=0\%$ means PRE's set size and $C=100\%$ means HPS-Refresh's (computed only
when the denominator is clearly positive).

PrefixCP recovers $\N{ts.closed.refresh.range}\%$ of the gap across the eleven
settings, with zero churn (Fig.~\ref{fig:pareto}a; Table~\ref{tab:frontier}
shows the nine with $m=20$).
HPS-Refresh is smaller in every setting, and every paired interval for
PrefixCP minus HPS-Refresh lies above zero. On Food-101/SigLIP with $m=20$, for
example, PrefixCP returns $\N{ts.food-siglip-m20.prefix.r2}$ labels on average and
HPS-Refresh $\N{ts.food-siglip-m20.refresh.r2}$.

Why about half? An intuition: after one update, PrefixCP scores old labels
with the same reference as PRE and new labels with the same registry as
HPS-Refresh. Set size adds up over labels, so PrefixCP recovers roughly the
part of the gap that new labels contribute. With half the classes new, that is
about half whenever old and new labels contribute similarly. PrefixCP does not
repair classes that were already under-covered.

Conf-OT returns smaller sets on the three matched settings (Food-101/SigLIP
and Stanford Cars with both encoders), but it reads the whole
unlabeled test batch and guarantees only marginal coverage (supplement
Sec.~J.2).

\begin{table}[t]
\centering\small\setlength{\tabcolsep}{3pt}
\begin{tabular}{@{}rrrrrr@{}}
\toprule
$T/T_0$ & PRE & PrefixCP & \makecell{HPS-\\Refresh$^{\dagger}$} & \makecell{PrefixCP\\coverage} & \makecell{PrefixCP\\churn rate}\\
\midrule
\N{tab.main.temperature}
\bottomrule
\end{tabular}
\caption{\textbf{Temperature sensitivity} on Stanford Cars/SigLIP
($m=20$, target $0.9$; $T_0$ native). Labels per image and macro
coverage, averaged over five class splits; each temperature is calibrated
separately. $^{\dagger}$Run with its random keys held fixed;
expected set size does not depend on those keys, so its value at
$T/T_0=1.00$ matches Table~\ref{tab:frontier}.}
\label{tab:temperature}
\end{table}

\begin{table*}[t]
\centering\small\setlength{\tabcolsep}{4pt}
\begin{tabular}{@{}lrrrrl@{}}
\toprule
& & \multicolumn{2}{c}{PrefixCP} & & \\
\cmidrule(lr){3-4}
Setting & PRE & \makecell{all at\\once} & \makecell{one class\\per release} & \makecell{advantage over\\PRE used up (\%)} & \makecell[l]{worst tested\\order}\\
\midrule
\N{tab.main.schedule}
\bottomrule
\end{tabular}
\caption{\textbf{Release schedules.} Labels per image, averaged over
five class splits, with the final label set and the first registry fixed. We
compare PrefixCP with all new labels released at once against one class per
release in the worst tested order, named in the last column (near-base-first:
labels most similar to the first registry first; high-confusion-first: most
confusable first). ``Used up'' is the share of PrefixCP's advantage over PRE
lost by releasing one class at a time in that order; it is the worst tested
order, not a worst-case bound. Every schedule keeps its own old decisions
fixed.}
\label{tab:schedule}
\end{table*}

\paragraph{Which method should a deployment use?}
It depends on what a set triggers. In the catalog example, PrefixCP keeps the
review trigger working as the registry grows, at the cost of more candidate
labels to review. If changing old actions is acceptable and small sets matter
most, HPS-Refresh is the better choice, provided the old calibration images or
their embeddings are kept. We do not measure application costs, so we cannot
give a universal threshold.

\paragraph{Prompts and temperature.}
Stability, onboarding-rule equality, and the set-size ordering hold for three
prompt styles on three settings (supplement Sec.~F.8) and for the native
temperature and $\pm20\%$ on Stanford Cars/SigLIP (Table~\ref{tab:temperature}). Each variant is calibrated
separately; we do not claim stability \emph{across} a prompt or temperature
change.

\paragraph{Calibration redraws and worst-class coverage.}
Redrawing the calibration examples \N{cr.meta.n-calibration-draws} times on
Food-101/SigLIP and ImageNet-Sketch/SigLIP keeps PrefixCP stable and between PRE and
HPS-Refresh every time. Its worst-class coverage is
$\N{cr.food-siglip-m20.pfx.coverage.worst.ci}$ and
$\N{cr.sketch-siglip-m20.pfx.coverage.worst.ci}$, respectively (mean and
$2.5$th--$97.5$th percentiles over draws). Such low minima are not specific to
PrefixCP: every valid method in Table~\ref{tab:mutation} shows similar
worst-class values. Theorem~\ref{thm:coverage} promises coverage on average
over calibration draws, so the realized worst of many classes can fall well
below the target; these minima do not separate sampling variation from shift
(supplement Secs.~F.1 and G).

\subsection{RQ4: what happens over many releases?}
\label{sec:rq4}

\textbf{Old decisions stay fixed, but set size depends on the release
history.} Over six releases that grow the registry from a quarter of the
classes to all of them, PrefixCP and PRE never change an old decision,
while the share of images affected by HPS-Frozen and HPS-Refresh keeps growing
(Fig.~\ref{fig:pareto}b). PrefixCP also moves closer to HPS-Refresh in set
size: gap recovery rises from $\N{sg.food.closure.first}$ to
$\N{sg.food.closure.last}\%$ on Food-101 (Fig.~\ref{fig:pareto}c) and from
$\N{sg.cars.closure.first}$ to $\N{sg.cars.closure.last}\%$ on Stanford Cars
(supplement Sec.~F.6), in
line with the intuition of Sec.~\ref{sec:rq3}: each release adds classes whose
references are more recent than $V_1$. Each new class
matches its onboarding rule in all $\N{sg.cells}$ version--seed runs.

Because each class keeps the registry it was added with, the order and grouping
of releases decide which rules a deployment ends up with. Holding the final
label set and the first registry fixed, we compare \N{rc.schedules} release
schedules on three settings. They order the new labels randomly, by text
similarity to the first registry (nearest or farthest first), or
most-confusable first, and release them in batches of decreasing size, from
all at once to one class at a time.
Releasing one class at a time uses up $\N{rc.consumed.worst.range}\%$ of
PrefixCP's set-size advantage over PRE in the worst tested order
(Table~\ref{tab:schedule}) and $\N{rc.consumed.best.range}\%$ in the best
(supplement Sec.~H), but
PrefixCP stays smaller than PRE in every schedule.
On average, fewer, larger releases give smaller sets in every setting and
order we tested (individual seeds occasionally reverse this), consistent with
the same intuition: in a larger release, more classes are scored against a
larger, more recent registry.

Schedules shift mean set size by $\N{rc.meandelta.range}$ labels yet change
the full set on $\N{rc.imgdiff.range}\%$ of images, while macro coverage moves
by at most $\N{rc.cov.spread.max}$, the paper's theme again. These are designed
orders, not natural arrivals, so a deployment should fix its release policy
before seeing calibration results and audit its cost (supplement Secs.~D.5
and H).

\subsection{RQ5: what does PrefixCP cost to serve?}
\label{sec:rq5}

\textbf{Little.} We compare the shared prefix computation with a direct
implementation that computes each historical normalizer separately, using
identical rules and stored state. Once similarities are available, PrefixCP's
normalizers take $\N{rb.real.blockedoverfloor}\times$ the time of one
current-registry normalizer across the measured batch sizes, rather than
repeating one computation per release; the ratio rises with the number of
releases, and for small registries at batch size 1 a simple sequential scan
can be faster (supplement Sec.~K). On ImageNet-Sketch ($K=930$, batch size
$\N{rb.batch.real}$, one CPU thread), similarity plus normalizer time exceeds
that of HPS by $\N{rb.e2e.T2.overhead}\%$ with $\nver=2$ releases and
$\N{rb.e2e.T6.overhead}\%$ with $\nver=6$, excluding image encoding and
membership tests. PrefixCP stores the descriptors, the release boundaries, the
version at which each class was added, and fixed calibration values; old calibration images
can be deleted after fitting, whereas HPS-Refresh must keep them or their
embeddings.

\subsection{RQ6: do the findings transfer to a new dataset?}
\label{sec:rq6}

\textbf{Stability does; efficiency only partly.} We ran
GeoDE~\cite{ramaswamy2023geode}, which groups images by world region, after the
method was fixed, as an exploratory test with each region run separately.
Stability and onboarding-rule equality hold in all \N{gd.cells} runs
(\N{gd.regions} regions, \N{gd.seeds} seeds, two encoders), and PrefixCP is smaller than PRE in every region.
It recovers only $\N{gd.clip.all.closed}\%$ (CLIP) and
$\N{gd.siglip.all.closed}\%$ (SigLIP) of the gap, however, below the
development range. On images of new classes alone, raw cosine beats PrefixCP by
$\N{gd.siglip.novel.rawdelta.abs}$ labels on average on GeoDE/SigLIP, and in
all six regions,
although PrefixCP is smaller on all images for both encoders (supplement
Sec.~L). Efficiency therefore depends on the corpus.

\section{Limitations and conclusion}
\label{sec:limitations}

\noindent\textbf{Limitations.}
\emph{Assumptions:} PrefixCP requires append-only growth, a fixed encoder and
label descriptors, labeled data for each new class, and frozen old-class
calibration. Deleting labels, changing prompts, or updating the encoder needs
a new, recalibrated deployment outside Theorem~\ref{thm:serving}; renaming a
label for display only, with its identifier and descriptor unchanged, leaves
the rule unchanged.
\emph{Guarantees:} coverage is class-conditional and on average under
exchangeability; it does not hold per image or under arbitrary shift. A stable
rule also preserves its mistakes, and in our experiments smaller sets require
giving up stability. Exact-arithmetic identities do not imply byte-identical
replay across hardware. Stability holds for any arrival order; set size
depends on release history.
\emph{Evidence:} we test two ViT-B/16 encoders on the development settings;
calibration redraws cover two settings at one class split, designed release
orders cover three settings, and GeoDE is exploratory. We have not established
the quantitative findings for larger encoders or for natural arrivals of new
labels.

\noindent\textbf{Conclusion.}
When a label registry grows, coverage can stay nearly unchanged while many old
decisions change. Audits of growing registries should therefore report churn
on new images alongside coverage, release history, and set size. PrefixCP
shows that old decisions can be preserved without giving up class-conditional
coverage, at a measured set-size cost relative to recalibration. Whether to
recalibrate then becomes an explicit choice about which old actions a
deployment may change.

\clearpage
{
  \small
  \phantomsection\label{pagecheck:refstart}%
  \bibliographystyle{ieeenat_fullname}
  \bibliography{main}
}

\end{document}